\documentclass{article}
\usepackage{iclr2027_conference,times}
\iclrfinalcopy 

\usepackage{booktabs,amsmath,amssymb,graphicx,microtype,tikz}
\usepackage[table]{xcolor}     
\usepackage{array}             
\usepackage{longtable}         
\usepackage[group-separator={,}]{siunitx}

\definecolor{oursbg}{RGB}{233,240,250}
\usepackage{hyperref}
\usepackage{url}
\usepackage{caption}
\usepackage{alphalph}
\usepackage[section]{placeins}
\renewcommand{\topfraction}{0.92}
\renewcommand{\bottomfraction}{0.75}
\renewcommand{\textfraction}{0.06}
\renewcommand{\floatpagefraction}{0.72}

\newcommand{\addedpct}{11.9}

\newcommand{\brierdrop}{0.207}
\newcommand{\brierfull}{0.241}
\newcommand{\cddexlibfull}{57.1}
\newcommand{\cddexlibgain}{+13.9}
\newcommand{\cdlibdexfull}{74.1}
\newcommand{\cdlibdexgain}{+2.5}
\newcommand{\cdmean}{-1.34}
\newcommand{\cdn}{8}
\newcommand{\cdwins}{3}

\newcommand{\corrts}{0.86}

\newcommand{\devcount}{100}
\newcommand{\devlevel}{99.1}
\newcommand{\devmask}{100}

\newcommand{\dropssix}{-0.45}
\newcommand{\dropsthree}{-0.10}

\newcommand{\ecedrop}{0.114}
\newcommand{\eceful}{0.163}
\newcommand{\famk}{10}

\newcommand{\grladv}{72.80}
\newcommand{\grlctrl}{71.38}

\newcommand{\hppdr}{4963}

\newcommand{\hppearr}{0.358}

\newcommand{\hppgapan}{0}

\newcommand{\hppgapanwnthirty}{99}
\newcommand{\hppgapbt}{91}
\newcommand{\hppgapbthi}{201}
\newcommand{\hppgapbtlo}{34}
\newcommand{\hppgapbtpninezero}{398}
\newcommand{\hppgapbtwninety}{50}
\newcommand{\hppgapbtwnthirty}{23}

\newcommand{\hppmargdrall}{2.1}
\newcommand{\hppmargdrtight}{1.6}
\newcommand{\hppmargoball}{3.7}
\newcommand{\hppmargobtight}{3.8}
\newcommand{\hppmeanauc}{65.4}

\newcommand{\hppreps}{2000}

\newcommand{\hppsingledev}{9251}

\newcommand{\hpptightn}{2873}

\newcommand{\hppwins}{95445}
\newcommand{\hsicsa}{0.07}
\newcommand{\hsicta}{0.04}
\newcommand{\hsicts}{0.73}
\newcommand{\infparams}{506550}

\newcommand{\inlpdelta}{+0.63}
\newcommand{\inlpdeltahi}{+1.35}
\newcommand{\inlpdeltalo}{-0.09}
\newcommand{\inlpgain}{+4.19}
\newcommand{\inlpn}{12}
\newcommand{\inlpp}{0.079}
\newcommand{\inlpt}{1.93}
\newcommand{\inlpwins}{10}

\newcommand{\mdcfullseven}{68.06}

\newcommand{\mdcthree}{68.41}

\newcommand{\mdgainone}{+3.52}

\newcommand{\mdgainseven}{+0.21}

\newcommand{\meancsixwindow}{68.44}

\newcommand{\meancwindow}{68.01}

\newcommand{\objalone}{+0.71}

\newcommand{\objgivenvib}{+2.67}

\newcommand{\posthocdelta}{-0.08}
\newcommand{\posthocn}{12}
\newcommand{\posthocpos}{7}
\newcommand{\posthoct}{-0.32}
\newcommand{\ppgrbaseglucwithin}{0.437}
\newcommand{\ppgrbetweenhi}{0.699}
\newcommand{\ppgrbetweenlo}{0.550}
\newcommand{\ppgrbetweenvar}{22.9}
\newcommand{\ppgrcarbmax}{14,705}
\newcommand{\ppgrcarbpear}{0.072}
\newcommand{\ppgrcarbqfour}{1,911}
\newcommand{\ppgrcarbqone}{1,001}
\newcommand{\ppgrcarbspear}{0.197}
\newcommand{\ppgrcarbwins}{0.183}

\newcommand{\ppgrfmgain}{+0.0186}
\newcommand{\ppgrfmgainhi}{+0.0235}
\newcommand{\ppgrfmgainlo}{+0.0138}

\newcommand{\ppgrisolated}{270373}
\newcommand{\ppgritems}{2640959}
\newcommand{\ppgriters}{3}

\newcommand{\ppgrmeals}{136374}
\newcommand{\ppgrmednull}{148.5}
\newcommand{\ppgrmedreal}{1,026}
\newcommand{\ppgrnread}{13}
\newcommand{\ppgrobsfrac}{0.326}
\newcommand{\ppgroccasions}{828623}
\newcommand{\ppgrpeakmin}{45}
\newcommand{\ppgrpeaknull}{10.3}
\newcommand{\ppgrpeakreal}{26.2}
\newcommand{\ppgrpeakrise}{13.6}
\newcommand{\ppgrpermeal}{16.5}
\newcommand{\ppgrreps}{2000}
\newcommand{\ppgrretention}{0.985}
\newcommand{\ppgrsubj}{8282}
\newcommand{\ppgrwithinhi}{0.440}
\newcommand{\ppgrwithinlo}{0.096}
\newcommand{\ppgrwithinvar}{77.1}

\newcommand{\prdelta}{+1.16}
\newcommand{\prholm}{0.2080}
\newcommand{\prn}{14}
\newcommand{\prsigcount}{3}
\newcommand{\prwins}{9}

\newcommand{\roza}{68.88}

\newcommand{\rozspca}{+5.74}
\newcommand{\rozsrand}{+1.67}
\newcommand{\rozt}{74.19}
\newcommand{\roztgap}{1.65}
\newcommand{\roztpca}{+8.65}
\newcommand{\roztrand}{+3.27}
\newcommand{\rozts}{72.54}
\newcommand{\sdcellc}{1.56}

\newcommand{\sdcsixwindow}{0.53}

\newcommand{\sdcwindow}{0.15}

\newcommand{\shrinkpct}{29.7}
\newcommand{\trainparams}{806168}

\newcommand{\trdcp}{0.0161}

\newcommand{\trdropc}{4.82}

\newcommand{\vibalone}{+0.11}

\newcommand{\auc}{AUROC}
\newcommand{\zt}{z_{T}}\newcommand{\zs}{z_{S}}\newcommand{\za}{z_{A}}
\newcommand{\head}[1]{\vspace{2pt}\noindent\textbf{#1}}

\title{GlucoPRISM: Protocol-Supervised Pretraining for Cross-Cohort Glucose Transfer}
\author{%
\textbf{Adinath Madhavrao Dukre$^{1,2,*}$, Abdullah Mohammad$^{1,*}$, Yifan Lu$^{1}$,} \\
\textbf{Eran Segal$^{1}$, Ahmed Metwally$^{3}$, Imran Razzak$^{1,2}$} \\[4pt]
$^{1}$Mohamed bin Zayed University of Artificial Intelligence, Abu Dhabi, UAE \\
$^{2}$Vitalverse, Abu Dhabi, UAE \\
$^{3}$Google, USA \\
$^{*}$Equal contribution \\[2pt]
\texttt{\{adinath.dukre, imran.razzak\}@mbzuai.ac.ae}
}

\begin{document}
\maketitle
\lhead{} 
\renewcommand{\headrulewidth}{0pt} 
\begin{abstract}
Continuous glucose monitoring (CGM) foundation models encode each day as one vector that mixes who the person is, what happened that day and how the sensor measured it, so measurement differences travel with the representation into new cohorts. We ask whether pretraining can place the measurement component that harms transfer in known coordinates, so that deployment drops them with a fixed projection and no target data. GlucoPRISM supervises a GlucoFM backbone with response contracts, rules for how each part of the representation should change when only the day or only the sensor changes: a 64-dimensional Trait band should change with neither, a 48-dimensional State band only across days, and a 16-dimensional Sensor band should track the device. Synthetic sensor views calibrated on paired Dexcom and Libre recordings, different-day pairs and glucose-summary targets supply this supervision without clinical labels. Across twelve cross-cohort directions, dropping the Sensor band from the same frozen encoder raises mean transfer AUROC from 68.75 to 73.57, against 69.3 for GlucoFM. The gain has two parts. Into CGMacros, the one target cohort recorded with two sensor types, filtering adds 11 to 19 points, and this holds when the generator is recalibrated on half of the participants and scored on the other half. On the other eight directions filtering is neutral and the encoder itself exceeds GlucoFM by 3.5 points. The device remains linearly decodable after filtering, so the filter removes a device-related component that harms transfer rather than all measurement information. The retained bands improve window AUROC on thirteen of fourteen public task--cohort cells (Holm-adjusted $p=0.031$) and correlation on fourteen of fifteen clinical endpoints in an external cohort of 9,303 participants.

\end{abstract}

\section{Introduction}
\begin{figure}[t]\centering
\includegraphics[width=\linewidth]{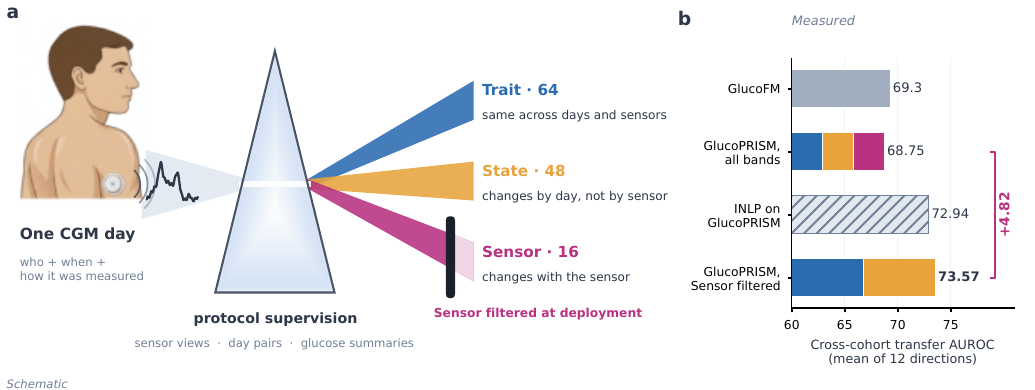}
\caption{GlucoPRISM treats a CGM day like light passing through a prism. (a) Schematic. A day of glucose readings from a CGM sensor worn on the back of the upper arm mixes who the person is, what happened that day and how it was measured. Protocol supervision disperses the representation into a Trait block (64 dimensions) that should not change across days or sensors, a State block (48) that may change across days but not across sensors, and a Sensor block (16) that tracks the device. At deployment the Sensor block is filtered out and the remaining 112 dimensions are used with the encoder frozen. (b) Measured. Mean cross-cohort transfer AUROC over twelve directions (Tables~\ref{tab:master} and~\ref{tab:erasure}). Bar colours show which blocks each readout keeps; the horizontal axis starts at 60. The average combines a large gain on the four directions into CGMacros with a neutral effect elsewhere (Section~\ref{sec:erasure}).}

\label{fig:teaser}
\end{figure}

Foundation models for continuous glucose monitoring (CGM) turn days of readings into reusable features for clinical prediction~\citep{lutsker2026gluformer,cgmformer2025,muhammad2026cgmjepa,li2026glucofm}. Their value depends on whether those features carry over to new cohorts, and cohorts differ in sensor hardware, sampling rate, calibration and study protocol. Measurement is also easy to learn: in the public corpus we use, a linear probe identifies the recording device from the observation mask alone at \devmask{} AUROC (Appendix~\ref{app:dependence}). A CGM day thus mixes three sources, the person, the day and the sensor, and they respond differently to change. The person persists across days and sensors, the events of a day persist across sensors only, and measurement changes with the sensor. Current CGM foundation models encode this mixture as one vector, so measurement moves with the model into every new cohort. Much as a prism separates light by how each wavelength bends, we separate these sources by how each responds to change.

Existing remedies act either after training or throughout the encoder. Post-hoc erasure fits a removal operator to labelled data~\citep{ravfogel2020null,belrose2023leace}, and invariance objectives suppress the nuisance during training, which can also remove useful signal~\citep{ganin2015dann,arjovsky2019irm}. We take a third route: \emph{if pretraining places a known nuisance in known coordinates, removing it at deployment becomes a fixed projection rather than an estimation problem.} GlucoPRISM does this with protocol supervision on a GlucoFM backbone~\citep{li2026glucofm}. Synthetic sensor views change only how a day is measured, different-day pairs change only the day, and each of three bands (Trait, State and Sensor) is trained to respond to these changes as its source would (Table~\ref{tab:refraction}). Glucose-summary targets further supervise Trait and State, and a Gaussian bottleneck bounds the Sensor band. At deployment we drop the 16 Sensor coordinates and keep the 112 Trait and State coordinates unchanged, so the full and filtered readouts differ only by the removed band.

On the same frozen encoder, removing the Sensor band raises mean cross-cohort AUROC from 68.75 to 73.57 across twelve transfer directions without fitting anything to the target cohort (Figure~\ref{fig:teaser}). The gain has two sources. On the four directions into CGMacros, the target cohort recorded with two sensor types, filtering adds 11 to 19 points, and recalibrating the generator without the scored participants leaves this gain intact. On the other eight directions filtering changes little, but the encoder itself already exceeds GlucoFM (70.8 against 67.3). The retained bands also carry physiology, within the public cohorts and in an external cohort of 9,303 participants. Our contributions are:

\begin{itemize}
\item \textbf{Response contracts for representation bands.} A response contract states how a band should change when only the day or only the sensor changes; each band is defined by one, and each contract is tested directly.

\item \textbf{A pretraining recipe with a target-free filter.} Sensor views, day pairs, glucose-summary targets and a Sensor bottleneck yield a measurement band that a projection fixed before any target data is seen removes at deployment.
\item \textbf{Evidence for where filtering helps and where it does not.} On the same frozen encoder, filtering raises transfer AUROC by 4.82 points, almost all of it on the two-sensor target cohort; elsewhere the advantage comes from the encoder. We test calibration leakage, fitted erasure and a domain-adversarial baseline, and report where the separation fails, including that the device stays decodable after filtering.

\end{itemize}

\section{Related Work}
\paragraph{CGM and time-series foundation models.} GluFormer, CGMformer and CGM-JEPA learn CGM representations by autoregressive, masked-token and latent prediction, X-CGM-JEPA adds a Glucodensity view~\citep{lutsker2026gluformer,cgmformer2025,muhammad2026cgmjepa}, and GlucoFM adds time alignment, observation masks and slow and residual streams~\citep{li2026glucofm}. Each returns a single vector in which the sources of variation remain mixed. We keep the GlucoFM backbone and change only how its representation is supervised and read out.

\paragraph{Measurement shortcuts and nuisance removal.} Models can rely on dataset or measurement features that do not transfer across sites~\citep{torralba2011unbiased,zech2018variable,degrave2021ai}, and in public CGM corpora the sensor is largely tied to the cohort. Domain-adversarial and invariant-risk objectives suppress such information throughout the encoder~\citep{ganin2015dann,arjovsky2019irm,krueger2021vrex}, and post-hoc erasure such as INLP and LEACE fits a removal operator afterwards~\citep{ravfogel2020null,belrose2023leace}. GlucoPRISM takes a third route: it places measurement information in a Sensor band whose coordinates are fixed during pretraining, so deployment only has to filter them. We compare directly with INLP on the same encoder.

\paragraph{Partitioned representations.} Domain Separation Networks and DIVA divide representations into shared, private or domain-specific parts~\citep{bousmalis2016dsn,ilse2020diva}, and variational bottlenecks limit what a block can hold~\citep{alemi2017deep}. Self-supervised learning with augmentations can isolate content that an augmentation preserves from style that it changes~\citep{vonkugelgen2021content,mitrovic2021relic}. GlucoPRISM differs in two ways: each block is defined by a response contract to simulated sensor changes and to changes of day rather than by domain labels, and the measurement block is removed at deployment by a projection fixed before any target data is seen. Because learned blocks are not identifiable as physiological factors without further assumptions~\citep{locatello2019challenging,locatello2020weakly}, we define them by response contracts and measure what each contains. OSF studies which input channels a sleep model sees during pretraining~\citep{shuai2026osf}; we change which learned features reach the predictor. Appendix~\ref{app:related} gives extended context.

\section{Method}
\subsection{Formulation: three bands from one encoder}
\label{sec:formulation}\label{sec:backbone}\label{sec:pooling}
We view a recorded day $x$ as a mixture of three sources: characteristics of the participant, events of that day and the measurement process of the sensor. Protocol supervision supplies two paired views: a sensor view $x'$ keeps the participant and the day and changes only the simulated measurement, and a day view $x''$ keeps the participant and changes the day. GlucoPRISM maps each day to three blocks, $z=[z_T;z_S;z_A]$, trained so that $z_T$ is unchanged under both views, $z_S$ is unchanged under the sensor view but may differ under the day view, and $z_A$ predicts the sensor (Table~\ref{tab:refraction}). We write Sensor as $z_A$, for acquisition, to keep $S$ for State.

The backbone is GlucoFM~\citep{li2026glucofm}. A 24-hour profile $x\in\mathbb{R}^{288}$ on a five-minute grid, with observation mask $m$, is split into 24 hourly patches; a causal, mask-aware filter separates slow and residual streams, and a three-layer transformer produces tokens $H\in\mathbb{R}^{24\times128}$. Masked context reconstruction $L_{\mathrm{MCR}}$ and temporal dynamics $L_{\mathrm{TD}}$ train the encoder against an exponential-moving-average target. We pool the tokens by coordinate-wise mean and standard deviation, project them to $u\in\mathbb{R}^{128}$ with a linear layer and LayerNorm, and split $u=[T_{64};S_{48};A_{16}]$. Trait and State pass through width-preserving MLPs $h_T,h_S$. Sensor is a Gaussian channel with mean $\mu$ and log-variance $\ell$, sampled during training and set to $\mu$ at evaluation, followed by an MLP $h_A$:
\begin{equation}
z=[z_T;z_S;z_A]=[h_T(T);h_S(S);h_A(A)]\in\mathbb{R}^{128},\qquad z_{TS}=[z_T;z_S]\in\mathbb{R}^{112}.
\label{eq:blocks}
\end{equation}
The full readout is $z$ and the filtered readout is $z_{TS}$; filtering leaves the Trait and State values unchanged. Appendix~\ref{app:method} gives the pooling, clipping and head details.

\begin{figure}[t]\centering
\includegraphics[width=\linewidth]{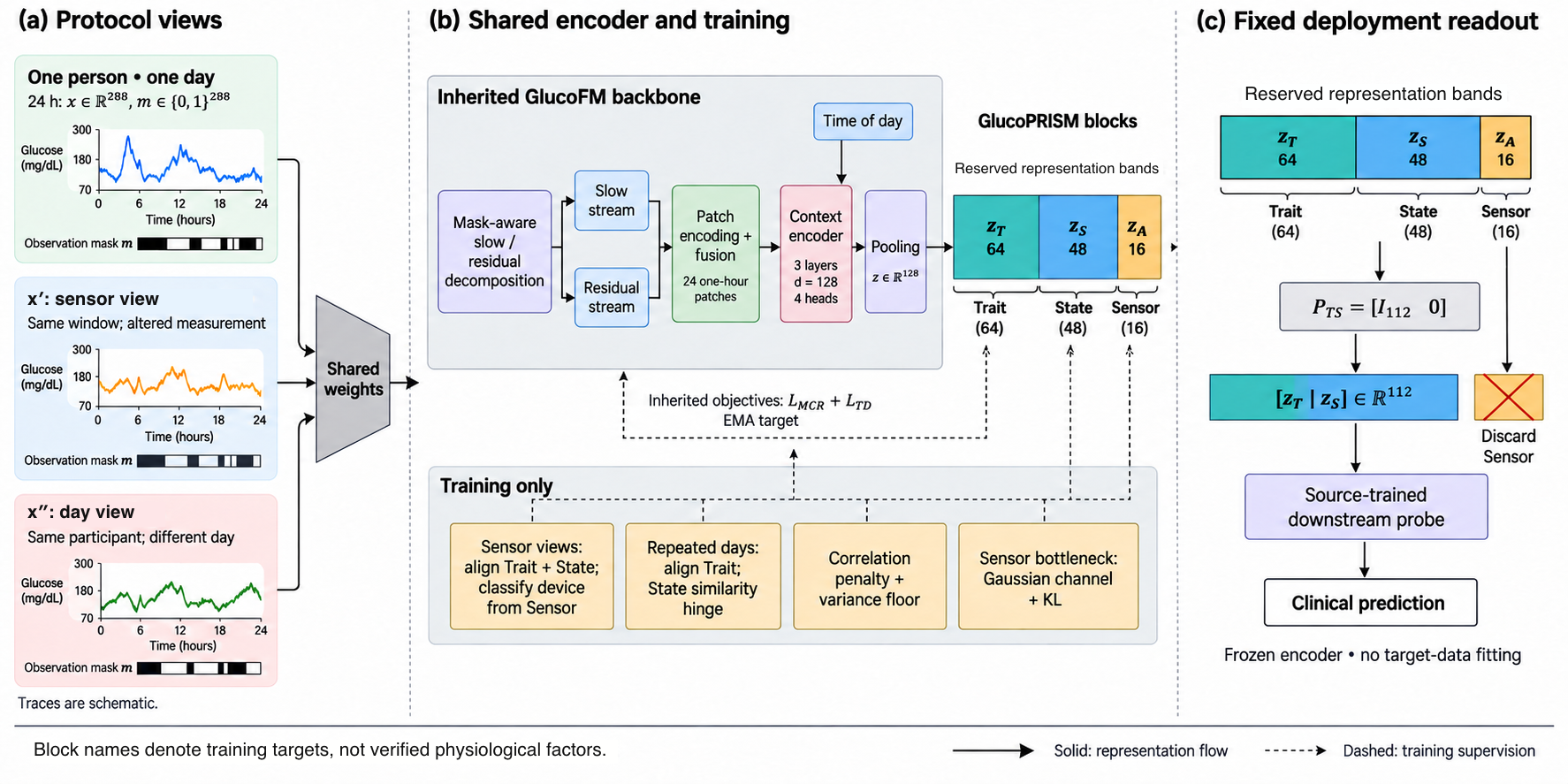}
\caption{GlucoPRISM pretraining and fixed readout. (a) Synthetic sensor views and repeated-day views provide distinct supervision. (b) The shared GlucoFM backbone produces a 128-dimensional representation partitioned into Trait (64), State (48) and Sensor (16); auxiliary objectives and a Sensor bottleneck shape these blocks during training. (c) At deployment the fixed filter $P_{TS}=[I_{112}\;0]$ removes the Sensor band and passes the 112 Trait and State dimensions to the downstream probe. The drawing summarizes the flow; statistical pooling (256 to 128), block projections and raw-block CMP heads are specified in the method. The Sensor mean and projection remain active during feature extraction. Traces are schematic.}
\label{fig:arch}
\end{figure}
\subsection{Protocol supervision: dispersing a CGM day}
\label{sec:views}\label{sec:objectives}
\begin{table}[t]\centering\small
\caption{Protocol supervision as response contracts. Each block is defined by how it should respond when only the recording day or only the sensor changes. The supervision terms are given in Equations~\ref{eq:sensor} to~\ref{eq:klclosed}.}
\label{tab:refraction}
\begin{tabular}{@{}p{0.15\linewidth}p{0.26\linewidth}p{0.26\linewidth}p{0.24\linewidth}@{}}
\toprule
& \textcolor[HTML]{2B6CB0}{\textbf{Trait}} (64) & \textcolor[HTML]{C7892A}{\textbf{State}} (48) & \textcolor[HTML]{B83280}{\textbf{Sensor}} (16) \\
\midrule
Across days & unchanged & may change & not constrained \\
Across sensors & unchanged & unchanged & changes \\
Supervision & day-pair alignment, sensor-view alignment, level targets & day contrast, sensor-view alignment, variability targets & device prediction, KL bottleneck \\
\bottomrule
\end{tabular}
\end{table}
A synthetic sensor partner is the same recorded day passed through a generator that mimics measurement differences: random-phase downsampling from five to fifteen minutes, smoothing, gain changes, compression dropout and low-frequency noise. Its parameters are fitted on 374 paired Dexcom and Libre windows from 44 CGMacros participants excluded from pretraining (Appendix~\ref{app:genrobust}). GlucoPRISM does not apply the measured level shift, so that sensor views do not change the glucose level that Trait is supervised to predict. With $D(a,b)=\mathbb{E}[1-\langle\hat a,\hat b\rangle]$ the cosine distance between unit-normalized features, sensor pairs align Trait and State while Sensor predicts the device label (synthetic partners receive the Libre label):
\begin{equation}
L_{\mathrm{sensor}}=D(z_T,z'_T)+D(z_S,z'_S)+L_{\mathrm{device}}.
\label{eq:sensor}
\end{equation}
A day pair takes a different day from the same participant. Trait is kept similar, and State is penalized when its similarity after a small head $g_S$ exceeds 0.3:
\begin{equation}
L_{\mathrm{day}}=\mathbb{E}_{\mathrm{eligible}}\!\left[1-\cos(z_T,z''_T)+0.5\max\{0,\cos(g_S(z_S),g_S(z''_S))-0.3\}\right].
\label{eq:day}
\end{equation}
Glucose-summary heads ($L_{\mathrm{CMP}}$, SmoothL1) make raw Trait predict five level and time-in-range summaries and raw State five variability summaries of the input; no clinical labels are used.

\subsection{Keeping the Sensor band narrow, and filtering it at deployment}
\label{sec:vib}\label{sec:readout}\label{sec:addressability}
A KL penalty keeps the Sensor Gaussian close to a standard normal, which bounds how much the band can hold:
\begin{equation}
L_{\mathrm{KL}}=\mathbb{E}_x\tfrac12\textstyle\sum_{j=1}^{16}(\mu_j^2+\exp\ell_j-\ell_j-1).
\label{eq:klclosed}
\end{equation}
A correlation penalty $L_{\mathrm{indep}}$ averages the squared batch correlations between coordinates of different raw blocks~\citep{zbontar2021barlow,bardes2022vicreg}, and a variance floor $L_{\mathrm{var}}$ keeps Trait and State coordinates from collapsing (Appendix~\ref{app:method}). The complete objective is
\begin{equation}
\begin{split}
L={}&L_{\mathrm{MCR}}+L_{\mathrm{TD}}+0.2L_{\mathrm{sensor}}+0.2L_{\mathrm{day}}\\
&+0.1L_{\mathrm{indep}}+0.1L_{\mathrm{KL}}+L_{\mathrm{var}}+L_{\mathrm{CMP}},
\end{split}\label{eq:total}
\end{equation}
trained with AdamW for 300 epochs (Appendix~\ref{app:impl}). At deployment the fixed matrix $P_{TS}=[I_{112}\;0]$ removes the Sensor band, $z_{TS}=P_{TS}z$, using no target data. Each readout receives its own probe fitted on the same source data, so comparing $z$ with $z_{TS}$ isolates the effect of filtering with the encoder fixed. The separation arises in the block projections: dropping the raw Sensor block before $h_A$ does not change transfer (Appendix~\ref{app:erasure-fm}).

\section{Experiments}\label{sec:eval}
\paragraph{Setup.} The prism view makes four predictions, which we test in turn: filtering the Sensor band should improve transfer between cohorts, most where source and target are measured differently (Section~\ref{sec:erasure}); the retained bands should keep physiological information in an external cohort (Section~\ref{sec:hpp}); each band should respond to changes of day and sensor as its contract specifies (Section~\ref{sec:spectrum}); and dispersion should cost nothing within a cohort (Section~\ref{sec:headline}). All encoders are pretrained on 514 participants and 10,952 daily windows from five public cohorts, frozen, and read by logistic probes on subject-disjoint folds (five folds repeated ten times, shared by all models, three matched pretraining seeds). The public benchmark has fourteen task--cohort cells from CGMacros, Stanford, Hall and ShanghaiT2DM. Transfer uses twelve source--target directions for diabetes risk and insulin resistance among CGMacros, Stanford and Hall, with the probe fitted on the source and scored on the target. GlucoFM is retrained on the same corpus. Stanford and ShanghaiT2DM appear in both pretraining and evaluation with participants intended to be disjoint; one participant overlaps, which changes 14-cell averages by at most 0.17 points (Appendix~\ref{app:corpus}). Unless stated otherwise, GlucoPRISM uses the filtered readout $z_{TS}$. Holm correction applies to the declared window-AUROC family, and other analyses are exploratory. Appendices~\ref{app:mainextra}, \ref{app:corpus} and~\ref{app:percell} give the evaluation map, protocol and per-cell results.

\begin{table}[!tbp]\centering\footnotesize\setlength{\tabcolsep}{2.6pt}
\caption{Frozen-encoder results averaged over fourteen within-cohort cells and twelve transfer directions. Window and Subject rows are within-cohort evaluations. GlucoPRISM uses the filtered readout $z_{TS}$. Trained models use three shared seeds. PR, ROC and F1 are reported in percentage points; $p_{\mathrm{Holm}}$ refers to the window-AUROC comparison with GlucoFM in the declared family. Subject-level and transfer tests are given separately in the appendix.}
\label{tab:master}

\resizebox{\linewidth}{!}{%
\begin{tabular}{@{}ll ccccccc ccc >{\columncolor{oursbg}}c@{}}
\toprule
& & \multicolumn{7}{c}{Frozen third-party checkpoints} & \multicolumn{3}{c}{Retrained, public corpus} & \textbf{Ours} \\
\cmidrule(lr){3-9}\cmidrule(lr){10-12}\cmidrule(lr){13-13}
& & \shortstack{Chronos-2\\small} & \shortstack{Chronos\\-2} & \shortstack{MOMENT\\large} & \shortstack{MOMENT\\small} & \shortstack{CGM-\\former} & Mantis & \shortstack{Mantis\\V2} & \shortstack{CGM-\\JEPA} & \shortstack{X-CGM-\\JEPA} & \shortstack{Gluco\\FM} & \shortstack{\textbf{Gluco}\\\textbf{PRISM}} \\
\midrule
Window & PR & 48.7 & 49.9 & 50.3 & 52.1 & 53.7 & 54.0 & 55.3 & 52.3 & 52.1 & 57.9 & \textbf{59.1} \\
& ROC & 58.5 & 59.6 & 59.9 & 62.5 & 62.9 & 63.6 & 64.7 & 60.9 & 60.8 & 65.9 & \textbf{68.0} \\
& F1 & 52.8 & 53.5 & 53.5 & 55.2 & 55.8 & 56.2 & 56.4 & 53.8 & 53.5 & 57.6 & \textbf{57.8} \\
\midrule
Subject & PR & 64.3 & 64.1 & 67.7 & 66.2 & 64.9 & \textbf{68.1} & 67.8 & 66.9 & 67.2 & 66.3 & 67.2 \\
& ROC & 66.1 & 65.7 & 69.3 & 67.3 & 66.3 & \textbf{70.0} & 69.2 & 67.3 & 67.8 & 69.0 & 69.4 \\
& F1 & 57.3 & 58.1 & 60.1 & 59.6 & 58.3 & 59.9 & 60.0 & 58.3 & \textbf{60.2} & \textbf{60.2} & 59.4 \\
\midrule
Transfer & ROC & 62.6 & 61.9 & 63.7 & 64.9 & 67.1 & 68.8 & 68.5 & 65.7 & 66.1 & 69.3 & \textbf{73.6} \\
\midrule
\multicolumn{2}{@{}l}{$p_{\mathrm{Holm}}$ (window ROC vs GlucoFM)} & 0.107 & 0.221 & 0.282 & 0.541 & 0.392 & 0.541 & 0.541 & 0.293 & 0.293 & --- & \textbf{0.031} \\
\bottomrule
\end{tabular}}

\end{table}
\subsection{Filtering the Sensor band improves cross-cohort transfer}\label{sec:erasure}
\begin{table}[!tbp]\centering\footnotesize\setlength{\tabcolsep}{3.4pt}
\caption{Readout interventions on GlucoPRISM: transfer AUROC, ECE and Brier score over three seeds. Differences are relative to the same encoder's full readout. INLP fits eight projections using real device labels and accesses target inputs when CGMacros is the target; the fixed Sensor filter does neither. The paired filter-minus-INLP AUROC difference is $\inlpdelta$ points (95\% interval $[\inlpdeltalo,\inlpdeltahi]$, $p=\inlpp$).}
\label{tab:erasure}

\begin{tabular}{lrrrr}
\toprule
Readout & Transfer ROC-AUC & $\Delta$ & ECE & Brier \\
\midrule
Full readout (all three bands) & 68.75\,{\scriptsize$\pm$}{\scriptsize 2.67} & --- & 0.163\,{\scriptsize$\pm$}{\scriptsize 0.022} & 0.241\,{\scriptsize$\pm$}{\scriptsize 0.014} \\
INLP \citep{ravfogel2020null}, 8 iterations & 72.94\,{\scriptsize$\pm$}{\scriptsize 0.40} & $+4.19$\,{\scriptsize$\pm$}{\scriptsize 2.44} & 0.136\,{\scriptsize$\pm$}{\scriptsize 0.024} & 0.218\,{\scriptsize$\pm$}{\scriptsize 0.009} \\
\textbf{Filtered readout (ours)} & \textbf{73.57}\,{\scriptsize$\pm$}{\scriptsize 0.54} & $+4.82$\,{\scriptsize$\pm$}{\scriptsize 2.18} & \textbf{0.114}\,{\scriptsize$\pm$}{\scriptsize 0.009} & \textbf{0.207}\,{\scriptsize$\pm$}{\scriptsize 0.006} \\
\bottomrule
\end{tabular}

\vspace{4pt}\par\footnotesize\begin{minipage}{\textwidth}When CGMacros is the transfer target, INLP fits its projection using target-domain inputs and device labels, but not target clinical-outcome labels. The Sensor filter requires no target-specific projection fitting. The synthetic-view generator nevertheless uses calibration data from CGMacros, as described in Section~\ref{sec:views}.\end{minipage}
\end{table}
\begin{table}[!tbp]\centering\footnotesize\setlength{\tabcolsep}{3pt}
\caption{Cross-dataset transfer following \citet{li2026glucofm}: a frozen-encoder probe is fitted on the source cohort and evaluated on the target cohort for diabetes and insulin resistance (IR). GlucoPRISM uses the filtered Trait--State readout $z_{TS}$. Values are means over three pretraining seeds in percentage points. \textbf{Bold} marks the best and \underline{underline} the second-best model in each row; the bracket gives GlucoPRISM's difference from the best baseline. ROC-AUC is shown; the full table with PR-AUC and seed standard deviations is Table~\ref{tab:xfer-sd} in Appendix~\ref{app:percell}. $^\dagger$ marks directions into CGMacros, the target cohort recorded with two sensor types.}
\label{tab:xfer}
\resizebox{\linewidth}{!}{%
\begin{tabular}{llrrrr>{\columncolor{oursbg}}r}
\toprule
Direction & Task & CGM-JEPA & X-CGM-JEPA & GluFormer (tiny) & GlucoFM & \textbf{GlucoPRISM} \\
\midrule
Stanford $\to$ Hall & Diabetes & 70.4 & 71.5 & \underline{72.0} & 68.4 & \textbf{76.6\,{\scriptsize(+4.6)}} \\
\addlinespace[1pt]
 & IR & \underline{63.9} & \textbf{64.0} & 60.7 & 61.7 & 62.1\,{\scriptsize($-$1.9)} \\
\midrule
Hall $\to$ Stanford & Diabetes & 63.9 & 64.0 & 64.7 & \underline{66.3} & \textbf{76.0\,{\scriptsize(+9.7)}} \\
\addlinespace[1pt]
 & IR & 56.6 & 56.7 & 56.2 & \textbf{58.2} & \underline{58.0\,{\scriptsize($-$0.2)}} \\
\midrule
CGMacros $\to$ Hall & Diabetes & 69.6 & 70.7 & 72.1 & \textbf{78.2} & \underline{78.0\,{\scriptsize($-$0.2)}} \\
\addlinespace[1pt]
 & IR & 61.1 & 61.6 & 63.2 & \underline{68.5} & \textbf{70.0\,{\scriptsize(+1.5)}} \\
\midrule
Hall $\to$ CGMacros$^\dagger$ & Diabetes & 75.2 & 75.7 & 72.6 & \underline{77.2} & \textbf{81.7\,{\scriptsize(+4.5)}} \\
\addlinespace[1pt]
 & IR & 73.2 & 73.0 & 71.9 & \textbf{80.3} & \underline{76.6\,{\scriptsize($-$3.7)}} \\
\midrule
CGMacros $\to$ Stanford & Diabetes & 64.6 & 64.5 & 69.4 & \underline{74.3} & \textbf{80.9\,{\scriptsize(+6.6)}} \\
\addlinespace[1pt]
 & IR & 52.3 & 52.5 & 61.9 & \underline{62.6} & \textbf{65.8\,{\scriptsize(+3.2)}} \\
\midrule
Stanford $\to$ CGMacros$^\dagger$ & Diabetes & 76.4 & 76.5 & \underline{77.0} & 71.1 & \textbf{85.6\,{\scriptsize(+8.6)}} \\
\addlinespace[1pt]
 & IR & 61.7 & 62.0 & \underline{70.0} & 65.3 & \textbf{71.4\,{\scriptsize(+1.4)}} \\
\midrule
\multicolumn{2}{l}{Average (12 directions)} & 65.7 & 66.1 & 67.6 & \underline{69.3} & \textbf{73.6\,{\scriptsize(+4.3)}} \\
\bottomrule
\end{tabular}}
\end{table}
\paragraph{Filtering improves the same encoder.} The cleanest test of the design holds everything fixed except the Sensor band. With the same encoder, source data and probe procedure, removing the 16 Sensor coordinates raises mean transfer AUROC from 68.75 to 73.57 and lowers expected calibration error from 0.163 to 0.114 and Brier score from 0.241 to 0.207 (Table~\ref{tab:erasure}). Averaged over all twelve directions the full readout is close to GlucoFM (68.75 against 69.3), but this average combines two opposite effects, which we separate below. INLP on the same encoder reaches 72.94 using device labels and, for CGMacros targets, target inputs; the fixed filter uses neither and is 0.63 points higher (95\% interval $[-0.09,1.35]$, $p=0.079$), so the two are comparable. In separate same-corpus retrains, erasure fitted to GlucoFM does not close the gap: GlucoFM with LEACE stays $3.94$ AUROC points below the fixed filter (95\% interval $[-6.77,-0.51]$), while LEACE on GlucoPRISM matches the filter ($-0.05$, $[-1.27,1.18]$; Appendix~\ref{app:erasure-fm}). In these retrains GlucoFM transfers about 2 points lower than in Table~\ref{tab:xfer}, so the gap is approximate; a domain-adversarial GlucoFM also stays near its baseline (67.19 against 73.14; Appendix~\ref{app:dann}).

\paragraph{Filtering helps where the target adds a second sensor.}
 The four largest filtering gains, +11.3 to +18.8 AUROC points in the rounded means of Table~\ref{tab:transfer}, are the four directions into CGMacros, the only transfer cohort recorded with both Dexcom and Libre sensors. There the full readout averages 64.7, below GlucoFM (73.5), and filtering raises it to 78.8. On the other eight directions filtering changes transfer by $+0.17$ (95\% interval $[-0.72,1.01]$, resampling directions and seeds), whereas the full readout already exceeds GlucoFM by $+3.49$ $[0.93,6.16]$ and is higher in 7 of 8 directions (Figure~\ref{fig:deletion}; Table~\ref{tab:xfer-groups}). This advantage depends on the protocol objectives: in a controlled retraining that removes one objective at a time, dropping sensor-view alignment lowers full-readout transfer by 5.82 points $[3.60,7.82]$ (Appendix~\ref{app:controlled-positive}), although that study does not separate the effect by direction.
 On same-day paired CGMacros windows, where only the sensor differs, filtering improves Dexcom-to-Libre transfer within the same encoder by \cddexlibgain{} points (Table~\ref{tab:crossdevice}). The generator was calibrated on CGMacros participants. Recalibrating it on 22 of the 44 paired participants and scoring CGMacros targets on the other 22 leaves the filtering gain unchanged ($+17.16$ against $+16.80$ for the original generator), and a generator whose only device knowledge is 15-minute Libre sampling keeps $+13.35$ (Appendix~\ref{app:gen-recal}). Both halves still come from one study with the same two devices, and on paired devices
 the filtered readout does not exceed GlucoFM on average (\cdmean{} points, \cdwins{} of \cdn{} comparisons won).

\paragraph{Width does not explain the gain.} Truncating GlucoFM to 112 or 64 dimensions changes its transfer AUROC by at most 0.16 points. Within GlucoPRISM, restoring the four least device-predictive Sensor coordinates brings transfer back to 68.49, close to the full readout (Appendices~\ref{app:controls} and~\ref{app:partial}).

\paragraph{Comparison with other CGM models.} Among the CGM models of Table~\ref{tab:xfer}, GlucoPRISM has the highest ROC-AUC in 8 of 12 directions and the highest PR-AUC in 9 of 12 (Appendix~\ref{app:percell}); counting the general time-series models of Table~\ref{tab:transfer}, the ROC-AUC count is 6 of 12. Average PR-AUC is 72.7 against 69.9 for GlucoFM. With twelve non-independent directions, no transfer comparison survives Holm correction (Table~\ref{tab:transfer}), so these rankings are exploratory.
 GlucoFM stays ahead for diabetes risk from CGMacros to Hall (by 0.2) and for insulin resistance from Hall to Stanford and from Hall to CGMacros. For every model, insulin resistance transfers worse on average than diabetes risk, which suggests that part of its shift lies in how the label is defined in each cohort rather than in how glucose is measured.

\subsection{The retained bands carry clinical information in an external cohort}\label{sec:hpp}
\begin{figure}[!tbp]
\centering
\includegraphics[width=\linewidth]{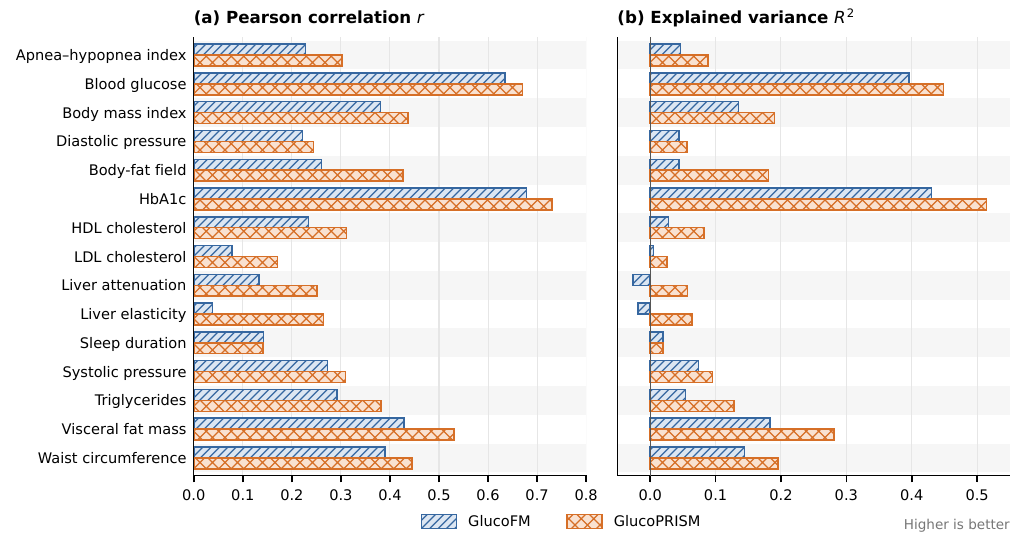}
\caption{Frozen-encoder prediction of all 15 baseline clinical endpoints on HPP. Paired bars show GlucoFM (blue, diagonal hatching) and GlucoPRISM (orange, cross-hatching) for each endpoint; the two panels show correlation and explained variance. Scores use embedding-only Ridge probes on the common test pool, with 301--1,214 labelled participants per endpoint. Labels are concurrent clinical measurements, not prospective disease outcomes. Bars are reported estimates without uncertainty intervals; the panels use different horizontal scales.}\label{fig:clinical-regression}
\end{figure}
\begin{table}[t]\centering\footnotesize\setlength{\tabcolsep}{5pt}
\caption{External frozen-encoder evaluation on HPP: subject-level AUROC from ten repetitions of five-fold subject-grouped probing. Clinical labels train the probes within HPP; encoders remain frozen and were not pretrained on HPP. DR: laboratory HbA1c $\geq5.7\%$; HL: hyperlipidemia; OB: obesity. Kept/deleted rows use the same encoder. Classical indices are shown with full wear and with input days matched to the model. These comparisons are descriptive and separate from the confirmatory family in Table~\ref{tab:master}.}
\label{tab:hpp}

\resizebox{\linewidth}{!}{%
\begin{tabular}{@{}l cccc ccccc >{\columncolor{oursbg}}c cccc@{}}
\toprule
& \multicolumn{4}{c}{No learned representation} & \multicolumn{5}{c}{CGM foundation models} & \multicolumn{5}{c}{GlucoPRISM readouts} \\
\cmidrule(lr){2-5}\cmidrule(lr){6-10}\cmidrule(lr){11-15}
& \shortstack{Mean\\glucose} & \shortstack{Raw\\window} & \shortstack{Indices\\full wear} & \shortstack{Indices\\day-matched} & \shortstack{CGM-\\JEPA} & \shortstack{X-CGM-\\JEPA} & \shortstack{Glu-\\Former} & \shortstack{\\GlucoFM} & \shortstack{GlucoFM\\trunc-112} & \shortstack{\textbf{Filtered}\\$[\zt\|\zs]$} & \shortstack{$\za$\\kept} & \shortstack{$\zt$\\only} & \shortstack{$\zs$\\only} & \shortstack{$\za$\\only} \\
\midrule
DR & 65.4 & 63.6 & 70.3 & 70.5 & 67.3 & 67.5 & 68.6 & 69.5 & 69.5 & \textbf{71.8} & 71.7 & 71.5 & 71.4 & 68.8 \\
HL & 53.1 & 53.0 & 59.7 & 58.2 & 55.8 & 56.7 & 58.0 & 59.8 & \textbf{59.9} & \textbf{59.9} & 59.7 & 59.7 & 59.5 & 58.1 \\
OB & 57.0 & 65.1 & 69.7 & 68.8 & 66.5 & 66.7 & 68.1 & 68.7 & 68.7 & 71.2 & \textbf{71.3} & 69.4 & 69.3 & 65.8 \\
\midrule
Avg & 58.5 & 60.6 & 66.6 & 65.8 & 63.2 & 63.6 & 64.9 & 66.0 & 66.0 & \textbf{67.6} & \textbf{67.6} & 66.9 & 66.7 & 64.2 \\
\bottomrule
\end{tabular}}

\vspace{4pt}\par\footnotesize\begin{minipage}{\textwidth}Per-endpoint PR-AUC, the cross-device split and fold standard deviations are reported in Table~\ref{tab:hppfull}.\end{minipage}
\end{table}
The Human Phenotype Project (HPP) tests whether filtering discards physiology. Its 9,303 participants were never seen in pretraining, and all were recorded with one manufacturer's 15-minute sensor, so filtering has little measurement shift to remove; an impoverished 112-dimensional readout would show up here as lost clinical signal. The public-corpus encoders stay frozen, and probes are fitted on HPP participants only.

\paragraph{Clinical classification.} The filtered readout reaches 67.6 mean AUROC across diabetes risk, hyperlipidemia and obesity, against 66.0 for GlucoFM and 63.2 to 64.9 for the other CGM foundation models (Table~\ref{tab:hpp}). It is the only model above classical CGM indices computed from the full wear (66.6); a single mean-glucose feature reaches 58.5. In participant-clustered paired bootstraps, the gain over GlucoFM is +2.24 AUROC for diabetes risk and +2.51 for obesity, both with intervals excluding zero (Appendix~\ref{app:hpp}). Full and filtered readouts score alike (67.6 each), consistent with little shift to remove.

\paragraph{Fifteen clinical endpoints.} Ridge probes predicting fifteen baseline measurements from the embedding alone improve Pearson correlation and $R^2$ on fourteen endpoints and mean absolute error on all fifteen (Figure~\ref{fig:clinical-regression}; 301--1,214 test participants per endpoint). The largest gains are in adiposity and liver phenotypes: body-fat correlation rises from 0.260 to 0.427, visceral fat from 0.429 to 0.531 and liver elasticity from 0.038 to 0.265, while HbA1c rises from 0.679 to 0.731 ($R^2$ 0.430 to 0.514). The one exception is sleep duration, where both models are weak (about 0.14). With age, sex and device group added as covariates, GlucoPRISM remains ahead on fourteen endpoints, and for two-year future levels it leads GlucoFM on twelve of fifteen correlations (Appendix~\ref{app:downstream-regression}). Trait and State together exceed either band alone on nine of fifteen endpoints (body fat: 0.367 for Trait, 0.378 for State, 0.427 jointly), so both physiological bands contribute (Appendix~\ref{app:block-functions}).

\paragraph{Where the gains stop.} The gains are in slow phenotypes. On the seven HPP meal-response and next-hour or next-day glucose tasks GlucoFM is stronger, and within-participant postprandial correlation is 0.212 for GlucoPRISM against 0.355 for GlucoFM (Appendices~\ref{app:downstream-regression} and~\ref{app:ppgr}). Under the error-based postprandial protocol of \citet{li2026glucofm}, GlucoPRISM has lower error than GlucoFM in 15 of 20 settings (Table~\ref{tab:ppgr-mae}).

\subsection{The bands respond as their contracts predict}\label{sec:spectrum}
\begin{figure}[t]\centering
\includegraphics[width=\linewidth]{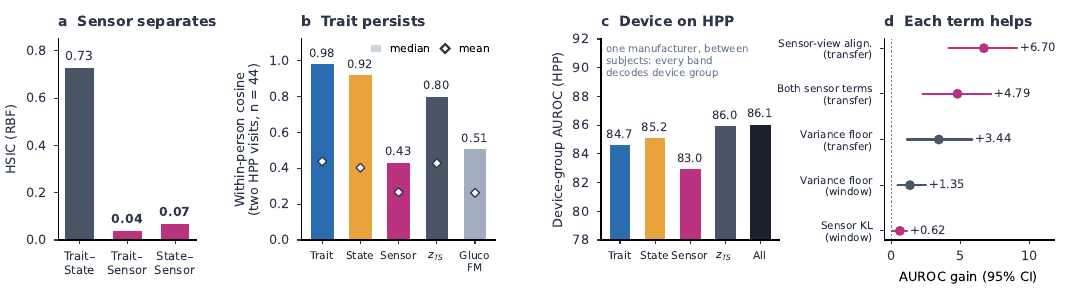}
\caption{Do the bands keep their contracts? (a) Kernel dependence (HSIC) between bands: Sensor is nearly independent of Trait and State, while Trait and State remain dependent. (b) Within-person cosine between two HPP clinic visits (44 participants; bars are medians, diamonds are means). (c) Device-group AUROC on HPP by readout. Device group varies between participants wearing one manufacturer's sensor, and every band decodes it, so this contract is not established in that setting. (d) AUROC gain from keeping each term in a controlled one-term-off study with its own protocol (24 transfer cells, three seeds); lines are 95\% participant-bootstrap intervals.}
\label{fig:contracts}
\end{figure}
\paragraph{The contracts hold where they are tested.} Table~\ref{tab:refraction} states how each band should respond when only the day or only the sensor changes. Sensor is nearly independent of Trait and State (HSIC \hsicta{} and \hsicsa{}), whereas Trait and State remain dependent (\hsicts{}; Figure~\ref{fig:contracts}a). Across two HPP clinic visits of 44 participants, the median within-person cosine is 0.985 for Trait, 0.923 for State and 0.435 for Sensor, and the filtered readout is more stable than GlucoFM (mean 0.428 against 0.263; Figure~\ref{fig:contracts}b). One contract is not established: on HPP, where device group varies between participants wearing one manufacturer's sensor, every single band decodes device group (83.0 to 85.2 AUROC; Figure~\ref{fig:contracts}c). For Dexcom and Libre it is only partly met: filtering improves transfer across the two devices, yet a linear probe still identifies the device from the filtered readout on same-day CGMacros pairs (99.97 AUROC; Appendix~\ref{app:dann}), so the filter removes a device-related component that harms transfer, not all device information.

\paragraph{Why a band rather than an adversary.} A linear probe identifies the device at \devmask{} AUROC from the observation mask alone, so measurement is available to any encoder. A gradient-reversal adversary on Sensor did not remove it and left the block more clinically readable (\grladv{} against \grlctrl{} AUROC; Appendix~\ref{app:extendedablations}); a domain-adversarial GlucoFM also fails to improve transfer (Section~\ref{sec:erasure}). GlucoPRISM therefore bounds the band with a bottleneck and removes it with a fixed filter.

\paragraph{Which terms matter.} Fitting the same block structure post hoc to a frozen encoder changes AUROC by \posthocdelta{} (Appendix~\ref{app:posthoc}), so the bands depend on pretraining with protocol views. In a controlled study that removes one term at a time, sensor-view alignment adds 6.70 points to filtered transfer AUROC and variance regularization adds 3.44, both with intervals above zero (Figure~\ref{fig:contracts}d; Appendix~\ref{app:controlled-positive}).

\subsection{Filtering is free within a cohort}\label{sec:headline}
Filtered and full readouts reach 68.01 and 67.90 window AUROC, so removing the band costs nothing where there is no measurement shift. Separately, the filtered readout exceeds GlucoFM (65.85) by 2.16 points (Holm-adjusted $p=0.031$) and improves thirteen of fourteen cells (Table~\ref{tab:master}; per-cell results in Table~\ref{tab:cellwise}). The PR-AUC gain is smaller and not significant after correction, and at subject level GlucoPRISM is competitive (69.4 against 69.0 for GlucoFM) but below Mantis (70.0). GlucoPRISM also has higher few-shot PR-AUC than GlucoFM at every support size and observation fraction (Figure~\ref{fig:fewshot}), and with multi-day subject representations it stays above GlucoFM at every number of days, with the largest gain from a single day (\mdgainone{} AUROC; Table~\ref{tab:multiday}).

\subsection{The transfer gap persists as the pretraining corpus grows}
In a single-seed study, we pretrain both models from scratch on nested HPP subsets of 500 to 8,000 participants  (up to 137k overlapping windows)
 and evaluate them on the public benchmark. GlucoPRISM has higher window and transfer scores at all nine sizes (Figure~\ref{fig:hpp_scaling}). Its transfer AUROC stays near 73.3, while GlucoFM's falls from 72.30 to 70.07; at 8,000 participants the transfer advantage is 3.24 points. One untested explanation is that every HPP subset shares one sensor protocol, which an undispersed encoder absorbs as the corpus grows and GlucoPRISM places in the filtered band.

\section{Discussion and Conclusion}\label{sec:discussion}
\paragraph{What the evidence supports.} The transfer advantage has two parts. Where source and target are measured differently (CGMacros targets, paired Dexcom-to-Libre windows), removing the Sensor band supplies the gain, and without filtering the encoder falls below GlucoFM there. Elsewhere filtering is roughly neutral and the advantage comes from the pretrained encoder itself.
 Both parts rest on twelve non-independent directions among three cohorts, and no transfer comparison survives Holm correction. Why the harmful device component becomes separable only after the block projections, while the device stays decodable, remains open.

\paragraph{Spectral leakage.} The separation is not complete. Trait and State remain dependent (HSIC \hsicts{}), and State predicts HbA1c nearly as well as the joint readout. Sensor also carries physiological signal: keeping all three bands gives the highest correlation on seven of fifteen HPP endpoints and improves every HPP meal-response and short-horizon task (Appendix~\ref{app:block-functions}), so fine temporal structure partly shares the band with measurement. Trait alone transfers slightly better than Trait and State together, and subject-level filtering changes within-cohort AUROC by \dropsthree{}, so the contribution of State to transfer is not yet established. Device and cohort are confounded in the public corpus, so Sensor is best read as a measurement-protocol band rather than a pure device band.

\paragraph{Scope and beyond glucose.} Transfer is evaluated among three public cohorts, and the generator is calibrated on CGMacros, the target cohort with the largest gains; recalibrating on half of its participants leaves the gain intact, but calibrating on a second paired cohort with other devices remains the most direct test.
 Clinical probes measure representation quality and do not establish clinical utility. The principle is not specific to glucose: when a nuisance can be simulated, even approximately, pretraining can give it an address, and response contracts make that address testable.

\paragraph{Conclusion.} GlucoPRISM disperses a CGM day into Trait, State and Sensor bands with protocol supervision and filters the Sensor band at deployment without target data.
 This improves transfer into cohorts with a second sensor type
 while keeping within-cohort accuracy and clinical information.

\clearpage
\subsubsection*{Reproducibility Statement}
Code, configuration files and weights (seeds 0 to 2, which reproduce the reported means, and the released seed-5 model) are provided at
 \url{https://anonymous.4open.science/r/glucoprism-4762}. Appendix~\ref{app:impl} lists every configuration constant, probe setting and seed, and Appendix~\ref{app:repro} describes the released checkpoints and feature exporter. The subject-to-fold assignment shared by all models is released so that future methods can be compared on the same splits. HPP requires controlled access through its data owner; only aggregate results left its trusted research environment, and the released HPP evaluation driver lets groups with access rerun the protocol.
\subsubsection*{AI Use Statement}
An AI assistant helped revise prose, reorganize the manuscript, check numerical consistency across tables and prepare figure layouts. The authors designed and ran all experiments and are responsible for the manuscript and artifacts.
\subsubsection*{Ethics Statement}
This work analyzes existing CGM cohorts. HPP participant-level data remain in its authorized research environment, with aggregate outputs exported. Clinical probes assess representation quality and do not establish clinical safety or efficacy. Device--population associations and cohort imbalance limit generalization.
\bibliographystyle{iclr2027_conference}
\bibliography{iclr2026_conference,revised_references_polished}
\clearpage

\appendix
\renewcommand{\FloatBarrier}{}
\renewcommand{\topfraction}{0.95}\renewcommand{\bottomfraction}{0.9}\renewcommand{\textfraction}{0.05}\renewcommand{\floatpagefraction}{0.85}
\setcounter{topnumber}{4}\setcounter{bottomnumber}{3}\setcounter{totalnumber}{6}
\setlength{\floatsep}{8pt plus 2pt minus 2pt}\setlength{\textfloatsep}{10pt plus 2pt minus 2pt}\setlength{\intextsep}{8pt plus 2pt minus 2pt}
\renewcommand{\thesection}{\AlphAlph{\value{section}}}
\paragraph{Guide to the supporting evidence.}
The appendix follows the prism view of the main text. Appendices~\ref{app:repro} to~\ref{app:corpus} describe the release, the objective, the paired-sensor generator and the evaluation protocol, and Appendix~\ref{app:percell} gives every public-cohort result and paired test. The filter itself is characterised by matched-width controls, fitted erasure, partial filtering and bottleneck sweeps (Appendices~\ref{app:controls} to~\ref{app:capacity}). How each band responds to changes of day and sensor is measured in Appendices~\ref{app:dependence} and~\ref{app:block-functions}, and the clinical information kept by the filtered readout in Appendices~\ref{app:hpp}, \ref{app:ppgr} and~\ref{app:downstream-regression}. Corpus and geometry studies are in Appendix~\ref{app:corpusfrac}, and the controlled study of individual objectives is in Appendix~\ref{app:controlled-positive}.

\section{Supplementary tables and figures for Section~\ref{sec:eval}}\label{app:mainextra}
This appendix collects the tables and figures that support the main-text results, in the order they are discussed: the evaluation map, per-cell within-cohort results, per-direction filtering gains, postprandial error, few-shot adaptation and the corpus-size study.

\begin{table}[!htbp]
\centering\small
\caption{Evaluation settings. Encoders are frozen for all downstream probes. Public refers to the five-cohort pretraining corpus; the public benchmark contains four evaluation cohorts. The two HPP uses have opposite train--evaluation directions.}
\label{tab:evaluation-map}
\begin{tabular}{@{}p{0.20\linewidth}p{0.19\linewidth}p{0.26\linewidth}p{0.25\linewidth}@{}}
\toprule
Evaluation & Encoder pretraining & Probe fitting and scoring & Question tested\\
\midrule
Public windows / subjects & Public corpus & Subject-disjoint folds within each cohort & Representation quality\\
Cross-cohort transfer & Public corpus & Fit on source; score on target & Filtering the Sensor band\\
External HPP & Public corpus & HPP training/test participants & Clinical information\\
HPP corpus-size study & Nested HPP subsets & Public benchmark splits and transfer directions & Transfer at larger pretraining sizes\\
\bottomrule
\end{tabular}
\end{table}
\begin{table}[!htbp]\centering
\caption{Linear-probe performance on every public task--cohort cell (window level, frozen encoders, subject-disjoint folds, three shared seeds for trained models). \textbf{Bold} marks the best and \underline{underline} the second-best value in each column for each metric. ROC-AUC cells are given to one decimal; PR-AUC and Macro-F1 cells are rounded to integers, and Avg is the unrounded 14-cell mean from Table~\ref{tab:master}. Tasks: DR diabetes risk, DR3 3-class diabetes risk, IR insulin resistance, $\beta$C $\beta$-cell dysfunction, HL hyperlipidemia, Ob obesity, Hy hypoglycemia, GT glucotype.}
\label{tab:cellwise}
\resizebox{\textwidth}{!}{%
\begin{tabular}{llrrrrrrrrrrrrrr>{\columncolor{oursbg}}r}
\toprule
& & \multicolumn{4}{c}{CGMacros} & \multicolumn{3}{c}{ShanghaiT2DM} & \multicolumn{3}{c}{Stanford} & \multicolumn{4}{c}{Hall} & \\
\cmidrule(lr){3-6}\cmidrule(lr){7-9}\cmidrule(lr){10-12}\cmidrule(lr){13-16}
Method & Metric & DR3 & HL & IR & Ob & HL & Hy & IR & $\beta$C & DR & IR & DR & GT & HL & IR & \textbf{Avg} \\
\midrule
\multicolumn{17}{l}{\emph{General time-series foundation models (frozen third-party checkpoints)}} \\
Chronos-2-small & PR & 55 & 28 & 84 & 61 & 32 & 16 & 61 & 60 & 66 & 58 & 51 & 49 & 20 & 41 & 48.7 \\
 & ROC & 70.0 & 46.8 & 68.9 & 58.7 & 49.5 & 53.3 & 52.2 & 63.2 & 64.8 & \textbf{67.3} & 64.5 & 59.0 & 50.4 & 50.1 & 58.5 \\
 & F1 & 48 & 47 & 61 & 54 & 48 & 50 & 50 & \underline{58} & 60 & \textbf{57} & 57 & 53 & 47 & 49 & 52.8 \\
\addlinespace[2pt]
Chronos-2 & PR & 56 & 29 & 90 & 64 & 31 & 18 & 61 & 59 & 65 & 55 & 49 & 51 & 23 & 47 & 49.9 \\
 & ROC & 71.3 & 50.3 & 78.2 & 62.3 & 48.1 & 58.5 & 52.1 & 59.6 & 62.3 & 61.4 & 59.4 & 59.8 & 50.2 & 61.2 & 59.6 \\
 & F1 & 50 & 50 & \underline{67} & 57 & 47 & 51 & 50 & 54 & 58 & 53 & 55 & 53 & 50 & 55 & 53.5 \\
\addlinespace[2pt]
MOMENT-small & PR & 59 & 30 & 88 & 67 & 35 & \underline{20} & 61 & 59 & 69 & 57 & 52 & 56 & \textbf{31} & 45 & 52.1 \\
 & ROC & 73.8 & 53.7 & 75.7 & \underline{65.3} & 51.3 & 57.6 & 53.8 & 62.2 & 66.8 & 65.4 & 61.9 & 60.8 & \textbf{68.4} & 57.6 & 62.5 \\
 & F1 & 53 & 50 & \underline{67} & \underline{59} & \textbf{50} & \underline{52} & \textbf{52} & 56 & 60 & \underline{56} & 56 & 56 & \textbf{54} & 53 & 55.2 \\
\addlinespace[2pt]
MOMENT-large & PR & 58 & 31 & 88 & 64 & 35 & 16 & 59 & 57 & 64 & 54 & 47 & 58 & 22 & 51 & 50.3 \\
 & ROC & 72.4 & 54.2 & 77.0 & 63.6 & 49.9 & 56.0 & 49.4 & 58.1 & 61.6 & 61.7 & 56.7 & 61.9 & \underline{55.3} & 61.3 & 59.9 \\
 & F1 & 52 & \underline{51} & 66 & 58 & \textbf{50} & 50 & 48 & 54 & 57 & 52 & 53 & 56 & 47 & 55 & 53.5 \\
\addlinespace[2pt]
Mantis & PR & 60 & 32 & 89 & 64 & 29 & \textbf{22} & 60 & 63 & 76 & 61 & 53 & 75 & 20 & 52 & 54.0 \\
 & ROC & 75.3 & 56.5 & 77.2 & 61.8 & 43.9 & \textbf{63.2} & 51.1 & 65.1 & \underline{73.7} & \underline{65.9} & 65.5 & 82.6 & 46.1 & 62.7 & 63.6 \\
 & F1 & 53 & \underline{51} & \underline{67} & 57 & 43 & \underline{52} & 49 & \textbf{59} & \underline{66} & \underline{56} & 58 & 71 & 48 & \underline{57} & 56.2 \\
\addlinespace[2pt]
MantisV2 & PR & 62 & 33 & \underline{91} & 62 & 35 & 18 & \textbf{65} & 63 & 74 & 57 & 56 & 82 & 22 & 53 & 55.3 \\
 & ROC & 76.4 & 57.8 & \underline{80.3} & 60.5 & 50.1 & 59.8 & \underline{55.1} & 65.2 & 72.7 & 61.8 & 65.9 & 85.6 & 52.3 & 62.8 & 64.7 \\
 & F1 & \underline{55} & \underline{51} & \underline{67} & 56 & 47 & 49 & \underline{51} & \underline{58} & 64 & 54 & 58 & 74 & 49 & 56 & 56.4 \\
\midrule
\multicolumn{17}{l}{\emph{Externally pretrained CGM model (frozen checkpoint)}} \\
CGMformer & PR & 59 & 32 & 89 & 61 & 36 & 14 & \underline{62} & 61 & 72 & 59 & 50 & 81 & \underline{25} & 50 & 53.7 \\
 & ROC & 74.0 & 55.2 & 77.5 & 60.0 & 52.6 & 52.4 & 54.5 & 62.2 & 69.9 & 61.2 & 61.1 & 88.6 & 54.5 & 56.9 & 62.9 \\
 & F1 & 53 & 50 & 66 & 56 & \textbf{50} & 48 & \underline{51} & 56 & 61 & 52 & 56 & 78 & \underline{51} & 54 & 55.8 \\
\midrule
\multicolumn{17}{l}{\emph{CGM foundation models retrained on our public-only corpus}} \\
CGM-JEPA & PR & 57 & 32 & 88 & 56 & 36 & 21 & 64 & 50 & 68 & 49 & 58 & \textbf{87} & 16 & 51 & 52.3 \\
 & ROC & 72.0 & 52.0 & 74.7 & 53.7 & 50.4 & 61.6 & 54.8 & 49.3 & 64.3 & 51.2 & 72.0 & \textbf{91.5} & 39.9 & 65.3 & 60.9 \\
 & F1 & 50 & 46 & 62 & 51 & 46 & 51 & 51 & 47 & 58 & 45 & 64 & \textbf{81} & 47 & 55 & 53.8 \\
\addlinespace[2pt]
X-CGM-JEPA & PR & 58 & 32 & 87 & 55 & 35 & 19 & 64 & 50 & 68 & 49 & 59 & \textbf{87} & 17 & 51 & 52.1 \\
 & ROC & 72.2 & 51.6 & 74.6 & 53.2 & 50.2 & \underline{62.2} & \textbf{55.2} & 48.8 & 63.5 & 51.1 & 72.1 & \underline{91.4} & 39.8 & 65.2 & 60.8 \\
 & F1 & 50 & 47 & 62 & 51 & 45 & 50 & 51 & 47 & 57 & 45 & 63 & 79 & 46 & 55 & 53.5 \\
\addlinespace[2pt]
GlucoFM & PR & \underline{64} & \textbf{42} & \underline{91} & \underline{68} & \underline{40} & 13 & 58 & \textbf{67} & \underline{77} & \textbf{64} & \underline{63} & \textbf{87} & 20 & \textbf{57} & \underline{57.9} \\
 & ROC & \underline{77.2} & \underline{57.9} & \underline{80.3} & 64.8 & \underline{59.4} & 52.2 & 46.0 & \underline{66.3} & 73.5 & 65.5 & \underline{75.1} & 90.6 & 45.0 & \textbf{68.0} & \underline{65.9} \\
 & F1 & \underline{55} & \textbf{55} & \underline{67} & \underline{59} & \underline{49} & 51 & 43 & \textbf{59} & 64 & \underline{56} & \underline{63} & \textbf{81} & 47 & \textbf{58} & \underline{57.6} \\
\midrule
\rowcolor{oursbg}\textbf{GlucoPRISM} & PR & \textbf{68} & \underline{41} & \textbf{92} & \textbf{69} & \textbf{45} & 14 & 59 & \underline{66} & \textbf{82} & \underline{63} & \textbf{66} & \underline{86} & 21 & \underline{56} & \textbf{59.1} \\
\rowcolor{oursbg} & ROC & \textbf{80.0} & \textbf{59.6} & \textbf{82.0} & \textbf{68.6} & \textbf{62.7} & 53.9 & 47.1 & \textbf{66.8} & \textbf{79.7} & \underline{65.9} & \textbf{77.8} & 90.7 & 51.3 & \underline{66.0} & \textbf{68.0} \\
\rowcolor{oursbg} & F1 & \textbf{56} & 48 & \textbf{70} & \textbf{61} & \underline{49} & \textbf{53} & 42 & \textbf{59} & \textbf{69} & 54 & \textbf{66} & \underline{79} & 48 & 56 & \textbf{57.8} \\
\bottomrule
\end{tabular}}
\end{table}
\begin{figure}[!htbp]\centering
\includegraphics[width=\linewidth]{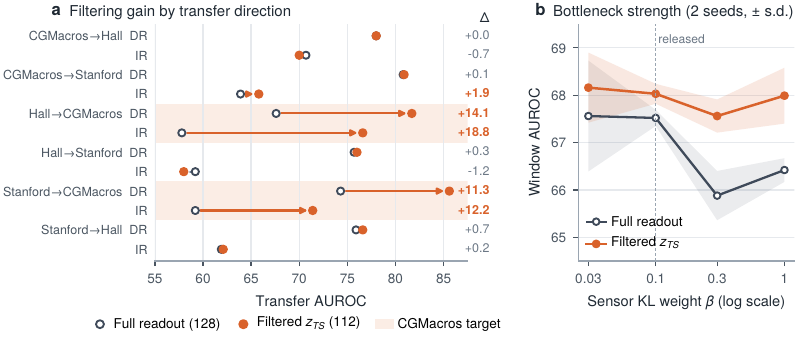}
\caption{Filtering the Sensor band. (a) Transfer AUROC of the full readout (open circles) and the filtered readout $z_{TS}$ (filled circles) for each direction, from the rounded means of Table~\ref{tab:transfer}; the column on the right gives the difference, and shading marks CGMacros targets. DR: diabetes risk; IR: insulin resistance. (b) Absolute AUROC in the separate two-seed KL sweep (mean $\pm$ seed SD). Panel (b) measures window AUROC, distinct from the transfer metric in panel (a). Larger gaps need not imply higher filtered-readout accuracy.}
\label{fig:deletion}
\end{figure}
\begin{table}[!htbp]\centering\footnotesize\setlength{\tabcolsep}{3.5pt}
\caption{Cross-cohort transfer ROC-AUC by target group, computed from per-seed, per-direction values over the three shared seeds. Differences are averaged over directions; brackets are 95\% intervals from resampling directions and seeds together ($n=8$, $n=4$ and $n=12$ directions). Directions share cohorts, so Table~\ref{tab:xfer-loo} also holds out one target cohort at a time.}
\label{tab:xfer-groups}
\resizebox{\linewidth}{!}{%
\begin{tabular}{lccc}
\toprule
& 8 directions not into CGMacros & 4 directions into CGMacros & All 12 directions \\
\midrule
Full readout & \underline{70.76} & 64.71 & 68.75 \\
Filtered $z_{TS}$ & \textbf{70.94} & \textbf{78.83} & \textbf{73.57} \\
GlucoFM & 67.27 & \underline{73.45} & \underline{69.33} \\
\midrule
Filtered $-$ full & $+0.17$ $[-0.72, +1.01]$ & $+14.12$ $[+5.03, +21.34]$ & $+4.82$ $[+1.02, +10.04]$ \\
Full $-$ GlucoFM & $+3.49$ $[+0.93, +6.16]$ & $-8.74$ $[-21.51, +4.24]$ & $-0.59$ $[-7.12, +4.49]$ \\
Filtered $-$ GlucoFM & $+3.66$ $[+0.73, +6.59]$ & $+5.38$ $[-3.09, +12.42]$ & $+4.24$ $[+0.86, +7.39]$ \\
Directions where filtered exceeds GlucoFM & 6/8 & 3/4 & 9/12 \\
\bottomrule
\end{tabular}}
\end{table}
\begin{table}[!htbp]\centering\footnotesize\setlength{\tabcolsep}{5pt}
\caption{Leave-one-target-cohort-out transfer: differences averaged over the directions whose target is not the held-out cohort, with 95\% intervals from resampling directions and seeds together. Filtered minus GlucoFM stays positive whichever target cohort is held out; its interval spans zero only when Stanford is held out.}
\label{tab:xfer-loo}
\begin{tabular}{lcc}
\toprule
Held-out target cohort & Filtered $-$ full & Filtered $-$ GlucoFM \\
\midrule
CGMacros & $+0.17$ $[-0.74, +1.01]$ & $+3.66$ $[+0.80, +6.55]$ \\
Stanford & $+7.09$ $[+1.44, +14.16]$ & $+3.94$ $[-0.56, +8.31]$ \\
Hall & $+7.20$ $[+1.74, +14.42]$ & $+5.11$ $[+0.56, +9.35]$ \\
\bottomrule
\end{tabular}
\end{table}

\begin{table}[!htbp]\centering\footnotesize\setlength{\tabcolsep}{3pt}
\caption{Postprandial glycemic response (PPGR) across cumulative context, averaged over Dexcom (5-min) and Libre (15-min) evaluations, following the protocol of \citet{li2026glucofm}. Entries are mean absolute error (lower is better) over 10 probe seeds; \textbf{bold} marks the best and \underline{underline} the second-best model in each row, ranked on unrounded values. All baselines are our reproductions on the public-only pretraining corpus, so they differ from published values. Device-specific results with seed standard deviations are in Appendix~\ref{app:ppgr}.}
\label{tab:ppgr-mae}
\resizebox{\linewidth}{!}{%
\begin{tabular}{llrrrr>{\columncolor{oursbg}}r}
\toprule
Context & Endpoint & GluFormer (tiny) & CGM-JEPA & X-CGM-JEPA & GlucoFM & \textbf{GlucoPRISM} \\
\midrule
Frozen representation & Trajectory (mg/dL) & 25.86 & \underline{23.73} & \textbf{23.43} & 25.84 & 25.87 \\
 & Positive iAUC (mg/dL$\cdot$h) & 36.05 & \underline{32.59} & \textbf{32.04} & 35.80 & 36.20 \\
 & Peak rise (mg/dL) & 32.59 & \underline{29.74} & \textbf{29.31} & 32.62 & 33.32 \\
 & Peak time (min) & 30.89 & 30.51 & \underline{30.47} & 31.27 & \textbf{30.36} \\
\midrule
+ 1 h pre-meal CGM & Trajectory (mg/dL) & 24.96 & \underline{23.27} & \textbf{23.08} & 23.46 & 23.31 \\
 & Positive iAUC (mg/dL$\cdot$h) & 34.69 & 32.16 & \textbf{31.86} & 32.20 & \underline{32.08} \\
 & Peak rise (mg/dL) & 30.79 & \underline{29.18} & \textbf{28.77} & 29.51 & 29.58 \\
 & Peak time (min) & 30.73 & 30.43 & \underline{30.20} & 30.41 & \textbf{29.75} \\
\midrule
+ Meal nutrition & Trajectory (mg/dL) & 23.45 & 21.84 & \textbf{21.72} & 21.92 & \underline{21.75} \\
 & Positive iAUC (mg/dL$\cdot$h) & 31.71 & 29.42 & \underline{29.23} & \textbf{29.14} & 29.23 \\
 & Peak rise (mg/dL) & 27.91 & 26.64 & \textbf{26.38} & 26.58 & \underline{26.45} \\
 & Peak time (min) & 30.60 & \underline{30.20} & 30.33 & 30.63 & \textbf{29.53} \\
\midrule
+ Fasting glucose & Trajectory (mg/dL) & 23.43 & 21.74 & \underline{21.67} & 21.93 & \textbf{21.59} \\
 & Positive iAUC (mg/dL$\cdot$h) & 31.72 & 29.33 & \underline{29.13} & 29.29 & \textbf{28.95} \\
 & Peak rise (mg/dL) & 27.71 & \underline{26.11} & \textbf{26.00} & 26.39 & 26.16 \\
 & Peak time (min) & 30.78 & \underline{29.48} & 29.72 & 30.01 & \textbf{29.28} \\
\midrule
+ BMI + diabetes status & Trajectory (mg/dL) & 23.74 & 22.12 & \textbf{21.91} & 22.27 & \underline{22.09} \\
 & Positive iAUC (mg/dL$\cdot$h) & 32.34 & 29.97 & \textbf{29.52} & 29.95 & \underline{29.91} \\
 & Peak rise (mg/dL) & 28.04 & 26.63 & \textbf{26.14} & 26.84 & \underline{26.59} \\
 & Peak time (min) & 30.71 & 30.27 & \underline{30.13} & 30.57 & \textbf{29.50} \\
\bottomrule
\end{tabular}}
\end{table}
\begin{figure}[!htbp]\centering
\includegraphics[width=0.86\linewidth]{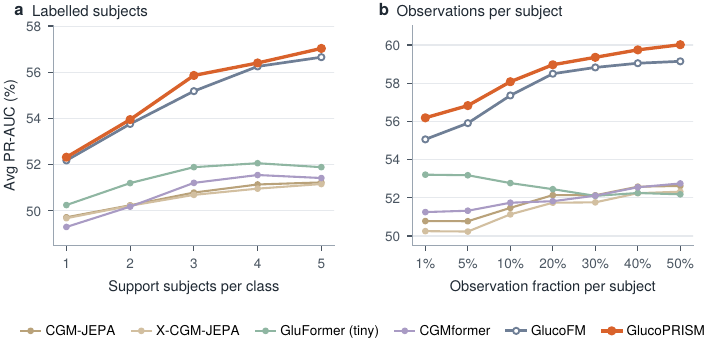}
\caption{Few-shot adaptation with frozen encoders, reported as task-averaged PR-AUC over the public within-cohort cells, following the protocol of \citet{li2026glucofm}. Left: number of labelled support subjects per class. Right: fraction of each subject's observations available for probe fitting. Points are mean estimates without uncertainty intervals; vertical axes differ between panels.}
\label{fig:fewshot}
\end{figure}
\begin{figure}[!htbp]\centering
\includegraphics[width=0.79\linewidth]{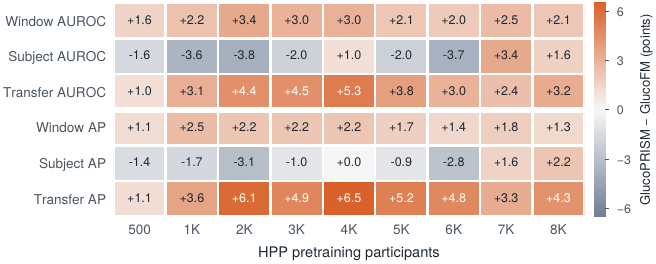}
\caption{Transfer advantage across HPP pretraining corpus sizes. Each row is one metric and each column one nested corpus size (500 to 8,000 participants); cells give GlucoPRISM minus GlucoFM in percentage points, with orange where GlucoPRISM is higher and grey-blue where GlucoFM is higher. Window and subject scores average 14 task--cohort cells; transfer scores average 12 directions. One pretraining seed is available, so no uncertainty or significance is inferred. Absolute scores are shown in Figure~\ref{fig:hpp_scaling_abs}.}
\label{fig:hpp_scaling}
\end{figure}

\section{Release, model identity and provenance}\label{app:repro}
GlucoPRISM uses Sensor KL weight 0.1 and does not apply the measured level shift in its synthetic sensor views.

The release contains the feature exporter used for all public-cohort results, the checkpoints for seeds 0, 1 and 2, from which every reported mean is computed, and the checkpoint distributed as the main GlucoPRISM model (seed 5). Seed 5 is the highest of the six trained seeds on public-benchmark window ROC-AUC, a selection rule fixed before release. On the same benchmark it scores 69.2 window, 70.6 subject and 74.9 transfer ROC-AUC, against 68.0, 69.4 and 73.6 for the three-seed means reported in the paper. Because it was selected on the benchmark, these scores are given for transparency and are not used in any comparison.
 The full and filtered exports have identical Trait and State coordinates, so the filter changes nothing except the removal of the Sensor band.


Table~\ref{tab:repro} preserves the reproduced and published baseline summaries. The published GlucoFM corpus includes non-public Wear-CGM data and differs from the public-only corpus here. Consequently, these columns are contextual comparisons rather than matched-corpus experiments.
\begin{table}[!htbp]\centering\footnotesize\setlength{\tabcolsep}{2.0pt}
\caption{\textbf{Retrained CGM baselines and published reference results.} Task-averaged over the same 14 task--cohort cells, the same frozen folds and the same untuned probe as everywhere else in this paper. \emph{Reproduced} is what this work produces; \emph{published} is GlucoFM's Table~3, whose pretraining corpus combines public data with non-public Wear-CGM recordings. It reports 477 subjects and 109,066 recording hours, compared with 514 subjects and 10,952 daily windows covering 237,879 distinct hours here.
 The published scores are therefore a reference under a different corpus and preprocessing pipeline, rather than a like-for-like column. The last two columns give per-cell agreement: mean absolute ROC-AUC difference over the 14 cells, and how many of them land within 2 points. \textbf{GlucoFM's lead over the next-best model reproduces at $+4.5$ \auc{} against $+4.4$ published}, which is the ordering this benchmark rests on. The three models beneath it span $0.5$ \auc{} here against $0.4$ published, so neither table resolves their internal order.}
\label{tab:repro}

\begin{tabular}{lrrrrrrrr}
\toprule
Model & Params & \multicolumn{3}{c}{Reproduced} & \multicolumn{2}{c}{Published} & mean & cells \\
\cmidrule(lr){3-5}\cmidrule(lr){6-7}
 & (M) & PR & ROC & F1 & PR & ROC & $|\Delta$ROC$|$ & $\leq 2$ \\
\midrule
\textbf{GlucoFM~\citep{li2026glucofm}} & 0.72 & \textbf{57.9}\,{\scriptsize$\pm$}{\scriptsize 0.6} & \textbf{65.9}\,{\scriptsize$\pm$}{\scriptsize 1.1} & \textbf{57.6}\,{\scriptsize$\pm$}{\scriptsize 0.6} & \textbf{58.8} & \textbf{66.7} & 3.53 & 5\,/\,14 \\
GluFormer-tiny~\citep{lutsker2026gluformer} & 0.65 & \underline{52.3}\,{\scriptsize$\pm$}{\scriptsize 0.3} & \underline{61.3}\,{\scriptsize$\pm$}{\scriptsize 0.5} & \underline{55.1}\,{\scriptsize$\pm$}{\scriptsize 0.2} & 53.0 & 61.9 & \underline{3.43} & 6\,/\,14 \\
CGM-JEPA~\citep{muhammad2026cgmjepa} & 0.52 & \underline{52.3}\,{\scriptsize$\pm$}{\scriptsize 0.1} & 60.9\,{\scriptsize$\pm$}{\scriptsize 0.2} & 53.8\,{\scriptsize$\pm$}{\scriptsize 0.3} & \underline{54.7} & \underline{62.4} & 3.47 & 5\,/\,14 \\
X-CGM-JEPA~\citep{muhammad2026cgmjepa} & 0.52 & 52.1\,{\scriptsize$\pm$}{\scriptsize 0.1} & 60.8\,{\scriptsize$\pm$}{\scriptsize 0.1} & 53.5\,{\scriptsize$\pm$}{\scriptsize 0.1} & 54.6 & 62.1 & \textbf{3.23} & 7\,/\,14 \\
\addlinespace
raw 288-point window & --- & 51.2 & 59.9 & 53.6 & --- & --- & --- & --- \\
observation mask only & --- & 45.8 & 51.0 & 44.6 & --- & --- & --- & --- \\
\bottomrule
\end{tabular}

\vspace{4pt}\par\footnotesize\begin{minipage}{\textwidth}CGM-JEPA and X-CGM-JEPA use the same encoder architecture (\num{522160} parameters) with different objectives. Their score difference is descriptive and is not an independent estimate of an objective effect.\end{minipage}
\end{table}
\section{Corpus, evaluation protocol and statistics}\label{app:corpus}

\begin{table}[!htbp]\centering\footnotesize\setlength{\tabcolsep}{2.6pt}
\caption{Pretraining corpus. All five cohorts are public and redistributable. ShanghaiT2DM's low observation fraction is not missingness: a 15-minute sensor can fill at most 96 of 288 five-minute grid slots, so coverage thresholds are applied relative to device rate. REPLACE-BG is capped at 40 windows per subject; uncapped it would contribute roughly 28,000 windows and turn the corpus into a type~1 diabetes distribution.}
\label{tab:corpus}

\begin{tabular}{llrrrr}
\toprule
Cohort & Device & Rate (min) & Subjects & Windows & Obs.\ frac. \\
\midrule
REPLACE-BG \citep{aleppo2017replacebg} & Dexcom G4 & 5 & 226 & 9,040 & 0.978 \\
Stanford \citep{hall2018glucotypes} & Dexcom & 5 & 27 & 622 & 0.983 \\
ShanghaiT2DM \citep{zhao2023shanghai} & FreeStyle Libre & 15 & 40 & 722 & 0.333 \\
Col\'as \citep{colas2019detection} & Medtronic iPro & 5 & 205 & 391 & 0.997 \\
BIG IDEAs \citep{bent2021bigideas} & Dexcom & 5 & 16 & 177 & 0.935 \\
\midrule
\textbf{Total (public only)} & --- & --- & \textbf{514} & \textbf{10,952} & --- \\
\addlinespace
\emph{GlucoFM's original corpus} & --- & --- & \emph{477} & --- & --- \\
\bottomrule
\end{tabular}

\vspace{4pt}\par\footnotesize\begin{minipage}{\textwidth}Our 10,952 windows cover 237,879 distinct hours; consecutive windows overlap by 40\% (Appendix~\ref{app:corpusfrac}), so this is less than 10,952 $\times$ 24. The original GlucoFM corpus has 109,066 recording hours and includes non-public Wear-CGM recordings.
\end{minipage}
\end{table}
Pretraining and downstream subject sets were constructed to be disjoint. Stanford and ShanghaiT2DM contribute participants to both, and one ShanghaiT2DM participant contributes one session to pretraining and a different session (ten windows) to evaluation. The main benchmark keeps these windows. Excluding them changes any model's ShanghaiT2DM results by at most 0.8 points and the 14-cell averages by at most 0.17, and GlucoPRISM's ShanghaiT2DM margin over GlucoFM rises from +2.05 to +2.28. The controlled study of Appendix~\ref{app:controlled-positive} excludes them.
 Membership-level verification is distinct from counting subjects, since counts alone cannot establish non-overlap~\citep{kaufman2012leakage}.
CGMacros and Hall never enter pretraining, and
CGMacros' real paired windows are reserved on top of that for the falsification
test. REPLACE-BG is capped at 40 windows per subject, because uncapped it would
supply roughly 28,000 windows and turn the corpus into a type~1 diabetes
distribution. ShanghaiT2DM's observation fraction of $0.333$ is not missingness.
It is the arithmetic consequence of a 15-minute sensor on a five-minute grid, so
we apply coverage thresholds relative to device rate.

Probing uses logistic regression with an $\ell_2$ penalty and lbfgs, no inner
search over $C$ and no class weighting, under 5-fold subject-grouped
cross-validation repeated 10 times. The subject-to-fold assignment was written
to disk before the first model was trained and is shared by every model,
baselines included. Shared assignments enable paired comparisons and account for variation common to the two models; they do not eliminate seed or fold uncertainty. We release the fold file so
that future methods can be compared paired rather than re-split. Holm correction applies to the declared confirmatory family. Ablation tests are reported without that correction and interpreted as exploratory; their uncorrected significance does not constitute confirmatory evidence.

\section{Implementation and hyperparameters}\label{app:impl}
\paragraph{Configuration and aggregation clarification.}
GlucoPRISM uses a Sensor KL weight of 0.1 with no synthetic level shift. It has six trained seeds; headline comparisons use the three seeds shared with every baseline (0, 1 and 2). The separately distributed seed-5 checkpoint and its selection rule are described in Appendix~\ref{app:repro}.
 The model rankings are descriptive and do not establish a deployment rule for unseen devices.

Internal subject-level probes concatenate the coordinate-wise mean, standard deviation, 10th and 90th percentiles of each subject's window embeddings. This produces $4d$ inputs, including 512 for a 128-dimensional embedding and 448 for the retained 112-dimensional readout. HPP uses the mean alone ($d$ inputs), while the multi-day analysis averages $K$ randomly sampled windows rather than adjacent days. Within each internal table all models use the same aggregator and subject-grouped folds, with scaler and probe fitted only on training subjects. The additional summary width is not separately controlled and may contribute to variability in small cohorts.

The following table lists the configuration constants.

\begingroup\footnotesize\setlength{\tabcolsep}{2.6pt}
\begin{longtable}{llr}
\caption{GlucoPRISM configuration, generator settings and probe parameters. HPP scaling uses 96 positions and patch width four instead of the public-corpus grid.}\label{tab:hpall}\\
\toprule
Setting & Value & Provenance \\
\midrule\endfirsthead
\multicolumn{3}{l}{\emph{\footnotesize Table~\ref{tab:hpall}, continued}}\\
\toprule
Setting & Value & Provenance \\
\midrule\endhead
\bottomrule\endlastfoot
\multicolumn{3}{l}{\textbf{Architecture}} \\
Window grid & 288 steps at 5 min (24 h) & \emph{inherited} \\
Patches & 24 of 12 steps (1 h) & \emph{inherited} \\
Causal filter $\sigma$ & init 6.0, learned in $[2.0, 12.0]$ & \emph{inherited} \\
Filter truncation / lookback & 3$\sigma$ / 9 steps & \emph{inherited} \\
Stream dim & 64 per stream & \emph{inherited} \\
Context encoder & 3 layers, $d=128$, 4 heads, FFN 256 & \emph{inherited} \\
Predictor & 1 layer & \emph{inherited} \\
Dropout & 0.1 & \emph{inherited} \\
\addlinespace
Block widths $(d_T, d_S, d_A)$ & $(64, 48, 16)$, summing to 128 & \textbf{ours} \\
Sensor bottleneck & variational; $\beta=0.1$; free bits 0.0 & \textbf{ours} \\
Sensor source & shared pooled Sensor slice & \textbf{ours} \\
Device classifier logits & 8 (three observed native labels) & \textbf{ours} \\
\addlinespace
Trainable at pretraining & \num{\trainparams} ($\addedpct$\% over backbone) & --- \\
Shipped at inference & \num{\infparams} ($\shrinkpct$\% \emph{below} backbone) & --- \\
\addlinespace
\multicolumn{3}{l}{\textbf{Pretraining optimisation}} \\
Optimizer & AdamW & \\
Learning rate & 0.0001 (filter $\sigma$: 0.001) & \\
Weight decay & 0.01 & \\
Warmup & 0.05 of total steps, then cosine & \\
Gradient clipping & 1.0 & \\
Batch size & 128 windows & \\
Epochs & 300 & \\
Mask ratio & sampled in $[0.5, 0.6]$ per batch & \\
EMA momentum & 0.997 $\rightarrow$ 1, scheduled & \\
Precision & fp32 & \\
Hardware & one NVIDIA T4 & \\
Wall-clock & $\approx$36 min per run & \\
\addlinespace
Seeds & $\{0,1,2\}$ for every arm; $\{3,4,5\}$ additionally for GlucoPRISM & \\
Model selection & held-out pretraining split only (released seed: App.~\ref{app:repro}) & \\

\addlinespace
\multicolumn{3}{l}{\textbf{Objective constants and the $V_1$ generator}} \\
$\lambda_1$ sensor, $\lambda_2$ day, $\lambda_3$ indep & $(0.2, 0.2, 0.1)$ & Eq.~\ref{eq:total} \\
VIB weight $\beta$ & 0.1 & Eq.~\ref{eq:total} \\
Variance floor weight & 1.0 & \S\ref{sec:objectives} \\
$D(\cdot,\cdot)$ & cosine on normalised blocks; see objective definitions & \\
State day hinge & cosine similarity above 0.3 & \\
Day-term hinge margin & 0.3 & \\
Day-term weight $\beta_d$ & 0.5 & \\
Sub-sampling stride & every 4 steps & \\
\addlinespace\multicolumn{3}{l}{\emph{$V_1$ synthetic paired-sensor generator, constants fitted on real pairs}} \\
\quad Decimation & $p=0.2$, keep 1 in 3, random phase & \\
\quad Compression drop & $p=0.1$, length [6, 12], depth [0.4, 0.7] & \\
\quad Dropout blocks & $p=0.25$, [1, 3] blocks of [6, 24] & \\
\quad Baseline wander & $p=0.25$, amplitude [5.0, 15.0] mg/dL & \\
\quad Level shift & not applied & App.~\ref{app:genrobust} \\
\quad Measured calibration & offset $-31.12$ mg/dL; gain $0.878\pm0.217$ & App.~\ref{app:genrobust} \\
\addlinespace
\multicolumn{3}{l}{\textbf{Downstream probe}} \\
Classifier & logistic regression, $\ell_2$ penalty, lbfgs & \\
Regularisation $C$ & 1.0 (scikit-learn default, no inner search) & \\
\texttt{max\_iter} & 2000 & \\
Standardisation & per-feature, fitted on train folds only & \\
Class weighting & none & \\
Cross-validation & 5-fold subject-grouped, $\times 10$ repeats & \\
Fold assignment & frozen before the first model trained & \\
Subject aggregation, internal & mean, SD, p10, p90 of embeddings; concatenate ($4d$) & \\
Subject aggregation, HPP & mean of window embeddings ($d$) & \\
Multi-day aggregation & mean of $K$ randomly drawn 24-hour windows & \\
Encoder & frozen; no fine-tuning anywhere in this paper & \\
\end{longtable}
\endgroup

\section{Seed variability}\label{app:seedsd}

We measured the seed-to-seed spread for every arm that trained
more than one model, with folds, probe and evaluation held fixed so the only
thing varying is the pretraining seed. We report two standard deviations there,
and they are not interchangeable. The
\emph{task-averaged} s.d.\ averages the 14 cells within a seed and then varies
the seed. It is the run-to-run variability of the quantity every headline
compares, and it is the error bar we attach to those headlines. The
\emph{per-cell} s.d.\ is the spread inside a single task--cohort cell, and runs
roughly tenfold larger
 ($\sdcellc$ against $\sdcwindow$ for GlucoPRISM),
because per-cell noise partly cancels once 14 cells are averaged. Reporting the
per-cell figure beside a task-averaged mean would overstate the uncertainty on
that mean by about that factor, so we keep the two apart.

GlucoPRISM ran six seeds where every comparator ran three,
and a six-seed mean cannot be paired against a three-seed one. The seed-matched
row is the one every headline, significance test and effect size in this paper
uses; the six-seed row is shown only so that the difference is visible. Between
the two, its window-level mean moves from $\meancwindow$ to $\meancsixwindow$
and its s.d.\ from $\sdcwindow$ to $\sdcsixwindow$. The extra seeds widen the
spread rather than shifting the estimate, which is what one expects when three
additional draws come from the same distribution.

\section{Objective implementation and notation}\label{app:method}
\paragraph{Pooling and feature export.}
Mean and sample standard deviation are computed across all 24 contextual tokens, including artificially masked positions. Their 256-dimensional concatenation passes through a learned projection and LayerNorm before the 64/48/16 split. Observation-density weighting and separate block normalization are disabled in the audited configurations. The Sensor channel uses a sample during training and its mean during evaluation. The public exporter concatenates the projected blocks; its full and retained features have identical Trait--State coordinates.

\paragraph{Repeated-day objective.}
Eligible pairs share the same subject identifier and have different day values. Another day is sampled uniformly, followed by a uniform window draw from that day. No minimum time gap or nonoverlap constraint is enforced. Both the masked anchor and unmasked partner use the online encoder. Equation~\ref{eq:day} averages eligible pairs in the batch and is skipped when none are available. The State hinge penalizes cosine similarity above 0.3, with internal weight 0.5 and overall day-loss weight 0.2, giving an effective hinge coefficient of 0.1. It does not require negative cosine similarity. The additional day head reads projected State; Trait alignment in this objective also uses projected features.

\paragraph{Variance floor.}
For a batch $Z\in\mathbb{R}^{B\times d}$, define $U_i=Z_i/\|Z_i\|_2$ and
\begin{equation}
\sigma_j=\sqrt{\operatorname{Var}_{B-1}(U_{\cdot j})+10^{-8}},\quad
\gamma_d=\frac1{\sqrt d},\quad
V(Z)=\frac1{d\gamma_d}\sum_j\max(0,\gamma_d-\sigma_j).
\label{eq:variance-exact}
\end{equation}
The implementation uses $L_{\mathrm{var}}=[V(T)+V(z_T)+V(S)+V(z_S)]/4$ on the masked anchor batch, with weight one. Thresholds are 0.125 for Trait and approximately 0.14434 for State. This operates on unit-normalized embeddings rather than imposing a raw-activation variance of one. Sensor is excluded under the active bottleneck. In the reported diagnostic without a variance floor, within- and between-subject cosine similarity both reach 0.9999.

\paragraph{Cross-block correlation and Sensor KL.}
The correlation penalty standardizes raw $T,S,A$ across the batch. Its mask excludes every within-block entry, including within-block off-diagonal terms. $L_{\mathrm{indep}}$ averages 9,728 entries, counting both symmetric directions. The sampled Sensor channel participates during training. KL is evaluated analytically from its mean and clipped log-variance, summed over dimensions, averaged within a view, and then averaged equally across available views. The coefficient is 0.1. The information-bound interpretation applies to the stochastic channel, not automatically to its deterministic evaluation mean.

\paragraph{Summary and device heads.}
CMP sums two independently mean-reduced SmoothL1 losses with overall weight one. Five level/occupancy targets supervise raw Trait and five dispersion targets supervise raw State; circadian CMP targets are disabled. The device head maps projected Sensor to eight logits. Native labels in the audited full corpus are Dexcom (9,839 windows), Libre (722) and iPro (391); synthetic partners receive Libre. The other configured outputs do not denote additional observed or synthetic device classes.

\paragraph{Level preservation.}
Input normalization and representation normalization serve different roles. Inputs preserve glucose level in mg/dL. Under the reported $+60$ mg/dL perturbation, the representation changes by 0.136 under level-preserving preprocessing and by $8.7\times10^{-6}$ when that level information is removed. This is an offset-sensitivity control, separate from clinical prediction performance.

\section{The paired-sensor generator, and its robustness}\label{app:genrobust}

\begin{figure}[!htbp]\centering
\includegraphics[width=0.76\linewidth]{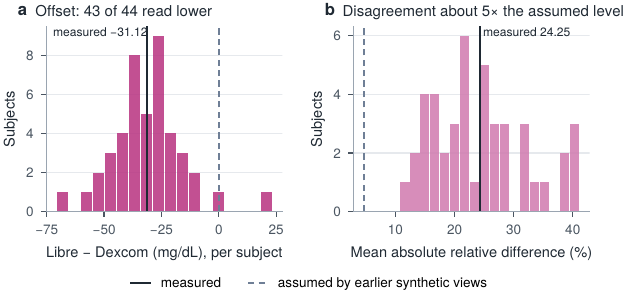}
\caption{Real Dexcom--Libre disagreement on same-day paired windows. The offset
is systematic and one-directional rather than symmetric noise, which is what
makes it a learnable device signature rather than a nuisance that averages out.}
\label{fig:sensor}
\end{figure}
Two preprocessing choices affect the paired-sensor estimates. First, array
index~0 denotes a window's first reading rather than midnight. A
Dexcom window beginning at 11:39 and its same-day Libre partner beginning at
10:30 are offset by roughly 14 grid steps. Left uncorrected, the measured
correlation reads $0.31$ with the estimated lag pegged at the search boundary.
Second, ordinary least
squares is attenuated when both variables carry noise: OLS reports a slope of
$0.674$ where Deming
regression~\citep{deming1943statistical,bland1986statistical} reports $0.878$.
Fitting on the OLS slope would have built a systematically flattened sensor.
\begin{table}[!htbp]\centering\footnotesize\setlength{\tabcolsep}{5pt}
\caption{Subject-level sensitivity of the paired-sensor generator. Parameters are fitted on 374 windows from 44 subjects and refitted using leave-one-subject-out analysis and subject bootstrap resampling. The calibration offset remains negative in all 2,000 bootstrap refits; the reported ranges quantify variation in its magnitude and the other fitted parameters.}
\label{tab:genrobust}

\begin{tabular}{lrrr}
\toprule
Fitted parameter & Point estimate & Leave-one-subject-out range & 95\% bootstrap \\
\midrule
Calibration offset (mg/dL) & -31.12 & [-32.39, -30.20] & [-34.60, -27.71] \\
Deming slope & 0.88 & [0.86, 0.89] & [0.83, 0.92] \\
Mean abs.\ relative difference (\%) & 24.25 & [23.86, 24.56] & [22.54, 25.99] \\
Correlation with source window & 0.74 & [0.73, 0.75] & [0.69, 0.78] \\
\bottomrule
\end{tabular}

\vspace{4pt}\par\footnotesize\begin{minipage}{\textwidth}43 of 44 individual subjects show the offset in the same direction, so the effect is not carried by a few outlying people.\end{minipage}
\end{table}

\subsection{Calibration without the evaluated participants}\label{app:gen-recal}
The largest filtering gains occur on CGMacros targets, and the generator was calibrated on CGMacros pairs. To test whether the gain depends on that calibration, GlucoPRISM was retrained from scratch with three seeds per arm, with everything else unchanged, under two generators that do not use the evaluated participants (Table~\ref{tab:gen-recal-arms}). No second cohort with same-day Dexcom and Libre pairs is available, so the two arms are (a) an uncalibrated generator whose only device knowledge is the 15-minute Libre sampling rate, a public device specification, and (b) a generator calibrated on 22 of the 44 paired participants. Arm (b) is scored on CGMacros-target transfer restricted to the other 22 participants, and every arm is also scored on that held-out half for a like-for-like comparison. These runs use the same separate same-corpus retrains as Appendix~\ref{app:erasure-fm}, so absolute values differ from Table~\ref{tab:erasure}.

Calibrating on the evaluated participants does not produce the gain. On the held-out 22 participants the half-calibrated generator gives a filtering gain of $+17.16$ against $+16.80$ for the control, which was calibrated on all 44 participants including these 22 (Table~\ref{tab:gen-recal}). The uncalibrated generator keeps most of the gain ($+13.35$); its mean difference from the control is $-3.46$, with per-seed differences of $-0.05$, $-5.54$ and $-4.78$, so a contribution of calibration of a few points is possible but not resolved with three seeds. Filtered transfer over all twelve directions and the near-zero gain on the eight directions not into CGMacros are unchanged across arms. On the raw pre-head blocks no arm shows a filtering gain, consistent with Appendix~\ref{app:erasure-fm}.

\begin{table}[!htbp]\centering\footnotesize\setlength{\tabcolsep}{4pt}
\caption{Synthetic-sensor generator arms used to test calibration leakage. Lag is in five-minute steps; all arms apply no level shift (bias 0).}
\label{tab:gen-recal-arms}
\begin{tabular}{llccl}
\toprule
Arm & Lag & Smoothing $\sigma$ & Bias & Fitted on \\
\midrule
Control (released recipe) & $-1\pm1.0$ & 1.5 & 0 & all 44 CGMacros paired participants \\
(a) Uncalibrated & 0 & 0 & 0 & nothing (only the Libre 15-minute decimation) \\
(b) Half-calibrated & $-1\pm1.48$ & 1.29 & 0 & 22 of the 44 paired participants \\
\bottomrule
\end{tabular}
\end{table}

\begin{table}[!htbp]\centering\footnotesize\setlength{\tabcolsep}{3.5pt}
\caption{Filtering gain (filtered minus full transfer ROC-AUC) under each generator, mean $\pm$ SD over three seeds. Held-out 22 restricts CGMacros-target scoring to the participants not used by arm (b). The last column gives each arm minus the control on the held-out 22, averaged over seeds, with per-seed values in brackets.}
\label{tab:gen-recal}
\resizebox{\linewidth}{!}{%
\begin{tabular}{lccccc}
\toprule
Arm & Filtered ROC, all 12 & Gain, into CGMacros (all 44) & Gain, into CGMacros (held-out 22) & Gain, 8 not into CGMacros & Arm $-$ control (held-out 22) \\
\midrule
\multicolumn{6}{l}{\emph{Projected readout (deployed)}} \\
Control & 73.14 $\pm$ 0.62 & 14.39 $\pm$ 2.44 & 16.80 $\pm$ 4.05 & 0.03 $\pm$ 0.77 & --- \\
(a) Uncalibrated & 73.05 $\pm$ 0.53 & 12.42 $\pm$ 5.42 & 13.35 $\pm$ 6.83 & 0.45 $\pm$ 0.89 & $-3.46$ [$-0.05$, $-5.54$, $-4.78$] \\
(b) Half-calibrated & 73.17 $\pm$ 0.58 & 15.14 $\pm$ 5.30 & 17.16 $\pm$ 7.13 & 0.03 $\pm$ 0.91 & $+0.35$ [$+4.73$, $-0.00$, $-3.67$] \\
\midrule
\multicolumn{6}{l}{\emph{Raw pre-head blocks}} \\
Control & 67.69 $\pm$ 3.12 & $-0.29$ $\pm$ 0.88 & $-0.70$ $\pm$ 0.97 & 0.25 $\pm$ 0.21 & --- \\
(a) Uncalibrated & 67.82 $\pm$ 3.13 & $-0.02$ $\pm$ 4.25 & $-0.33$ $\pm$ 4.54 & 0.36 $\pm$ 0.25 & $+0.37$ [$+2.84$, $-3.94$, $+2.19$] \\
(b) Half-calibrated & 67.55 $\pm$ 3.38 & $-0.59$ $\pm$ 1.01 & $-1.10$ $\pm$ 1.02 & 0.15 $\pm$ 0.37 & $-0.40$ [$-0.17$, $-0.55$, $-0.47$] \\
\bottomrule
\end{tabular}}
\end{table}

\section{Full model table and paired significance}\label{app:percell}


This section reports results across window-level ranking metrics, subject-level
prediction and threshold-based decisions. Full task--cohort values and paired
tests show where model performance and uncertainty differ across these settings.

Table~\ref{tab:master} reports task-averaged scores for every model.
Table~\ref{tab:sigall} gives the paired tests on window ROC-AUC, window PR-AUC
and subject ROC-AUC together, all under the same pre-declared family and the
same Holm correction. 

\textbf{PR-AUC deserves particular attention}, both because it is the metric our
backbone leads with and because prevalence varies widely across these cells. Our
own gain there is $\prdelta$ on $\prwins$ of $\prn$ cells and does not survive
correction ($p_{\mathrm{Holm}} = \prholm$). The zero-shot
comparison shows a larger difference on PR than on ROC. There,
$\prsigcount$ of the seven general-purpose foundation models are
Holm-significantly worse than a 0.72M-parameter CGM-specific encoder under class
imbalance.

\begin{table}[!htbp]\centering\footnotesize\setlength{\tabcolsep}{2.6pt}
\caption{\textbf{Paired significance against GlucoFM, on all three metrics at once.} Wilcoxon signed-rank over the 14 task--cohort cells, Holm-corrected over the confirmatory family $k=\famk$ of declared before any result was read. $\Delta$ is the mean paired difference and ``won'' the number of cells in which the model beats GlucoFM. On the confirmatory axis, window ROC-AUC, exactly one comparison survives correction, and it is ours. The three further starred entries all sit in the PR column and are baselines falling \emph{below} GlucoFM, not improvements over it. Reading across a row rather than across three separate tables is the point: the ordering is stable on window ROC and PR, and dissolves at subject level.}
\label{tab:sigall}

\begin{tabular}{lrrrrrrrrr}
\toprule
& \multicolumn{3}{c}{window ROC-AUC} & \multicolumn{3}{c}{window PR-AUC} & \multicolumn{3}{c}{subject ROC-AUC} \\
\cmidrule(lr){2-4}\cmidrule(lr){5-7}\cmidrule(lr){8-10}
Model & $\Delta$ & won & $p_{\mathrm{Holm}}$ & $\Delta$ & won & $p_{\mathrm{Holm}}$ & $\Delta$ & won & $p_{\mathrm{Holm}}$ \\
\midrule
\textbf{GlucoPRISM} & \textbf{+2.16} & 13/14 & 0.031$^{*}$ & \textbf{+1.16} & 9/14 & 0.208 & +0.50 & 7/14 & 1.000 \\
Chr-2s & -7.38 & 4/14 & 0.107 & -9.21 & 3/14 & 0.019$^{*}$ & -2.85 & 4/14 & 1.000 \\
Chr-2 & -6.24 & 3/14 & 0.221 & -8.08 & 3/14 & 0.047$^{*}$ & -3.21 & 5/14 & 1.000 \\
MOM-l & -5.92 & 3/14 & 0.282 & -7.66 & 3/14 & 0.019$^{*}$ & +0.31 & 8/14 & 1.000 \\
CGM-JEPA & -4.94 & 3/14 & 0.293 & -5.64 & 3/14 & 0.122 & --- & --- & --- \\
X-CGM-JEPA & -5.05 & 3/14 & 0.293 & -5.78 & 3/14 & 0.122 & --- & --- & --- \\
CGMformer & -2.94 & 3/14 & 0.392 & -4.19 & 3/14 & 0.107 & -2.61 & 4/14 & 1.000 \\
Mantis & -2.24 & 5/14 & 0.541 & -3.95 & 3/14 & 0.116 & \textbf{+1.00} & 6/14 & 1.000 \\
MOM-s & -3.39 & 4/14 & 0.541 & -5.86 & 3/14 & 0.141 & -1.69 & 6/14 & 1.000 \\
MantisV2 & -1.10 & 4/14 & 0.541 & -2.63 & 4/14 & 0.203 & +0.26 & 7/14 & 1.000 \\
\bottomrule
\end{tabular}

\vspace{4pt}\par\footnotesize\begin{minipage}{\textwidth}$^{*}$ adjusted $p<0.05$. Our own PR-AUC gain is \prdelta{} on \prwins{} of \prn{} cells and does not survive correction. The PR column is also where the zero-shot comparison is strongest: \prsigcount{} of the seven general-purpose models are Holm-significantly worse than a 0.72M-parameter CGM-specific encoder under class imbalance.\end{minipage}
\end{table}
\begin{table}[!htbp]\centering\footnotesize\setlength{\tabcolsep}{3pt}
\caption{Cross-dataset transfer with seed variability: mean $\pm$ SD over three pretraining seeds, in percentage points, for all directions in Table~\ref{tab:xfer}. \textbf{Bold}: best; \underline{underline}: second best.}
\label{tab:xfer-sd}
\resizebox{\linewidth}{!}{%
\begin{tabular}{lllrrrr>{\columncolor{oursbg}}r}
\toprule
Direction & Task & Metric & CGM-JEPA & X-CGM-JEPA & GluFormer (tiny) & GlucoFM & GlucoPRISM \\
\midrule
Stanford $\to$ Hall & Diabetes & PR & 53.8\,{\scriptsize$\pm$\,0.7} & 55.2\,{\scriptsize$\pm$\,0.5} & \underline{63.2\,{\scriptsize$\pm$\,5.1}} & 59.5\,{\scriptsize$\pm$\,1.6} & \textbf{68.0\,{\scriptsize$\pm$\,2.0}} \\
 &  & ROC & 70.4\,{\scriptsize$\pm$\,0.9} & 71.5\,{\scriptsize$\pm$\,0.4} & \underline{72.0\,{\scriptsize$\pm$\,3.2}} & 68.4\,{\scriptsize$\pm$\,0.3} & \textbf{76.6\,{\scriptsize$\pm$\,0.8}} \\
\addlinespace[1pt]
 & IR & PR & 48.7\,{\scriptsize$\pm$\,2.0} & 48.5\,{\scriptsize$\pm$\,0.7} & 48.7\,{\scriptsize$\pm$\,2.5} & \textbf{53.1\,{\scriptsize$\pm$\,1.9}} & \underline{50.7\,{\scriptsize$\pm$\,2.6}} \\
 &  & ROC & \underline{63.9\,{\scriptsize$\pm$\,1.1}} & \textbf{64.0\,{\scriptsize$\pm$\,0.6}} & 60.7\,{\scriptsize$\pm$\,2.9} & 61.7\,{\scriptsize$\pm$\,1.4} & 62.1\,{\scriptsize$\pm$\,0.5} \\
\midrule
Hall $\to$ Stanford & Diabetes & PR & 66.3\,{\scriptsize$\pm$\,0.1} & 66.3\,{\scriptsize$\pm$\,0.4} & 68.4\,{\scriptsize$\pm$\,1.1} & \underline{69.9\,{\scriptsize$\pm$\,1.2}} & \textbf{78.5\,{\scriptsize$\pm$\,1.1}} \\
 &  & ROC & 63.9\,{\scriptsize$\pm$\,0.2} & 64.0\,{\scriptsize$\pm$\,0.3} & 64.7\,{\scriptsize$\pm$\,1.1} & \underline{66.3\,{\scriptsize$\pm$\,1.1}} & \textbf{76.0\,{\scriptsize$\pm$\,1.1}} \\
\addlinespace[1pt]
 & IR & PR & 51.8\,{\scriptsize$\pm$\,0.2} & 51.7\,{\scriptsize$\pm$\,0.2} & 50.9\,{\scriptsize$\pm$\,0.8} & \textbf{52.9\,{\scriptsize$\pm$\,0.9}} & \underline{52.7\,{\scriptsize$\pm$\,1.0}} \\
 &  & ROC & 56.6\,{\scriptsize$\pm$\,0.3} & 56.7\,{\scriptsize$\pm$\,0.1} & 56.2\,{\scriptsize$\pm$\,0.9} & \textbf{58.2\,{\scriptsize$\pm$\,0.7}} & \underline{58.0\,{\scriptsize$\pm$\,0.7}} \\
\midrule
CGMacros $\to$ Hall & Diabetes & PR & 49.3\,{\scriptsize$\pm$\,1.6} & 50.5\,{\scriptsize$\pm$\,0.6} & 58.9\,{\scriptsize$\pm$\,1.4} & \underline{65.2\,{\scriptsize$\pm$\,0.7}} & \textbf{66.8\,{\scriptsize$\pm$\,2.9}} \\
 &  & ROC & 69.6\,{\scriptsize$\pm$\,1.5} & 70.7\,{\scriptsize$\pm$\,0.2} & 72.1\,{\scriptsize$\pm$\,1.0} & \textbf{78.2\,{\scriptsize$\pm$\,0.9}} & \underline{78.0\,{\scriptsize$\pm$\,1.4}} \\
\addlinespace[1pt]
 & IR & PR & 48.9\,{\scriptsize$\pm$\,0.2} & 49.4\,{\scriptsize$\pm$\,1.1} & 50.2\,{\scriptsize$\pm$\,5.7} & \underline{55.2\,{\scriptsize$\pm$\,2.1}} & \textbf{56.5\,{\scriptsize$\pm$\,1.6}} \\
 &  & ROC & 61.1\,{\scriptsize$\pm$\,0.1} & 61.6\,{\scriptsize$\pm$\,1.0} & 63.2\,{\scriptsize$\pm$\,3.5} & \underline{68.5\,{\scriptsize$\pm$\,0.6}} & \textbf{70.0\,{\scriptsize$\pm$\,0.4}} \\
\midrule
Hall $\to$ CGMacros & Diabetes & PR & 85.6\,{\scriptsize$\pm$\,0.4} & 86.1\,{\scriptsize$\pm$\,0.1} & 84.5\,{\scriptsize$\pm$\,2.9} & \underline{87.9\,{\scriptsize$\pm$\,5.0}} & \textbf{89.0\,{\scriptsize$\pm$\,2.0}} \\
 &  & ROC & 75.2\,{\scriptsize$\pm$\,0.4} & 75.7\,{\scriptsize$\pm$\,0.1} & 72.6\,{\scriptsize$\pm$\,5.1} & \underline{77.2\,{\scriptsize$\pm$\,8.8}} & \textbf{81.7\,{\scriptsize$\pm$\,1.7}} \\
\addlinespace[1pt]
 & IR & PR & 87.4\,{\scriptsize$\pm$\,0.5} & 86.8\,{\scriptsize$\pm$\,0.6} & 85.8\,{\scriptsize$\pm$\,1.7} & \textbf{91.0\,{\scriptsize$\pm$\,0.6}} & \underline{87.9\,{\scriptsize$\pm$\,3.5}} \\
 &  & ROC & 73.2\,{\scriptsize$\pm$\,0.2} & 73.0\,{\scriptsize$\pm$\,0.4} & 71.9\,{\scriptsize$\pm$\,1.6} & \textbf{80.3\,{\scriptsize$\pm$\,0.9}} & \underline{76.6\,{\scriptsize$\pm$\,4.2}} \\
\midrule
CGMacros $\to$ Stanford & Diabetes & PR & 66.2\,{\scriptsize$\pm$\,0.3} & 65.9\,{\scriptsize$\pm$\,0.1} & 72.4\,{\scriptsize$\pm$\,0.8} & \underline{79.6\,{\scriptsize$\pm$\,1.0}} & \textbf{82.9\,{\scriptsize$\pm$\,0.6}} \\
 &  & ROC & 64.6\,{\scriptsize$\pm$\,0.4} & 64.5\,{\scriptsize$\pm$\,0.0} & 69.4\,{\scriptsize$\pm$\,0.8} & \underline{74.3\,{\scriptsize$\pm$\,1.9}} & \textbf{80.9\,{\scriptsize$\pm$\,0.2}} \\
\addlinespace[1pt]
 & IR & PR & 47.6\,{\scriptsize$\pm$\,0.2} & 47.7\,{\scriptsize$\pm$\,0.4} & 56.0\,{\scriptsize$\pm$\,1.2} & \underline{56.7\,{\scriptsize$\pm$\,2.0}} & \textbf{61.0\,{\scriptsize$\pm$\,1.2}} \\
 &  & ROC & 52.3\,{\scriptsize$\pm$\,0.3} & 52.5\,{\scriptsize$\pm$\,0.5} & 61.9\,{\scriptsize$\pm$\,0.9} & \underline{62.6\,{\scriptsize$\pm$\,2.9}} & \textbf{65.8\,{\scriptsize$\pm$\,0.4}} \\
\midrule
Stanford $\to$ CGMacros & Diabetes & PR & \underline{87.2\,{\scriptsize$\pm$\,0.4}} & \underline{87.2\,{\scriptsize$\pm$\,0.2}} & 87.0\,{\scriptsize$\pm$\,1.0} & 84.3\,{\scriptsize$\pm$\,2.2} & \textbf{91.6\,{\scriptsize$\pm$\,0.3}} \\
 &  & ROC & 76.4\,{\scriptsize$\pm$\,0.5} & 76.5\,{\scriptsize$\pm$\,0.3} & \underline{77.0\,{\scriptsize$\pm$\,1.9}} & 71.1\,{\scriptsize$\pm$\,3.1} & \textbf{85.6\,{\scriptsize$\pm$\,0.8}} \\
\addlinespace[1pt]
 & IR & PR & 80.1\,{\scriptsize$\pm$\,1.3} & 80.3\,{\scriptsize$\pm$\,1.0} & \underline{84.5\,{\scriptsize$\pm$\,2.5}} & 84.0\,{\scriptsize$\pm$\,2.9} & \textbf{86.7\,{\scriptsize$\pm$\,1.0}} \\
 &  & ROC & 61.7\,{\scriptsize$\pm$\,0.9} & 62.0\,{\scriptsize$\pm$\,1.3} & \underline{70.0\,{\scriptsize$\pm$\,2.5}} & 65.3\,{\scriptsize$\pm$\,6.9} & \textbf{71.4\,{\scriptsize$\pm$\,2.1}} \\
\midrule
\multicolumn{2}{l}{Average (12 directions)} & PR & 64.4 & 64.6 & 67.5 & \underline{69.9} & \textbf{72.7} \\
 &  & ROC & 65.7 & 66.1 & 67.6 & \underline{69.3} & \textbf{73.6} \\
\bottomrule
\end{tabular}}
\end{table}

\section{Extended public, device and HPP comparisons}\label{detail:sec:headline}\label{detail:sec:ladder}\label{detail:sec:erasure}\label{detail:sec:why}\label{detail:sec:fails}\label{detail:sec:hpp}
The tables and figures below report the public benchmark, transfer readouts, device diagnostics and HPP evaluation at their original granularity. Each comparison retains its model identity and aggregation protocol.

The public window benchmark supports the training recipe, and the same-encoder comparison of full and filtered readouts supports the deployment filter. Fitted INLP reaches a similar transfer point estimate under different fitting requirements. Paired-device evaluation gives a different result: the primary model averages 1.34 AUROC points below GlucoFM and wins three of eight comparisons. These tests distinguish cross-cohort transfer from device-pair generalization; within the same encoder, filtering still improves paired-device transfer (Table~\ref{tab:crossdevice}).

HPP classification uses locally trained probes on frozen public-corpus encoders. Appendices~\ref{app:downstream-regression} and~\ref{app:block-functions} provide the complete primary-model clinical and dynamic readout comparisons, including the benefit of Sensor retention within GlucoPRISM and GlucoFM's higher dynamic-task scores.

\begin{table}[!htbp]\centering\footnotesize\setlength{\tabcolsep}{2.6pt}
\caption{\textbf{Cross-cohort transfer: all twelve (source, target, task) directions.} The probe is fitted on the \emph{entire} source cohort and evaluated on the \emph{entire} target cohort; target clinical labels are used only for scoring, not fitting or tuning and there is no cross-validation, so this measures direct transfer rather than within-cohort generalisation. ROC-AUC, higher is better; \textbf{bold} is the best in a column, \underline{underline} the runner-up. The second row is the same model read \emph{without} discarding the Sensor block, so the first row minus the second isolates the intervention this paper proposes. Cohorts abbreviated CGM (CGMacros), Stan (Stanford), Hall; tasks DR (diabetes risk) and IR (insulin resistance).}
\label{tab:transfer}

\begin{tabular}{lrrrrrrrrrrrr >{\columncolor{oursbg}}r}
\toprule
& \multicolumn{2}{c}{CGM$\to$Hall} & \multicolumn{2}{c}{CGM$\to$Stan} & \multicolumn{2}{c}{Hall$\to$CGM} & \multicolumn{2}{c}{Hall$\to$Stan} & \multicolumn{2}{c}{Stan$\to$CGM} & \multicolumn{2}{c}{Stan$\to$Hall} & \\
\cmidrule(lr){2-3}\cmidrule(lr){4-5}\cmidrule(lr){6-7}\cmidrule(lr){8-9}\cmidrule(lr){10-11}\cmidrule(lr){12-13}
Model & DR & IR & DR & IR & DR & IR & DR & IR & DR & IR & DR & IR & \textbf{Avg} \\
\midrule
\rowcolor{oursbg}\textbf{GlucoPRISM} & \underline{78.0} & \underline{70.0} & \textbf{80.9} & \underline{65.8} & \textbf{81.7} & \underline{76.6} & \textbf{76.0} & 58.0 & \textbf{85.6} & 71.4 & \textbf{76.6} & 62.1 & \textbf{73.6} \\
\addlinespace[2pt]
\quad $z_A$ \emph{kept} & \underline{78.0} & \textbf{70.7} & \underline{80.8} & 63.9 & 67.6 & 57.8 & \underline{75.7} & \underline{59.2} & 74.3 & 59.2 & \underline{75.9} & 61.9 & 68.7 \\
\addlinespace[2pt]\multicolumn{14}{l}{\emph{Backbone and frozen third-party baselines}} \\
GlucoFM & \textbf{78.2} & 68.5 & 74.3 & 62.6 & 77.2 & \textbf{80.3} & 66.3 & 58.2 & 71.1 & 65.3 & 68.4 & 61.7 & \underline{69.3} \\
MantisV2 & 68.3 & 64.0 & 69.5 & \textbf{71.0} & \underline{77.6} & 64.8 & 65.8 & \textbf{65.0} & 68.3 & \underline{72.5} & 66.1 & \textbf{68.8} & 68.5 \\
CGMformer & 66.6 & 59.2 & 68.0 & 63.3 & 75.2 & 68.2 & 64.1 & 55.2 & \underline{78.0} & \textbf{74.4} & 69.6 & \underline{63.3} & 67.1 \\
MOMENT-l & 66.4 & 55.8 & 67.7 & 63.2 & 69.8 & 58.6 & 67.7 & \underline{59.2} & 67.7 & 62.0 & 64.6 & 61.1 & 63.7 \\
Chronos-2 & 58.1 & 57.3 & 62.5 & 63.9 & 65.3 & 58.2 & 60.9 & 57.5 & 68.8 & 66.5 & 61.8 & 62.0 & 61.9 \\
\bottomrule
\end{tabular}

\vspace{4pt}\par\footnotesize\begin{minipage}{\textwidth}Transfer PR-AUC depends on the target cohort's class balance. This table reports ROC-AUC; per-direction PR-AUC is retained in the released tables. The separate corpus-size figure reports macro-average AP over a fixed set of directions, which supports comparisons across models and corpus sizes within that protocol rather than comparisons between target prevalences. This protocol supplies only twelve non-independent directions, and no comparison on this axis survives Holm correction in either direction, including a baseline 7.4 ROC-AUC \emph{behind} GlucoFM.
\end{minipage}
\end{table}

\begin{table}[!htbp]\centering\footnotesize\setlength{\tabcolsep}{2.6pt}
\caption{Paired cross-device evaluation over all \cdn{} task--direction combinations. GlucoPRISM has a mean difference of \cdmean{} AUROC points versus GlucoFM and wins \cdwins{} comparisons. The same encoder's full readout is weaker for Dexcom $\rightarrow$ Libre (mean \cddexlibfull{}) than in the reverse direction (\cdlibdexfull{}). This asymmetry motivates further analysis of sampling, calibration and participant effects; it does not isolate a unique mechanism.}
\label{tab:crossdevice}

\begin{tabular}{lrrrrr}
\toprule
Endpoint & GlucoFM & GlucoPRISM & $\Delta$ vs FM & GlucoPRISM [$z_A$ kept] & deletion gain \\
\midrule
\addlinespace\multicolumn{6}{l}{\emph{Dexcom $\rightarrow$ Libre}} \\
Diabetes risk (3-cls) & \textbf{81.85} & \underline{80.21} & $-1.63$ & 61.09 & $+19.12$ \\
Hyperlipidemia & \textbf{63.91} & \underline{54.07} & $-9.84$ & 53.00 & $+1.07$ \\
Insulin resistance & \underline{83.27} & \textbf{85.25} & $+1.97$ & 54.09 & $+31.16$ \\
Obesity & \textbf{66.78} & \underline{64.66} & $-2.12$ & 60.41 & $+4.25$ \\
\addlinespace\multicolumn{6}{l}{\emph{Libre $\rightarrow$ Dexcom}} \\
Diabetes risk (3-cls) & 75.33 & \underline{78.82} & $+3.49$ & \textbf{79.28} & $-0.45$ \\
Hyperlipidemia & \textbf{65.95} & \underline{61.26} & $-4.68$ & 56.54 & $+4.72$ \\
Insulin resistance & 85.33 & \textbf{87.86} & $+2.53$ & \underline{85.81} & $+2.05$ \\
Obesity & \textbf{79.19} & \underline{78.77} & $-0.42$ & 74.92 & $+3.85$ \\
\bottomrule
\end{tabular}

\vspace{4pt}\par\footnotesize\begin{minipage}{\textwidth}Deletion improves the full readout by \cddexlibgain{} AUROC points for Dexcom $\rightarrow$ Libre and \cdlibdexgain{} in the reverse direction. These are comparisons within GlucoPRISM, distinct from its comparison with GlucoFM. All \cdn{} task--direction combinations are retained; the directional averages do not establish general device invariance.\end{minipage}
\end{table}
\section{Block controls at matched width}\label{app:controls}
\begin{figure}[!htbp]\centering
\includegraphics[width=0.73\linewidth]{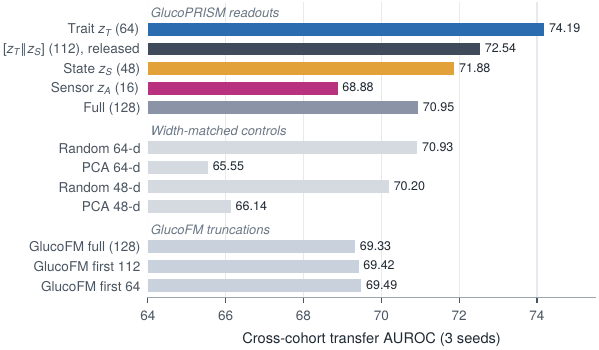}
\caption{\textbf{Individual block readouts and width-matched controls.} Cross-cohort
transfer ROC-AUC over three seeds. Trait and State each outperform their
controls. Trait alone and the released $[\zt\|\zs]$ readout differ by less than
the seed standard deviation. Applying the same truncations to the GlucoFM
embedding yields no corresponding gain. These values come from separate block-control runs on different checkpoints, not from the released model's runs in Table~\ref{tab:erasure}, so the two are not expected to match.}
\label{fig:readouts}
\end{figure}

A narrow probe input is better regularised than a wide one at 29--69 subjects,
so a small block can score well for reasons that have nothing to do with what it
encodes. Every claim about a named block therefore has to clear a projection of
the same width. Figure~\ref{fig:readouts} supplies that floor from both sides. It
scores each block alone against random and PCA projections of matched width, and
against the same truncation applied to the baseline embedding.

Both blocks outperform their matched-width controls: $\zt$ by $\roztrand$
and $\roztpca$ \auc, and $\zs$ by $\rozsrand$ and $\rozspca$. Each block
therefore contains useful predictive information. Trait alone reaches $\rozt$
against $\rozts$ for the released $[\zt\|\zs]$ readout. The gap of $\roztgap$
favors Trait on all three seeds but is not statistically resolved across seeds
($t=1.88$). This comparison does not establish an additional transfer benefit
from combining State with Trait. Within the prism view this ordering is expected: Trait is the band defined to be unchanged across both days and sensors, so it should transfer best on its own.

Sensor alone reaches $\roza$ transfer AUROC, indicating accessible predictive information despite its small width. This information may reflect clinical patterns, device--population associations or shared features. Its predictiveness does not establish a pure physiological or device factor. The comparison motivates evaluating deletion directly rather than inferring its value from a block's name.

\section{Erasure, calibration and partial deletion}\label{app:erasure}
\label{app:partial}
Table~\ref{tab:erasure} in the main paper reports the complete erasure and calibration comparison.
INLP improves the full readout by $\inlpgain$ AUROC points, compared with $\trdropc$ for the fixed filter. The paired deletion--INLP difference is $\inlpdelta$ over $\inlpn$ directions, with 95\% interval $[\inlpdeltalo,\inlpdeltahi]$ ($t=\inlpt$, $p=\inlpp$), positive in $\inlpwins$ directions. These results do not establish superiority or equivalence. The procedures use different information: INLP fits eight device probes and accesses target-domain inputs when CGMacros is the target, without using target clinical-outcome labels; the fixed filter uses a band specified during training. The comparison supports useful fitted erasure in this embedding, rather than uniquely identifying a nuisance mechanism.

Calibration matters separately from ranking. A screening threshold has to
transfer, and \auc{} says nothing about whether it will. Dropping $\za$ improves
expected calibration error from $\eceful$ to $\ecedrop$ and Brier score from
$\brierfull$ to $\brierdrop$, with INLP between the two on both. The ordering is
identical on all three metrics. That is reassuring given how modest our PR-AUC
and Macro-F1 gains are: the intervention is not buying ranking at the cost of
calibration. The last alternative is to delete only part of the block, which
Table~\ref{tab:partial} tests directly.
\begin{table}[!htbp]\centering\footnotesize\setlength{\tabcolsep}{2.0pt}
\caption{Partial Sensor deletion. Coordinates are ranked by device predictability using real Dexcom/Libre labels, and the least informative $m$ are restored. Removing all 16 coordinates gives the highest transfer and window AUROC among these readouts; the full readout has the highest subject AUROC. Restoring four coordinates reduces transfer AUROC from 73.57 to 68.49. ECE and Brier are lower-is-better.}
\label{tab:partial}

\begin{tabular}{lrrrrr}
\toprule
& \multicolumn{3}{c}{Cross-cohort transfer} & \multicolumn{2}{c}{Within-cohort} \\
\cmidrule(lr){2-4}\cmidrule(lr){5-6}
Readout & ROC-AUC & ECE & Brier & window & subject \\
\midrule
keep all 16 (full readout) & 68.75\,{\scriptsize$\pm$}{\scriptsize 2.67} & 0.163\,{\scriptsize$\pm$}{\scriptsize 0.022} & 0.241\,{\scriptsize$\pm$}{\scriptsize 0.014} & 67.90\,{\scriptsize$\pm$}{\scriptsize 0.66} & \textbf{69.55}\,{\scriptsize$\pm$}{\scriptsize 2.48} \\
keep 12 of 16 & 68.76\,{\scriptsize$\pm$}{\scriptsize 2.45} & 0.163\,{\scriptsize$\pm$}{\scriptsize 0.026} & 0.242\,{\scriptsize$\pm$}{\scriptsize 0.017} & --- & --- \\
keep 8 of 16 & 68.44\,{\scriptsize$\pm$}{\scriptsize 2.09} & 0.167\,{\scriptsize$\pm$}{\scriptsize 0.027} & 0.244\,{\scriptsize$\pm$}{\scriptsize 0.019} & --- & --- \\
keep 4 of 16 & 68.49\,{\scriptsize$\pm$}{\scriptsize 2.12} & 0.169\,{\scriptsize$\pm$}{\scriptsize 0.025} & 0.245\,{\scriptsize$\pm$}{\scriptsize 0.018} & 67.75\,{\scriptsize$\pm$}{\scriptsize 0.55} & 69.15\,{\scriptsize$\pm$}{\scriptsize 3.14} \\
keep 2 of 16 & \underline{71.63}\,{\scriptsize$\pm$}{\scriptsize 2.38} & \underline{0.137}\,{\scriptsize$\pm$}{\scriptsize 0.013} & \underline{0.218}\,{\scriptsize$\pm$}{\scriptsize 0.006} & \underline{67.97}\,{\scriptsize$\pm$}{\scriptsize 0.62} & 68.71\,{\scriptsize$\pm$}{\scriptsize 3.25} \\
\textbf{keep 0 of 16 (ours)} & \textbf{73.57}\,{\scriptsize$\pm$}{\scriptsize 0.54} & \textbf{0.114}\,{\scriptsize$\pm$}{\scriptsize 0.009} & \textbf{0.207}\,{\scriptsize$\pm$}{\scriptsize 0.006} & \textbf{68.01}\,{\scriptsize$\pm$}{\scriptsize 0.15} & \underline{69.45}\,{\scriptsize$\pm$}{\scriptsize 1.18} \\
\bottomrule
\end{tabular}

\vspace{4pt}\par\footnotesize\begin{minipage}{\textwidth}Within-cohort scores were run for the three readouts the paper released or ablated; $m=8$ and $m=12$ were run on transfer only, where the cliff is.\end{minipage}
\end{table}
Column standardization makes uniform nonzero rescaling of Sensor ineffective in this probe: the reported scores agree to three decimals. Coordinate removal is a distinct intervention. Restoring two coordinates reduces transfer AUROC to 71.63, and restoring four reduces it to 68.49. Calibration worsens alongside transfer ranking, but this association does not identify a unique mechanism. The subject-level maximum occurs with all coordinates retained, reinforcing the task dependence of deletion.

\subsection{Fitted erasure on GlucoFM}\label{app:erasure-fm}
This study asks whether fitting an eraser to the baseline embedding can match the fixed Sensor filter. It uses a separate set of same-corpus retrains with three pretraining seeds per encoder (GlucoPRISM seeds 0--2 and GlucoFM seeds 0--2), so absolute scores differ from Table~\ref{tab:erasure}; GlucoFM in particular transfers about 2 points lower than in Table~\ref{tab:xfer} (67.3 against 69.3 AUROC).
 The erasers follow the INLP arm of Table~\ref{tab:erasure} exactly: a standardizer and the eraser are fitted on all CGMacros windows from both sensors with Dexcom/Libre labels and applied to every cohort, so for the four CGMacros-target directions the eraser sees target inputs but never target clinical labels. INLP uses eight iterations; LEACE is the closed-form least-squares eraser~\citep{belrose2023leace}. Transfer uses the twelve directions of Table~\ref{tab:transfer} with one source-fitted logistic probe and window-level target scoring. Differences are paired by seed and direction, with 95\% intervals from a bootstrap that resamples seeds and directions jointly (10{,}000 replicates).

On the projected readout that the paper deploys, the fixed filter and fitted erasure on the same encoder are indistinguishable: INLP differs by $+0.33$ AUROC $[-0.83,+1.76]$ and LEACE by $-0.05$ $[-1.27,+1.18]$, although the filter uses no target data and no device labels. Fitted erasure on GlucoFM does not close the gap. GlucoFM with LEACE remains $3.94$ points below the fixed filter $[-6.77,-0.51]$ and GlucoFM with INLP $4.45$ points below $[-7.44,-1.29]$, with higher calibration error and Brier score (Table~\ref{tab:erasure-fm-delta}). The gap to erased GlucoFM is clearest on the eight directions not into CGMacros; on the four CGMacros-target directions the intervals include zero.

On the raw pre-head blocks, removing the Sensor block does not change transfer ($-0.07$ $[-0.63,+0.69]$), and the raw GlucoPRISM readout is close to GlucoFM. The removable measurement band is therefore formed by the block projections $h_T$, $h_S$ and $h_A$ of Equation~\ref{eq:blocks}, which define the deployed readout; the raw blocks are reported for completeness (Table~\ref{tab:erasure-fm-abs}, lower panel).

\begin{table}[!htbp]\centering\footnotesize\setlength{\tabcolsep}{4pt}
\caption{Fitted erasure versus the fixed Sensor filter: transfer scores averaged over twelve directions, mean $\pm$ SD over three pretraining seeds. Separate same-corpus retrains from Table~\ref{tab:erasure}. The upper panel uses the projected readout deployed in the paper; the lower panel uses the raw pre-head blocks. GlucoFM rows are identical in both panels because GlucoFM has no block heads.}
\label{tab:erasure-fm-abs}
\begin{tabular}{lcccc}
\toprule
Method & ROC-AUC & PR-AUC & ECE & Brier \\
\midrule
\multicolumn{5}{l}{\emph{Projected readout}} \\
GlucoPRISM, filtered (Sensor removed) & \underline{73.14} $\pm$ 0.62 & 72.33 $\pm$ 1.32 & \underline{0.128} $\pm$ 0.013 & \underline{0.214} $\pm$ 0.008 \\
GlucoPRISM, full readout & 68.35 $\pm$ 0.55 & 69.40 $\pm$ 1.62 & 0.167 $\pm$ 0.030 & 0.243 $\pm$ 0.022 \\
GlucoPRISM, full + INLP (8 it.) & \textbf{73.46} $\pm$ 0.88 & \textbf{72.46} $\pm$ 1.22 & \textbf{0.123} $\pm$ 0.007 & \textbf{0.213} $\pm$ 0.004 \\
GlucoPRISM, full + LEACE & 73.09 $\pm$ 1.01 & \underline{72.35} $\pm$ 1.24 & 0.136 $\pm$ 0.010 & 0.219 $\pm$ 0.010 \\
GlucoFM & 67.26 $\pm$ 2.53 & 68.27 $\pm$ 1.32 & 0.237 $\pm$ 0.026 & 0.283 $\pm$ 0.023 \\
GlucoFM + INLP (8 it.) & 68.69 $\pm$ 0.40 & 68.86 $\pm$ 0.27 & 0.207 $\pm$ 0.020 & 0.261 $\pm$ 0.010 \\
GlucoFM + LEACE & 69.20 $\pm$ 1.03 & 68.94 $\pm$ 0.91 & 0.209 $\pm$ 0.011 & 0.261 $\pm$ 0.007 \\
\midrule
\multicolumn{5}{l}{\emph{Raw pre-head blocks}} \\
GlucoPRISM, filtered (Sensor removed) & 67.69 $\pm$ 3.13 & \underline{69.17} $\pm$ 2.52 & \underline{0.212} $\pm$ 0.020 & \textbf{0.262} $\pm$ 0.018 \\
GlucoPRISM, full readout & 67.62 $\pm$ 2.77 & 69.05 $\pm$ 2.50 & 0.221 $\pm$ 0.015 & 0.270 $\pm$ 0.012 \\
GlucoPRISM, full + INLP (8 it.) & \underline{68.30} $\pm$ 3.09 & 69.11 $\pm$ 2.56 & \textbf{0.209} $\pm$ 0.015 & \textbf{0.262} $\pm$ 0.012 \\
GlucoPRISM, full + LEACE & \textbf{69.04} $\pm$ 3.06 & \textbf{69.83} $\pm$ 2.47 & \underline{0.212} $\pm$ 0.009 & \underline{0.263} $\pm$ 0.010 \\
\bottomrule
\end{tabular}
\end{table}

\begin{table}[!htbp]\centering\footnotesize\setlength{\tabcolsep}{3pt}
\caption{Each method minus the fixed filter (GlucoPRISM, filtered) on the projected readout, by direction group. Paired by seed and direction; brackets are 95\% intervals from a joint seed-and-direction bootstrap. Negative $\Delta$ROC and $\Delta$PR, and positive $\Delta$ECE and $\Delta$Brier, favour the fixed filter. The last column repeats $\Delta$ROC for the raw pre-head readout.}
\label{tab:erasure-fm-delta}
\resizebox{\linewidth}{!}{%
\begin{tabular}{llccccc}
\toprule
Method & Directions & $\Delta$ROC-AUC & $\Delta$PR-AUC & $\Delta$ECE & $\Delta$Brier & $\Delta$ROC-AUC (raw) \\
\midrule
GlucoPRISM, full readout & All 12 & $-4.78$ $[-9.21,-1.04]$ & $-2.93$ $[-6.24,-0.28]$ & $+0.039$ $[-0.004,+0.101]$ & $+0.030$ $[-0.000,+0.079]$ & $-0.07$ $[-0.63,+0.69]$ \\
 & 8 not into CGMacros & $-0.03$ $[-0.98,+1.14]$ & $+0.24$ $[-0.78,+1.55]$ & $+0.008$ $[-0.010,+0.035]$ & $+0.002$ $[-0.006,+0.013]$ & $-0.25$ $[-0.70,+0.25]$ \\
 & 4 into CGMacros & $-14.29$ $[-19.43,-9.77]$ & $-9.27$ $[-14.55,-4.60]$ & $+0.101$ $[-0.002,+0.225]$ & $+0.084$ $[+0.009,+0.182]$ & $+0.29$ $[-1.19,+1.95]$ \\
\addlinespace
GlucoPRISM, full + INLP & All 12 & $+0.33$ $[-0.83,+1.76]$ & $+0.13$ $[-1.03,+1.44]$ & $-0.006$ $[-0.039,+0.030]$ & $-0.001$ $[-0.017,+0.016]$ & $+0.61$ $[-0.98,+2.79]$ \\
 & 8 not into CGMacros & $-0.32$ $[-1.06,+0.48]$ & $-0.19$ $[-1.57,+1.17]$ & $+0.004$ $[-0.013,+0.029]$ & $+0.001$ $[-0.005,+0.011]$ & $-0.38$ $[-0.94,+0.17]$ \\
 & 4 into CGMacros & $+1.62$ $[-1.34,+4.53]$ & $+0.76$ $[-1.60,+3.16]$ & $-0.025$ $[-0.103,+0.062]$ & $-0.006$ $[-0.046,+0.041]$ & $+2.59$ $[-2.10,+7.64]$ \\
\addlinespace
GlucoPRISM, full + LEACE & All 12 & $-0.05$ $[-1.27,+1.18]$ & $+0.02$ $[-0.96,+1.28]$ & $+0.008$ $[-0.026,+0.045]$ & $+0.006$ $[-0.013,+0.026]$ & $+1.35$ $[-0.29,+3.40]$ \\
 & 8 not into CGMacros & $-0.29$ $[-1.18,+0.81]$ & $-0.23$ $[-1.23,+1.13]$ & $+0.012$ $[-0.003,+0.038]$ & $+0.004$ $[-0.004,+0.015]$ & $-0.36$ $[-0.97,+0.24]$ \\
 & 4 into CGMacros & $+0.44$ $[-2.12,+2.68]$ & $+0.53$ $[-0.88,+2.45]$ & $-0.000$ $[-0.083,+0.098]$ & $+0.009$ $[-0.033,+0.064]$ & $+4.76$ $[+0.81,+8.45]$ \\
\addlinespace
GlucoFM & All 12 & $-5.88$ $[-9.46,-2.26]$ & $-4.06$ $[-6.47,-1.75]$ & $+0.109$ $[+0.055,+0.168]$ & $+0.069$ $[+0.034,+0.107]$ & $-0.43$ $[-4.90,+5.99]$ \\
 & 8 not into CGMacros & $-4.69$ $[-6.99,-2.37]$ & $-4.22$ $[-7.66,-1.27]$ & $+0.125$ $[+0.064,+0.197]$ & $+0.066$ $[+0.037,+0.105]$ & $-3.22$ $[-5.91,-0.72]$ \\
 & 4 into CGMacros & $-8.26$ $[-15.55,+1.25]$ & $-3.74$ $[-6.79,+0.58]$ & $+0.077$ $[-0.008,+0.155]$ & $+0.074$ $[-0.008,+0.155]$ & $+5.14$ $[-4.79,+17.24]$ \\
\addlinespace
GlucoFM + INLP & All 12 & $-4.45$ $[-7.44,-1.29]$ & $-3.47$ $[-5.99,-1.21]$ & $+0.079$ $[+0.018,+0.141]$ & $+0.047$ $[+0.014,+0.079]$ & $+1.00$ $[-3.34,+6.92]$ \\
 & 8 not into CGMacros & $-4.47$ $[-6.97,-2.10]$ & $-3.87$ $[-7.05,-1.12]$ & $+0.113$ $[+0.056,+0.187]$ & $+0.059$ $[+0.032,+0.097]$ & $-3.01$ $[-6.07,-0.32]$ \\
 & 4 into CGMacros & $-4.40$ $[-12.70,+3.79]$ & $-2.67$ $[-7.51,+1.18]$ & $+0.011$ $[-0.107,+0.111]$ & $+0.023$ $[-0.055,+0.084]$ & $+9.01$ $[-0.25,+19.21]$ \\
\addlinespace
GlucoFM + LEACE & All 12 & $-3.94$ $[-6.77,-0.51]$ & $-3.39$ $[-5.99,-0.98]$ & $+0.081$ $[+0.017,+0.144]$ & $+0.047$ $[+0.011,+0.080]$ & $+1.51$ $[-3.00,+8.02]$ \\
 & 8 not into CGMacros & $-4.40$ $[-6.82,-2.05]$ & $-4.09$ $[-7.57,-1.10]$ & $+0.122$ $[+0.062,+0.192]$ & $+0.065$ $[+0.036,+0.102]$ & $-2.93$ $[-5.82,-0.33]$ \\
 & 4 into CGMacros & $-3.01$ $[-10.44,+5.45]$ & $-1.99$ $[-6.23,+2.23]$ & $-0.000$ $[-0.108,+0.080]$ & $+0.011$ $[-0.058,+0.062]$ & $+10.39$ $[+2.85,+21.20]$ \\
\bottomrule
\end{tabular}}
\end{table}

\subsection{Domain-adversarial GlucoFM and device information after filtering}\label{app:dann}
This study compares the band-and-filter design with standard invariance learning. GlucoFM with its native 128-dimensional readout was pretrained with a gradient-reversal device head~\citep{ganin2015dann} (128 to 128 to 5 units) on the unmasked pooled representation, using the same corpus, 300 epochs, batch size 128, sampler and recipe. Each anchor is paired with GlucoPRISM's synthetic sensor partner, and device targets follow GlucoPRISM (the anchor's real device family; partners are labelled Libre). The reversal coefficient ramps as $\lambda(2/(1+e^{-10p})-1)$ over training progress $p$. The weight $\lambda\in\{0.1,0.3,1,3\}$ was selected on the pretraining corpus without its 51 held-out subjects by a rule fixed in advance: the lowest held-out device-probe AUROC among weights whose held-out MCR+TD loss stays within 10\% of that at $\lambda=0.1$ (Table~\ref{tab:dann-select}). Only $\lambda=0.1$ met the loss budget. Final models use three seeds on the full corpus; the other weights are reported as a sensitivity analysis. All rows are re-evaluated in one job with identical code on the separate same-corpus retrains of Appendix~\ref{app:erasure-fm}, in which GlucoFM reaches 62.99 window AUROC rather than the 65.85 of Table~\ref{tab:master}.

The adversary does not improve transfer. The selected model reaches 67.19 transfer AUROC against 67.25 for GlucoFM and 73.14 for the filtered GlucoPRISM readout, and stronger weights lower it to between 61.73 and 65.17 (Table~\ref{tab:dann}). The adversary also lowers window-level scores and does not reach the calibration of the fixed filter. Weights of 0.3 and above raise the held-out reconstruction loss roughly seven- to tenfold,
 so in this setting the adversary disrupts pretraining rather than removing the device signal.

A linear device probe on same-day CGMacros pairs (Dexcom against Libre, participant-grouped five-fold logistic regression on all 838 real windows; 50 means no linear device information) decodes the device at 99.9 to 100 AUROC from every representation, including the filtered GlucoPRISM readout (99.97). Filtering therefore does not remove linearly decodable device identity. Sampling rate differs between the two sensors and is visible to any encoder (Appendix~\ref{app:dependence}), so the transfer gain from filtering is better described as removing the device-related component that the source-fitted clinical probe relies on than as removing all measurement information.

\begin{table}[!htbp]\centering\footnotesize\setlength{\tabcolsep}{5pt}
\caption{Selection of the adversarial weight on the pretraining held-out split, with no downstream data. The rule keeps weights whose held-out MCR+TD loss is within 10\% of that at $\lambda=0.1$ and picks the lowest held-out device AUROC among them.}
\label{tab:dann-select}
\begin{tabular}{cccc}
\toprule
$\lambda$ & Held-out MCR+TD loss & Held-out device AUROC & Within loss budget \\
\midrule
0.1 & 0.0300 & 99.91 & yes (selected) \\
0.3 & 0.2372 & 98.99 & no \\
1 & 0.2061 & 99.53 & no \\
3 & 0.3003 & 97.67 & no \\
\bottomrule
\end{tabular}
\end{table}

\begin{table}[!htbp]\centering\footnotesize\setlength{\tabcolsep}{2.5pt}
\caption{Domain-adversarial GlucoFM (DANN) against GlucoFM and GlucoPRISM, mean $\pm$ SD over three seeds, from the separate same-corpus retrains. Upper block: within-cohort 14-cell averages and 12-direction transfer ROC-AUC. Lower block: transfer PR-AUC and calibration, and linear device information on same-day CGMacros pairs. \textbf{Bold}: best; \underline{underline}: second best (device AUROC is not ranked).}
\label{tab:dann}
\resizebox{\linewidth}{!}{%
\begin{tabular}{lccccccc}
\toprule
Model & Window PR & Window ROC & Window F1 & Subject PR & Subject ROC & Subject F1 & Transfer ROC \\
\midrule
GlucoPRISM, filtered & \textbf{59.06} $\pm$ 0.49 & \textbf{67.90} $\pm$ 0.36 & \textbf{57.87} $\pm$ 0.11 & \underline{67.07} $\pm$ 0.29 & 69.26 $\pm$ 1.15 & 58.78 $\pm$ 0.28 & \textbf{73.14} $\pm$ 0.62 \\
GlucoPRISM, full & \underline{58.87} $\pm$ 0.45 & \underline{67.88} $\pm$ 0.41 & \underline{57.81} $\pm$ 0.36 & \underline{67.07} $\pm$ 2.20 & \textbf{70.06} $\pm$ 2.61 & 60.09 $\pm$ 1.69 & \underline{68.33} $\pm$ 0.53 \\
GlucoFM & 56.04 $\pm$ 0.30 & 62.99 $\pm$ 0.22 & 56.47 $\pm$ 0.26 & 66.17 $\pm$ 0.87 & 68.98 $\pm$ 0.75 & \underline{60.18} $\pm$ 0.36 & 67.25 $\pm$ 2.54 \\
DANN-GlucoFM, $\lambda=0.1$ (selected) & 54.30 $\pm$ 0.56 & 61.85 $\pm$ 0.74 & 55.68 $\pm$ 0.17 & \textbf{67.26} $\pm$ 3.02 & \underline{69.48} $\pm$ 4.42 & \textbf{60.50} $\pm$ 2.56 & 67.19 $\pm$ 1.77 \\
DANN-GlucoFM, $\lambda=0.3$ & 54.97 $\pm$ 1.25 & 63.71 $\pm$ 0.69 & 56.58 $\pm$ 0.81 & 64.73 $\pm$ 2.95 & 67.00 $\pm$ 3.82 & 58.19 $\pm$ 2.77 & 61.73 $\pm$ 3.21 \\
DANN-GlucoFM, $\lambda=1$ & 53.18 $\pm$ 1.60 & 61.43 $\pm$ 1.65 & 55.03 $\pm$ 1.04 & 63.84 $\pm$ 3.23 & 65.68 $\pm$ 3.68 & 56.56 $\pm$ 3.26 & 64.02 $\pm$ 3.93 \\
DANN-GlucoFM, $\lambda=3$ & 54.40 $\pm$ 1.40 & 63.38 $\pm$ 1.54 & 55.90 $\pm$ 1.01 & 63.29 $\pm$ 1.54 & 65.76 $\pm$ 1.50 & 57.25 $\pm$ 1.73 & 65.17 $\pm$ 0.89 \\
\midrule
& Transfer PR & ECE & Brier & \multicolumn{3}{c}{Device probe AUROC (CGMacros real pairs)} & \\
\midrule
GlucoPRISM, filtered & \textbf{72.34} $\pm$ 1.32 & \textbf{0.128} $\pm$ 0.013 & \textbf{0.214} $\pm$ 0.008 & \multicolumn{3}{c}{99.97 $\pm$ 0.03} & \\
GlucoPRISM, full & \underline{69.38} $\pm$ 1.59 & \underline{0.167} $\pm$ 0.031 & \underline{0.243} $\pm$ 0.022 & \multicolumn{3}{c}{100.00 $\pm$ 0.00} & \\
GlucoFM & 68.26 $\pm$ 1.33 & 0.238 $\pm$ 0.026 & 0.283 $\pm$ 0.023 & \multicolumn{3}{c}{100.00 $\pm$ 0.00} & \\
DANN-GlucoFM, $\lambda=0.1$ (selected) & 67.51 $\pm$ 0.60 & 0.217 $\pm$ 0.027 & 0.267 $\pm$ 0.020 & \multicolumn{3}{c}{100.00 $\pm$ 0.00} & \\
DANN-GlucoFM, $\lambda=0.3$ & 63.00 $\pm$ 2.05 & 0.242 $\pm$ 0.031 & 0.296 $\pm$ 0.026 & \multicolumn{3}{c}{99.92 $\pm$ 0.14} & \\
DANN-GlucoFM, $\lambda=1$ & 65.14 $\pm$ 2.74 & 0.253 $\pm$ 0.028 & 0.298 $\pm$ 0.028 & \multicolumn{3}{c}{100.00 $\pm$ 0.00} & \\
DANN-GlucoFM, $\lambda=3$ & 66.02 $\pm$ 1.15 & 0.225 $\pm$ 0.022 & 0.284 $\pm$ 0.012 & \multicolumn{3}{c}{100.00 $\pm$ 0.00} & \\
\bottomrule
\end{tabular}}
\end{table}

\section{Sensor-block capacity and KL weight}\label{app:capacity}\label{sec:why}
Two knobs govern how much the Sensor block can hold: its width $d_A$, and the
price $\beta$ charged per nat. We sweep both against the same frozen folds, two
seeds per arm, and read both the same way. What matters is not the arm's own
score but how much \emph{deleting} the block is worth in it.

The KL sweep reports deletion gains of $+0.60$, $+0.51$, $+1.68$ and $+1.57$ at $\beta=0.03$, $0.1$, $0.3$ and $1.0$, respectively (Figure~\ref{fig:deletion}b). The higher-weight settings have larger gains, but the sequence is not monotonic. With two seeds per arm and seed standard deviation near $1.0$, these observations do not distinguish channel specialization from other regularization effects. The released setting is $\beta=0.1$: full-readout AUROC is $67.52$, compared with $66.42$ at $\beta=1.0$, a decrease of $1.10$ points. Thus larger deletion gains need to be considered together with the predictive information lost from the full representation.

The deletion gains at Sensor widths 8, 16 and 32 are 0.35, 0.51 and 0.30 points, respectively. With two seeds per arm and overlapping variability, this sweep does not establish an optimal width or a systematic width effect. Trait remains fixed at 64 dimensions while State and Sensor share the remaining dimensions, so this is a comparison of allocations within a fixed total width.
\begin{table}[!htbp]\centering\footnotesize\setlength{\tabcolsep}{3.4pt}
\caption{Sensor width and KL-weight sweep with two matched seeds per arm. Width changes hold Trait at 64 dimensions and allocate the remaining width between State and Sensor. Larger deletion gains occur at the two higher KL weights, but the sequence is nonmonotonic and uncertain.}
\label{tab:capacity}

\begin{tabular}{lrrrr}
\toprule
Setting & Full readout & $z_A$ dropped & Gain (window) & Gain (subject) \\
\midrule
\multicolumn{5}{l}{\emph{Sensor-block width $d_A$ (at $\beta = 0.1$)}} \\
\quad 8 & \textbf{67.67}\,{\scriptsize$\pm$}{\scriptsize 0.43} & \underline{68.01}\,{\scriptsize$\pm$}{\scriptsize 0.06} & \underline{$+0.35$}\,{\scriptsize$\pm$}{\scriptsize 0.37} & \underline{$-1.11$}\,{\scriptsize$\pm$}{\scriptsize 1.89} \\
\quad 16\,\textbf{(released)} & 67.52\,{\scriptsize$\pm$}{\scriptsize 0.17} & \textbf{68.03}\,{\scriptsize$\pm$}{\scriptsize 0.21} & $\mathbf{+0.51}$\,{\scriptsize$\pm$}{\scriptsize 0.04} & $\mathbf{+0.56}$\,{\scriptsize$\pm$}{\scriptsize 1.02} \\
\quad 32 & \underline{67.62}\,{\scriptsize$\pm$}{\scriptsize 0.59} & 67.91\,{\scriptsize$\pm$}{\scriptsize 0.29} & $+0.30$\,{\scriptsize$\pm$}{\scriptsize 0.30} & $-1.49$\,{\scriptsize$\pm$}{\scriptsize 0.00} \\
\addlinespace
\multicolumn{5}{l}{\emph{KL weight $\beta$ (at $d_A = 16$)}} \\
\quad 0.03 & \textbf{67.56}\,{\scriptsize$\pm$}{\scriptsize 1.17} & \textbf{68.16}\,{\scriptsize$\pm$}{\scriptsize 0.74} & $+0.60$\,{\scriptsize$\pm$}{\scriptsize 0.43} & $-0.03$\,{\scriptsize$\pm$}{\scriptsize 3.21} \\
\quad 0.1\,\textbf{(released)} & \underline{67.52}\,{\scriptsize$\pm$}{\scriptsize 0.17} & \underline{68.03}\,{\scriptsize$\pm$}{\scriptsize 0.21} & $+0.51$\,{\scriptsize$\pm$}{\scriptsize 0.04} & \underline{$+0.56$}\,{\scriptsize$\pm$}{\scriptsize 1.02} \\
\quad 0.3 & 65.88\,{\scriptsize$\pm$}{\scriptsize 0.52} & 67.56\,{\scriptsize$\pm$}{\scriptsize 0.35} & $\mathbf{+1.68}$\,{\scriptsize$\pm$}{\scriptsize 0.17} & $\mathbf{+1.59}$\,{\scriptsize$\pm$}{\scriptsize 4.20} \\
\quad 1 & 66.42\,{\scriptsize$\pm$}{\scriptsize 0.25} & 67.99\,{\scriptsize$\pm$}{\scriptsize 0.59} & \underline{$+1.57$}\,{\scriptsize$\pm$}{\scriptsize 0.84} & $+0.34$\,{\scriptsize$\pm$}{\scriptsize 1.77} \\
\addlinespace
\bottomrule
\end{tabular}

\vspace{4pt}\par\footnotesize\begin{minipage}{\textwidth}At $\beta=1.0$, full and retained-readout AUROC are 66.42 and 67.99, compared with 67.52 and 68.03 at the released $\beta=0.1$. The larger deletion gain at $\beta=1.0$ therefore reflects a greater difference between readouts, not higher absolute performance. This post-release sweep uses two seeds per arm and was not used to reselect the released model. Subject-level gains vary substantially across settings. These results describe sensitivity to the bottleneck; they do not establish a monotonic dose response or a preferred replacement configuration.\end{minipage}
\end{table}

\section{Post-hoc factorization of a frozen encoder}\label{app:posthoc}
We test whether a comparable decomposition can be fitted after pretraining.
The encoder remains frozen while the decomposition is fitted using held-out
real paired windows, with the number of paired participants varied. This
comparison evaluates an alternative to learning the block structure jointly
with the encoder.

The prespecified Trait-versus-unsplit comparison averages $\posthocdelta$ AUROC over $\posthocn$ configurations ($t=\posthoct$), with positive differences in $\posthocpos$ configurations. The reported sweep shows no consistent improvement with more paired subjects. This result concerns the tested post-hoc objective and fitting procedure; it does not rule out other ways to learn a removable subspace. The fit uses real paired recordings and device labels. GlucoPRISM pretraining instead combines native device labels with synthetic sensor partners.

\begin{table}[!htbp]\centering\footnotesize\setlength{\tabcolsep}{5pt}
\caption{Fitting the block decomposition post hoc on a \emph{frozen} encoder, using real paired windows rather than synthetic ones, while sweeping how many paired subjects are available for the fit. The pre-specified comparison is the Trait block against the frozen encoder's own unsplit output: it is $-0.08$ ROC-AUC over 12 configurations ($t=-0.32$), positive in 7 of 12. The tested post-hoc procedure yields no resolved average improvement.}
\label{tab:posthoc}

\begin{tabular}{lrrrrr}
\toprule
Encoder & Paired subj. & Frozen encoder & $z_T$ & $z_T\|z_S$ & $\Delta$ \\
\midrule
$1\times$, no factorization loss & 15 & 65.5 & 64.7 & 64.6 & $-0.80$ \\
 & 20 & 65.5 & 66.0 & 65.7 & $+0.50$ \\
 & 30 & 65.5 & 66.8 & 65.7 & $+1.30$ \\
 & 45 & 65.5 & 65.6 & 65.2 & $+0.10$ \\
\addlinespace
$1\times$, post-hoc heads & 15 & 63.5 & 64.0 & 63.4 & $+0.50$ \\
 & 20 & 63.5 & 64.1 & 64.4 & $+0.60$ \\
 & 30 & 63.5 & 62.0 & 63.9 & $-1.50$ \\
 & 45 & 63.5 & 64.2 & 65.0 & $+0.70$ \\
\addlinespace
$4.97\times$, post-hoc heads & 15 & 64.0 & 63.4 & 62.9 & $-0.60$ \\
 & 20 & 64.0 & 63.0 & 63.0 & $-1.00$ \\
 & 30 & 64.0 & 62.7 & 61.6 & $-1.30$ \\
 & 45 & 64.0 & 64.5 & 63.7 & $+0.50$ \\
\addlinespace
\midrule\multicolumn{5}{l}{\textbf{Mean over all 12 configurations}} & $\mathbf{-0.08}$ \\
\bottomrule
\end{tabular}

\vspace{4pt}\par\footnotesize\begin{minipage}{\textwidth}Read instead as ``did \emph{any} of the five blocks win'', the same table returns 7 of 12, but the winning block is chosen after seeing the result and is never the same block twice, so that statistic is selection bias, not evidence.\end{minipage}
\end{table}

\section{Sensitivity to stronger auxiliary-objective weights}\label{app:arch}\label{sec:weight}
The architecture grid crosses model capacity with how the factorization is
obtained, three seeds per cell. The six cells read as follows. With the
factorization off, the backbone scores $65.85$ at $1\times$ and $65.17$ at
$4.97\times$, so capacity buys nothing except run-to-run stability (seed s.d.\
$1.08 \to 0.44$). With the objectives trained jointly at full weight the score
falls to $60.57$ at $1\times$ and $57.94$ at $4.97\times$. Fitted post hoc on a
frozen encoder it is $65.53$ and $65.78$, which is to say the same as not doing
it at all.

The grid shows that high-weight objectives reduce accuracy, with a larger reduction at greater model capacity. This is consistent with competing training objectives, but does not identify directional gradient conflict: gradient directions were not measured. Gradient surgery is therefore a proposed diagnostic and intervention rather than an established remedy. These results concern the tested objectives and weights, not the general viability of factorization.
\begin{figure}[!htbp]\centering
\includegraphics[width=0.74\linewidth]{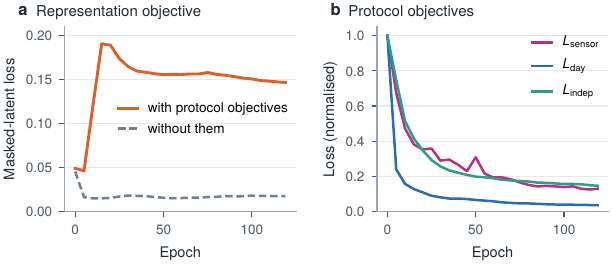}
\caption{Training losses with and without strong protocol objectives. The masked-latent objective increases during joint training while auxiliary losses decrease. This indicates competing optimization outcomes, without directly measuring gradient conflict.}
\label{fig:loss}
\end{figure}
At full objective weight, joint training costs $5.28$ \auc{} at $1\times$ and
$7.23$ at $4.97\times$ ($t=-5.0$, $-6.5$). Over the same run the masked-latent
objective rises from $0.049$ to $0.146$, against $0.018$ for an unfactorized
control. The paired-sensor augmentation alone, with the protocol objectives disabled,
is neutral. Corpus, backbone and augmentation distribution are matched across
the two arms. The updated comparison uses measured augmentation parameters and
two-layer projection heads. As documented in Appendix~\ref{app:impl}, the
objective switches also affect auxiliary regularization, so these results
compare training configurations rather than isolated losses.

\section{Controlled objective contributions}
\label{app:controlled-positive}

\subsection{Study design and uncertainty}
The controlled contribution study evaluates the primary GlucoPRISM configuration across three paired pretraining seeds, 0, 1 and 2. Its public experiment matrix contains 36 model--seed runs covering 12 configurations. Each run uses the same 10,952 pretraining windows, 300 epochs and 25,800 updates. The matched GlucoPRISM ablations preserve architecture, instantiated heads, shared initialization, schedules, views and stochastic computations, changing the selected final objective coefficient or coefficients. The sensor-bundle contrast removes sensor-view alignment and device classification together. Each objective effect is conditional on the remaining training recipe.

Full and retained readouts comprise the projected 128-dimensional representation and its exact first 112 coordinates, respectively. Each receives a separately fitted probe under the same prescribed procedure. The public evaluation spans four cohorts, with 14 within-cohort task cells at window and subject levels and 24 transfer task/direction/level cells. Transfer means give equal weight to these 24 cells. The evaluation manifest excludes ten windows belonging to the known Shanghai participant overlapping pretraining and preserves surviving participant fold assignments. These results follow their own protocol and are reported separately from the main benchmark in Table~\ref{tab:master}.

Intervals below are conditional 95\% Bayesian participant-bootstrap intervals from 1,000 draws, with shared participant weights across windows, sessions, folds, tasks, readouts and seeds. They are descriptive and unadjusted, conditional on the fixed fitted encoders, probes, splits, source samples and observed cohorts. Sample SD across pretraining seeds is reported separately. AUROC differences are calculated before rounding displayed means.

\subsection{Conditional contributions of the objectives}
Table~\ref{tab:controlled-positive-objectives} reports the objective contrasts whose intervals exclude zero; the remaining contrasts are uncertain. Sensor-view alignment improves full and retained transfer by 5.82 and 6.70 AUROC points, respectively, with positive effects in all three paired seeds. Removing both sensor objectives reduces retained transfer by 4.79 points, establishing a contribution from sensor supervision to the retained representation within this recipe. Variance regularization improves window discrimination by 1.35 points and retained transfer by 3.44 points. Sensor KL contributes 0.62 points to window discrimination when Gaussian heads, posterior sampling and evaluation-time posterior-mean export remain active.

\begin{table}[!htbp]
\centering\small
\caption{Objective contributions whose intervals exclude zero in the controlled study. All entries are AUROC percentage points. The complete configuration is compared with the indicated objective removed, with the other coefficients fixed. Intervals are conditional 95\% Bayesian participant-bootstrap intervals.}
\label{tab:controlled-positive-objectives}
\setlength{\tabcolsep}{3.5pt}
\begin{tabular}{@{}llrrrr@{}}
\toprule
Objective & Endpoint & Without & Complete & Gain & Interval\\
\midrule
Sensor-view alignment & Full transfer & 62.52 & 68.34 & +5.82 & [3.60, 7.82]\\
Sensor-view alignment & Retained transfer & 65.09 & 71.79 & +6.70 & [4.14, 9.10]\\
Sensor-objective bundle & Retained transfer & 67.00 & 71.79 & +4.79 & [2.26, 7.24]\\
Variance regularization & Full window & 66.89 & 68.23 & +1.35 & [0.44, 2.52]\\
Variance regularization & Retained transfer & 68.35 & 71.79 & +3.44 & [1.13, 5.86]\\
Sensor KL regularization & Full window & 67.61 & 68.23 & +0.62 & [0.05, 1.14]\\
\bottomrule
\end{tabular}
\end{table}

A separate strict multinomial participant bootstrap also gives positive alignment intervals of [3.71, 7.95] points for full transfer and [4.12, 9.14] for retained transfer. The variance objective's retained-transfer interval is [0.90, 6.08] points under that analysis. These comparisons support the specified transfer effects under both resampling procedures.

\subsection{Utility of the filtered readout}
The filtered representation increases transfer AUROC from $68.34\pm1.42$ to $71.79\pm0.21$ on the prespecified 24-cell panel. The paired deletion gain is $3.45\pm1.26$ points, with Bayesian interval [1.89, 4.94] and strict multinomial interval [1.75, 4.92]. As in the main benchmark, the gains are concentrated in the directions that target CGMacros, the cohort recorded with two sensor types. Table~\ref{tab:controlled-positive-readout} reports accompanying improvements in precision--recall performance, classification F1 and probabilistic prediction metrics.

\begin{table}[!htbp]
\centering\small
\caption{Favorable mean changes after removing the 16 Sensor coordinates in the controlled study's transfer panel. Changes use each metric's original scale. The accompanying intervals in the text apply to AUROC; the other entries are descriptive mean changes.}
\label{tab:controlled-positive-readout}
\begin{tabular}{@{}lrr@{}}
\toprule
Metric & Retained minus full & Favorable direction\\
\midrule
AUROC & +0.0345 & Higher\\
Average precision & +0.0221 & Higher\\
Macro-F1 & +0.0806 & Higher\\
Brier score & $-0.0717$ & Lower\\
Expected calibration error, ten bins & $-0.0968$ & Lower\\
\bottomrule
\end{tabular}
\end{table}

\subsection{Variance regularization and representation geometry}
Variance regularization increases singular-value entropy effective rank in both projected readouts (Table~\ref{tab:controlled-positive-geometry}). Each value averages 12 cohort--seed records from four cohorts and three seeds. The effective-rank increase occurs in every matched comparison for both readouts. Total variance also increases in all 12 comparisons for the full, retained and backbone representations. Effective rank here is computed from normalized singular values. Together with the downstream comparisons, these measurements connect the variance objective to increased representation spread and improved window discrimination and retained transfer.

\begin{table}[!htbp]
\centering\small
\caption{Mean singular-value entropy effective rank with and without variance regularization. Paired consistency counts the cohort--seed records with higher effective rank when the objective is active.}
\label{tab:controlled-positive-geometry}
\begin{tabular}{@{}lrrr@{}}
\toprule
Readout & Without variance & Complete & Paired consistency\\
\midrule
Full projected & 4.708 & 6.519 & 12/12\\
Retained projected & 3.310 & 5.210 & 12/12\\
\bottomrule
\end{tabular}
\end{table}

\section{The HPP external cohort: construction and audit}\label{app:hpp}

Section~\ref{sec:hpp} rests on a cohort we could not release and did not
assemble ourselves, so this appendix states how each number was produced, at the
level of method rather than of schema.

\paragraph{Access, and what leaves the enclave.} HPP is owned by Pheno.AI,
governed by a data use agreement, and reachable only inside its trusted research
environment; access is arranged with Pheno.AI directly and is not ours to
grant~\citep{shilo2021hpp}. No participant identifier, timestamp or glucose
value left that environment. What left is a small table of model $\times$
endpoint $\times$ metric aggregates, from which Tables~\ref{tab:hpp} and
\ref{tab:hppfull} and every HPP number in this paper are generated.

Because the dataset is not ours to describe, we give the construction below in
terms of the decisions we made and the evidence for them, and we name neither
the field identifiers nor the file layout of the source. The cohort-specific
loaders stay inside the enclave for the same reason. What we do release is the
evaluation driver itself, meaning the frozen-encoder embedding, per-subject
mean-embedding aggregation and linear probe for HPP, as a cohort-agnostic
script that consumes a window shard and a label table in the format
Section~\ref{sec:eval} already documents. Any group with HPP access can
reproduce Table~\ref{tab:hpp} with it; a group without HPP access can run the
identical protocol on a cohort of their own.

\paragraph{Cohort construction.} We read \num{9497} sensor connections, of which
\num{9303} participants yielded at least one usable window. The following preprocessing choices define the evaluation cohort.

\emph{Timestamps are local.} The source timestamps are timezone-aware. We drop
the zone keeping local wall time. Converting to UTC first would shift every
reading by the cohort's offset and silently misalign the circadian encoding of
Section~\ref{sec:backbone}; the model would be learning a shifted day while
believing it had learned a day.

\emph{Binning is \texttt{nearest}.} Inter-reading gaps are exactly $15.00$
minutes for \SI{100}{\percent} of readings, but the phase varies per connection:
one sampled wear sat entirely off the $:00/:15/:30/:45$ grid. The clock does not
drift, it starts elsewhere, so flooring would quantise a whole wear into the
wrong slots. This is the same setting ShanghaiT2DM uses for a different reason.

\emph{Device comes from the manifest, not the filename.} The cohort holds two
variants of one manufacturer's professional 15-minute sensor, variant~A (\num{5905} subjects; \num{104582} of the \num{160308} overlapping pretraining windows) and variant~B (\num{3450} subjects; \num{55726} windows),
 which we join to each wear
through the connection manifest rather than infer. That split is the
transfer column of Table~\ref{tab:hppfull}. It is between-subject: only \num{52}
participants hold
both, and those are the same participants who returned for a second visit, so
hardware and time are confounded. We report it as a weaker test than CGMacros'
same-day paired windows and use it only as a secondary device analysis.

Evaluation windows are cut exactly as for the other evaluation cohorts:
 24 hours, non-overlapping, aligned to the 288-slot
5-minute grid with an observation mask, segments split at gaps over an hour, and a coverage floor of
half of what the device can achieve. A 15-minute sensor fills at most 96 of 288 slots, so the floor is
relative to $96$ and not to $288$; an absolute threshold would discard every HPP
window. The resulting shard holds \num{\hppwins} windows at a mean observation
fraction of $0.333$, which is to say complete. For the corpus-size study (Figure~\ref{fig:hpp_scaling}), the same recordings are cut with 40\% overlap between consecutive days, as for the public pretraining corpus, which gives \num{160308} windows from the \num{9303} participants, of which the largest subset of 8,000 participants uses \num{137402}. HPP has essentially no missingness,
which is unusual and worth knowing when comparing its scores to cohorts that do.

\paragraph{Labels.} Every label is a laboratory assay or a physical measurement,
at the thresholds of Section~\ref{sec:eval}, so that an HPP positive means what
a Stanford positive means:

\begin{itemize}\itemsep1pt \parskip0pt
\item \textbf{Diabetes risk}: the \emph{assayed blood} HbA1c from the
      laboratory panel, thresholded at $5.7\%$. Not any CGM-derived estimate;
      see the audit below.
\item \textbf{Hyperlipidemia}: total cholesterol, LDL or triglycerides at or
      above $240/160/200$~\si{\milli\gram\per\deci\litre}. The panel reports
      lipids in \si{\milli\gram\per\deci\litre} already, so unlike ShanghaiT2DM
      no unit conversion is applied.
\item \textbf{Obesity}: measured BMI $\geq 30$~\si{\kilo\gram\per\metre\squared}.
\item \textbf{Insulin resistance}: \emph{unavailable}. The laboratory panel
      contains neither insulin nor C-peptide, so HOMA-IR cannot be formed. We
      report three endpoints rather than substitute a proxy for the one endpoint
      common to all four of our other cohorts.
\end{itemize}

HPP panels accrue over years, so a participant can hold several draws. We
collapse to the visit \emph{closest in time to that participant's CGM wear}
rather than to the most recent or the most complete one: a label measured years
from the recording is not a label for that recording.

\paragraph{How near is nearest.} That rule is worth characterising rather than
merely stating, because a laboratory value drawn far from the wear is a label
for a different metabolic state. The two sources differ sharply. BMI is measured
at the CGM appointment itself: median gap \num{\hppgapan} days, with
\hppgapanwnthirty\% of participants inside 30 days. The blood panel is a
separate visit: median \num{\hppgapbt} days, interquartile range
$[\hppgapbtlo, \hppgapbthi]$, 90th percentile \num{\hppgapbtpninezero}, and a
tail reaching several years: \hppgapbtwninety\% of participants are matched
inside 90 days and \hppgapbtwnthirty\% inside 30.

Because HbA1c integrates glycemia over the preceding two to three months, labels matched far from the wear are noisier for every model.

\paragraph{The circularity audit.} The cohort also distributes CGM-\emph{derived}
glycemic estimates, an eA1c and a variability index, alongside the
laboratory panel, and a curated diabetes category that may combine the two
sources. All three are, or may be, functions of the CGM trace: eA1c is
$(46.7 + \overline{g})/28.7$, so a model scored against it would be scored
against an affine transform of its own input. We use none of them. The label is
the assay.

Against the assayed label,
CGM-derived eA1c reaches \hppmeanauc{} \auc{} at Pearson $r=\hppearr$
($n=\num{\hppdr}$). Mean glucose and GMI return identical figures, which is a
consistency check rather than a coincidence: both are strictly increasing
functions of mean glucose, so all three induce the same ranking and $r$ is
invariant to a positive affine map. Mean glucose therefore explains roughly a
seventh of the variance in assayed HbA1c on this cohort. The label is not a
relabelling of the input, and \hppmeanauc{} is the floor every row of
Table~\ref{tab:hpp} must clear.

\paragraph{Probing.} This is the released part of the pipeline. Encoders are
frozen; nothing is fine-tuned. Each model receives the input its own paper
specifies: GlucoFM and GlucoPRISM take the masked window with its mask and
circadian index, CGM-JEPA and GluFormer a linearly interpolated dense series.
Window embeddings are averaged per subject and probed with the logistic
regression of Section~\ref{sec:eval} under $10\times5$ subject-grouped
cross-validation. The \emph{classical indices} row is the cohort's own
distributed table of \texttt{iglu} glycemic indices~\citep{broll2021iglu},
aggregated per participant across connections and median-imputed; \emph{mean
glucose} is the single mean-glucose column of that table.

\begingroup\footnotesize\setlength{\tabcolsep}{2.0pt}
\begin{longtable}{llrrrrrrrrr}
\caption{\textbf{HPP per cell: dispersion, PR-AUC and the cross-device split.} Every row of Table~\ref{tab:hpp}, with the two metrics that table omits. ROC-AUC and PR-AUC are mean $\pm$ standard deviation over $10\times5$ subject-grouped folds. \emph{Transfer} fits on the cohort's first sensor variant and scores on the second, dropping the \num{52} subjects who hold both (\num{\hppsingledev} remain); it is one fit on a fixed split, so it carries no fold dispersion. \emph{d} is the probe input width, which matters for reading the block rows: a narrow input is better regularised at any fixed $n$, so $\za$ at 16 dimensions is not comparable to $\zt$ at 64 without the width-matched controls of Appendix~\ref{app:controls}.}\label{tab:hppfull}\\
\toprule
Representation & $d$ & \multicolumn{3}{c}{ROC-AUC} & \multicolumn{3}{c}{PR-AUC} & \multicolumn{3}{c}{Transfer ROC-AUC} \\
\cmidrule(lr){3-5}\cmidrule(lr){6-8}\cmidrule(lr){9-11}
 & & DR & HL & OB & DR & HL & OB & DR & HL & OB \\
\midrule\endfirsthead
\multicolumn{11}{l}{\emph{\footnotesize Table~\ref{tab:hppfull}, continued}}\\
\toprule
Representation & $d$ & \multicolumn{3}{c}{ROC-AUC} & \multicolumn{3}{c}{PR-AUC} & \multicolumn{3}{c}{Transfer ROC-AUC} \\
\cmidrule(lr){3-5}\cmidrule(lr){6-8}\cmidrule(lr){9-11}
 & & DR & HL & OB & DR & HL & OB & DR & HL & OB \\
\midrule\endhead
\bottomrule\endlastfoot
\texttt{cgm\_mean} & 1 & 65.4\,\tiny$\pm$1.9 & 53.1\,\tiny$\pm$1.6 & 57.0\,\tiny$\pm$1.7 & 41.6\,\tiny$\pm$2.6 & 22.0\,\tiny$\pm$1.8 & 21.6\,\tiny$\pm$1.6 & 65.1 & 55.1 & 57.1 \\
\texttt{iglu} &  & 70.3\,\tiny$\pm$1.8 & 59.7\,\tiny$\pm$1.5 & 69.7\,\tiny$\pm$1.5 & 49.5\,\tiny$\pm$3.1 & 26.4\,\tiny$\pm$1.8 & 33.1\,\tiny$\pm$2.6 & 67.6 & \textbf{60.8} & 67.8 \\
\texttt{iglu\_daymatched} & 39 & 70.5\,\tiny$\pm$1.8 & 58.2\,\tiny$\pm$1.5 & 68.8\,\tiny$\pm$1.4 & -- & -- & -- & -- & -- & -- \\
\texttt{raw} & 288 & 63.6\,\tiny$\pm$1.9 & 53.0\,\tiny$\pm$1.8 & 65.1\,\tiny$\pm$1.6 & 40.5\,\tiny$\pm$2.4 & 21.5\,\tiny$\pm$1.7 & 27.0\,\tiny$\pm$2.0 & 64.3 & 53.3 & 64.8 \\
\texttt{cgm\_jepa} & 96 & 67.3\,\tiny$\pm$1.8 & 55.8\,\tiny$\pm$1.8 & 66.5\,\tiny$\pm$1.6 & 44.7\,\tiny$\pm$2.7 & 23.9\,\tiny$\pm$1.8 & 28.0\,\tiny$\pm$2.0 & 66.5 & 55.1 & 66.2 \\
\texttt{x\_cgm\_jepa} & 96 & 67.5\,\tiny$\pm$1.7 & 56.7\,\tiny$\pm$1.9 & 66.7\,\tiny$\pm$1.6 & 45.0\,\tiny$\pm$2.7 & 24.3\,\tiny$\pm$1.9 & 28.2\,\tiny$\pm$2.0 & 66.8 & 56.1 & 66.5 \\
\texttt{gluformer} & 128 & 68.6\,\tiny$\pm$2.0 & 58.0\,\tiny$\pm$1.7 & 68.1\,\tiny$\pm$1.3 & 46.6\,\tiny$\pm$2.9 & 24.7\,\tiny$\pm$2.1 & 30.6\,\tiny$\pm$2.4 & 65.0 & 55.4 & 64.5 \\
\texttt{glucofm} & 128 & 69.5\,\tiny$\pm$1.9 & \underline{59.8}\,\tiny$\pm$1.5 & 68.7\,\tiny$\pm$1.4 & 47.1\,\tiny$\pm$2.7 & 26.1\,\tiny$\pm$1.9 & 30.7\,\tiny$\pm$2.1 & 66.7 & 59.0 & 68.2 \\
\texttt{glucofm\_\_trunc112} & 112 & 69.5\,\tiny$\pm$1.9 & \textbf{59.9}\,\tiny$\pm$1.5 & 68.7\,\tiny$\pm$1.4 & 47.1\,\tiny$\pm$2.6 & 26.2\,\tiny$\pm$1.9 & 30.7\,\tiny$\pm$2.1 & 66.7 & 59.2 & 68.4 \\
\texttt{glucoprism\_\_full} & 128 & \underline{71.7}\,\tiny$\pm$2.0 & 59.7\,\tiny$\pm$1.6 & \textbf{71.3}\,\tiny$\pm$1.4 & \underline{51.2}\,\tiny$\pm$3.1 & 26.7\,\tiny$\pm$2.1 & \underline{33.7}\,\tiny$\pm$2.6 & \textbf{68.8} & 57.9 & \underline{70.5} \\
\texttt{glucoprism\_\_zTzS} & 112 & \textbf{71.8}\,\tiny$\pm$2.0 & \textbf{59.9}\,\tiny$\pm$1.7 & \underline{71.2}\,\tiny$\pm$1.4 & \textbf{51.3}\,\tiny$\pm$3.1 & \textbf{26.9}\,\tiny$\pm$2.2 & \textbf{33.9}\,\tiny$\pm$2.6 & \underline{68.5} & 58.3 & \textbf{71.6} \\
\texttt{glucoprism\_\_zT} & 64 & 71.5\,\tiny$\pm$2.0 & 59.7\,\tiny$\pm$1.7 & 69.4\,\tiny$\pm$1.4 & 50.7\,\tiny$\pm$3.1 & \underline{26.8}\,\tiny$\pm$2.0 & 32.0\,\tiny$\pm$2.5 & 68.2 & \underline{59.4} & 69.8 \\
\texttt{glucoprism\_\_zS} & 48 & 71.4\,\tiny$\pm$2.0 & 59.5\,\tiny$\pm$1.7 & 69.3\,\tiny$\pm$1.3 & 50.8\,\tiny$\pm$3.1 & 26.3\,\tiny$\pm$2.1 & 32.3\,\tiny$\pm$2.5 & 68.1 & 58.0 & 69.7 \\
\texttt{glucoprism\_\_zA} & 16 & 68.8\,\tiny$\pm$2.1 & 58.1\,\tiny$\pm$1.9 & 65.8\,\tiny$\pm$1.6 & 47.3\,\tiny$\pm$2.9 & 24.9\,\tiny$\pm$1.7 & 28.4\,\tiny$\pm$2.3 & 67.8 & 57.9 & 63.3 \\
\end{longtable}
\vspace{-6pt}\par\footnotesize\begin{minipage}{\textwidth}The transfer columns are a between-subject shift as well as a hardware one, since the two variants are held by disjoint participants, and both sample at 15 minutes, so this is a weaker test than CGMacros' same-day paired windows. It is nonetheless the only cross-device comparison available at this scale, at $206\times$ the subject count. Rows carry the representation identifiers the evaluation driver emits rather than display names, so a group with HPP access can check the table line by line against their own run of it.\end{minipage}
\endgroup

\begin{table}[!htbp]\centering\footnotesize\setlength{\tabcolsep}{2.0pt}
\caption{\textbf{Paired participant-clustered bootstrap on HPP.} \num{\hppreps} replicates; participants are resampled with replacement and every window and prediction from one participant travels together, because windows within a participant are not independent. Both arms of a comparison are scored on the \emph{same} resample, so these are intervals on the difference rather than on either score. Point estimates pool out-of-fold predictions and score once, where Table~\ref{tab:hpp} scores each fold and averages; the two differ by a few tenths because \auc{} is not linear in the pooled sample. $^{*}$ marks $p<0.05$ uncorrected. HPP stays descriptive and outside the family of \famk{} corrected in Section~\ref{sec:headline}.}
\label{tab:hppci}

\begin{tabular}{p{0.19\linewidth}lrrr}
\toprule
Comparison & Metric & DR & HL & OB \\
\midrule
GlucoPRISM $-$ GlucoFM & \auc & $+2.24$ [+1.36, +3.13]\,$^{*}$ & $-0.00$ [-1.20, +1.24] & $+2.51$ [+1.37, +3.66]\,$^{*}$ \\
GlucoPRISM $-$ GlucoFM & PR-AUC & $+4.36$ [+3.10, +5.53]\,$^{*}$ & $+0.69$ [-0.43, +1.78] & $+3.20$ [+1.65, +4.76]\,$^{*}$ \\
GlucoPRISM: $\za$ deleted $-$ kept & \auc & $+0.05$ [-0.11, +0.20] & $+0.18$ [-0.18, +0.55] & $-0.06$ [-0.41, +0.25] \\
GlucoPRISM: $\za$ deleted $-$ kept & PR-AUC & $+0.12$ [-0.12, +0.34] & $+0.12$ [-0.21, +0.45] & $+0.24$ [-0.18, +0.67] \\
indices (day-matched) $-$ GlucoFM & \auc & $+1.06$ [-0.04, +2.19] & $-1.57$ [-3.07, -0.18]\,$^{*}$ & $+0.07$ [-1.40, +1.60] \\
indices (day-matched) $-$ GlucoFM & PR-AUC & $+2.85$ [+1.32, +4.42]\,$^{*}$ & $-1.17$ [-2.41, +0.02] & $+1.28$ [-0.82, +3.21] \\
\bottomrule
\end{tabular}

\vspace{4pt}\par\footnotesize\begin{minipage}{\textwidth}The filtering rows compare the same encoder with and without the Sensor band. None of the six intervals excludes zero, so filtering neither helps nor hurts HPP classification, as expected for a cohort recorded with one sensor type.\end{minipage}
\end{table}
\section{Extended HPP downstream regression evaluation}
\label{app:downstream-regression}
\subsection{Design and predictor sets}
This evaluation uses the original released encoders, frozen throughout, rather than the HPP-pretrained models in the corpus-size sweep. GlucoFM uses its complete embedding and GlucoPRISM uses the fixed Trait--State concatenation (the primary configuration); no endpoint-specific model or block is selected. These tables compare these two prespecified models.

Native HPP CGM has 96 observations per daily window at 15-minute spacing. The original input configurations are retained: the adapter places these observations in every third position of a 288-slot grid, retains the observation mask, and fills using input context only. Evaluation augmentation is disabled. Baseline clinical features are the coordinate-wise mean over a participant's eligible baseline-window embeddings; this differs from the richer summary pooling used in the public-cohort benchmark. Event-level tasks encode their own preceding context.

The downstream pools contain 8,000 development participants, 1,219 common-test participants, and a separate 61-person repeat-visit manifest. Label availability determines the smaller endpoint-specific counts in the tables. Of the repeat-visit group, 55 have eligible baseline CGM and 44 have eligible paired CGM. Counts overlap across tasks and must not be summed. Test people are excluded from probe fitting and hyperparameter selection, and HPP is not part of the public pretraining corpus.

For each target, a Ridge probe with an intercept is fitted after median imputation and standardization. All preprocessing is fitted within the development folds. The penalty is selected from $\{0.1,1,10,100,1000\}$ by participant-weighted held-out mean squared error over three deterministic participant-grouped folds, followed by refitting on all eligible development people. Fold allocation uses SHA-256 ordering with seed 20260918. Tasks require at least 50 development and 10 test participants. For repeated examples, each example from person $i$ receives weight $1/n_i$, giving every person equal total weight. Individual predictions, inner-fold scores and fitted coefficients are not part of the aggregate capture.

The embedding-only and adjusted predictor sets are reported separately. Adjusted probes concatenate the same embedding with available baseline age, sex indicators and recorded device group. Future-level and change probes additionally include the baseline outcome when available by the prediction cutoff. Meal models additionally include logged energy, carbohydrate, protein and fat. No medication adjustment is made. These comparisons do not substitute for a covariate-only or persistence baseline.

\subsection{Concurrent clinical endpoints}
Clinical labels are independent visit measurements selected nearest the baseline visit anchor within 90 days. The anchor falls back to the first CGM-window date if necessary. Labels can occur after the CGM cutoff within that visit window, so this is concurrent phenotype estimation, not prospective forecasting. HbA1c, BMI and systolic blood pressure were the original primary endpoints; the other 12 extend the exploratory phenotype coverage. Blood glucose is not confirmed to be fasting. Body-fat, liver and sleep field units must not be inferred from their names; the tables preserve source units.

Table~\ref{tab:ds-baseline} reports every baseline endpoint with embedding-only probes. The embedding-only GlucoPRISM readout exceeds GlucoFM on correlation and $R^2$ for 14/15 endpoints and MAE for 15/15. Sleep correlation is 0.1407 versus 0.1424 for GlucoFM. With common covariates, it exceeds GlucoFM on all three metrics for 14/15, with sleep the exception. Absolute explained variance remains modest for several endpoints despite relative gains. Pearson $r$ describes linear agreement; $R^2=1-\mathrm{SSE}/\mathrm{SST}$ measures squared-error fit, and negative values indicate worse performance than the test-mean reference. The supplied values are point estimates, not significance tests.

\begin{table}[!htbp]
\centering\small
\caption{All baseline clinical endpoints, embedding-only probes. Higher $r$ and $R^2$, and lower MAE, are better. MAE retains the recorded target units.}\label{tab:ds-baseline}
\setlength{\tabcolsep}{3pt}
\resizebox{\linewidth}{!}{\begin{tabular}{lrr rrr rrr}
\toprule
& & & \multicolumn{3}{c}{GlucoFM} & \multicolumn{3}{c}{GlucoPRISM}\\
Endpoint & Dev. & Test & $r$ & MAE & $R^2$ & $r$ & MAE & $R^2$\\
\midrule
Apnea--hypopnea index & 4505 & 683 & 0.2273 & 7.7357 & 0.0467 & \textbf{0.3026} & \textbf{7.5818} & \textbf{0.0882}\\
Blood glucose & 3397 & 542 & 0.6348 & 6.8216 & 0.3956 & \textbf{0.6698} & \textbf{6.7141} & \textbf{0.4483}\\
Body mass index & 7956 & 1214 & 0.3803 & 3.0570 & 0.1347 & \textbf{0.4364} & \textbf{3.0105} & \textbf{0.1902}\\
Diastolic pressure & 7971 & 1214 & 0.2214 & 7.8867 & 0.0442 & \textbf{0.2442} & \textbf{7.8359} & \textbf{0.0562}\\
Body-fat field & 4892 & 736 & 0.2601 & 0.0649 & 0.0438 & \textbf{0.4266} & \textbf{0.0605} & \textbf{0.1813}\\
HbA1c & 1900 & 301 & 0.6789 & 0.3066 & 0.4304 & \textbf{0.7307} & \textbf{0.2927} & \textbf{0.5144}\\
HDL cholesterol & 2775 & 444 & 0.2341 & 9.9286 & 0.0280 & \textbf{0.3119} & \textbf{9.6142} & \textbf{0.0824}\\
LDL cholesterol & 2672 & 432 & 0.0781 & 25.3555 & 0.0051 & \textbf{0.1703} & \textbf{25.1441} & \textbf{0.0254}\\
Liver attenuation & 5033 & 750 & 0.1329 & 0.0969 & -0.0261 & \textbf{0.2514} & \textbf{0.0952} & \textbf{0.0573}\\
Liver elasticity & 5035 & 749 & 0.0377 & 0.8521 & -0.0188 & \textbf{0.2648} & \textbf{0.8346} & \textbf{0.0642}\\
Sleep duration & 4569 & 696 & \textbf{0.1424} & 3305.8197 & \textbf{0.0198} & 0.1407 & \textbf{3301.7810} & 0.0194\\
Systolic pressure & 7969 & 1214 & 0.2721 & 13.0048 & 0.0735 & \textbf{0.3094} & \textbf{12.7408} & \textbf{0.0952}\\
Triglycerides & 2834 & 453 & 0.2921 & 41.2313 & 0.0537 & \textbf{0.3819} & \textbf{40.4522} & \textbf{0.1284}\\
Visceral fat mass & 4826 & 726 & 0.4290 & 479.7442 & 0.1835 & \textbf{0.5305} & \textbf{448.6558} & \textbf{0.2809}\\
Waist circumference & 7956 & 1213 & 0.3895 & 8.8896 & 0.1441 & \textbf{0.4446} & \textbf{8.5975} & \textbf{0.1958}\\
\bottomrule
\end{tabular}}
\end{table}
\begin{table}[!htbp]
\centering\small
\caption{All baseline clinical endpoints, embeddings plus common covariates. Higher $r$ and $R^2$, and lower MAE, are better. MAE retains the recorded target units.}\label{tab:ds-baseline-adjusted}
\setlength{\tabcolsep}{3pt}
\resizebox{\linewidth}{!}{\begin{tabular}{lrr rrr rrr}
\toprule
& & & \multicolumn{3}{c}{GlucoFM} & \multicolumn{3}{c}{GlucoPRISM}\\
Endpoint & Dev. & Test & $r$ & MAE & $R^2$ & $r$ & MAE & $R^2$\\
\midrule
Apnea--hypopnea index & 4505 & 683 & 0.3395 & 7.3081 & 0.0903 & \textbf{0.4290} & \textbf{7.0974} & \textbf{0.1809}\\
Blood glucose & 3397 & 542 & 0.6553 & 6.8444 & 0.4229 & \textbf{0.6781} & \textbf{6.7304} & \textbf{0.4596}\\
Body mass index & 7956 & 1214 & 0.4009 & 3.0107 & 0.1518 & \textbf{0.4523} & \textbf{2.9717} & \textbf{0.2042}\\
Diastolic pressure & 7971 & 1214 & 0.2986 & 7.6949 & 0.0866 & \textbf{0.3120} & \textbf{7.6483} & \textbf{0.0964}\\
Body-fat field & 4892 & 736 & 0.6980 & 0.0464 & 0.4870 & \textbf{0.7405} & \textbf{0.0443} & \textbf{0.5439}\\
HbA1c & 1900 & 301 & 0.6942 & 0.2985 & 0.4496 & \textbf{0.7401} & \textbf{0.2870} & \textbf{0.5285}\\
HDL cholesterol & 2775 & 444 & 0.3895 & 9.1893 & 0.1300 & \textbf{0.4157} & \textbf{9.0837} & \textbf{0.1547}\\
LDL cholesterol & 2672 & 432 & 0.1158 & 25.1546 & 0.0124 & \textbf{0.1896} & \textbf{24.9416} & \textbf{0.0329}\\
Liver attenuation & 5033 & 750 & 0.2464 & 0.0937 & 0.0505 & \textbf{0.3425} & \textbf{0.0925} & \textbf{0.1075}\\
Liver elasticity & 5035 & 749 & 0.3052 & 0.7968 & 0.0925 & \textbf{0.3869} & \textbf{0.7848} & \textbf{0.1470}\\
Sleep duration & 4569 & 696 & \textbf{0.1876} & \textbf{3276.1524} & \textbf{0.0338} & 0.1743 & 3276.6819 & 0.0303\\
Systolic pressure & 7969 & 1214 & 0.4836 & 11.6621 & 0.2338 & \textbf{0.4943} & \textbf{11.5956} & \textbf{0.2441}\\
Triglycerides & 2834 & 453 & 0.3126 & 41.0398 & 0.0638 & \textbf{0.3856} & \textbf{40.5872} & \textbf{0.1268}\\
Visceral fat mass & 4826 & 726 & 0.5909 & 422.2313 & 0.3482 & \textbf{0.6395} & \textbf{410.5400} & \textbf{0.4084}\\
Waist circumference & 7956 & 1213 & 0.5294 & 7.9265 & 0.2725 & \textbf{0.5646} & \textbf{7.7544} & \textbf{0.3171}\\
\bottomrule
\end{tabular}}
\end{table}

\subsection{Two-year future levels and changes}
Future labels are measured 18--30 months after the end of baseline CGM and within 90 days of the year-two visit anchor. A baseline value used as a predictor or subtracted to form a change must be available by the baseline CGM cutoff. Measurements are not reused as both baseline and follow-up labels. Development probes for future levels are fitted to later outcomes of other development participants, using only baseline inputs. Future-level prediction estimates $Y_2$; change prediction estimates $Y_2-Y_0$. Good future-level performance does not imply accurate change prediction.

Table~\ref{tab:ds-future} reports all 15 common-test future-level endpoints. GlucoPRISM exceeds GlucoFM on 12/15 correlations, 12/15 MAEs and 13/15 $R^2$ values in the embedding-only comparison. Future HbA1c and LDL still have negative $R^2$ for both models. The common test includes 33--280 labelled people per endpoint. A separate reserved-repeat evaluation includes 35--48 people and supports 10 future-level endpoints; GlucoPRISM leads GlucoFM on 7 correlations, 5 MAEs and 6 $R^2$ values. These populations are not pooled.

Change prediction is less consistent. On 15 common-test change endpoints (12--280 test people), GlucoPRISM leads each of the three metrics on 9 endpoints, but has positive $R^2$ on only 4. The reserved change evaluation supports only BMI, waist and the two blood-pressure endpoints (48 people each); GlucoPRISM leads correlation and $R^2$ on one and MAE on two. These are descriptive endpoint counts, not independent replications. Baseline-trained clinical scores correlated with future values are a separate association analysis and are not treated as directly trained future-outcome models.

\begin{table}[!htbp]
\centering\small
\caption{Two-year future clinical levels on the common test, embedding-only probes. Higher $r$ and $R^2$, and lower MAE, are better. MAE retains the recorded target units.}\label{tab:ds-future}
\setlength{\tabcolsep}{3pt}
\resizebox{\linewidth}{!}{\begin{tabular}{lrr rrr rrr}
\toprule
& & & \multicolumn{3}{c}{GlucoFM} & \multicolumn{3}{c}{GlucoPRISM}\\
Endpoint & Dev. & Test & $r$ & MAE & $R^2$ & $r$ & MAE & $R^2$\\
\midrule
Apnea--hypopnea index & 1181 & 184 & 0.1252 & 6.7930 & 0.0007 & \textbf{0.2644} & \textbf{6.5406} & \textbf{0.0695}\\
Blood glucose & 338 & 53 & 0.6740 & 7.7609 & 0.3696 & \textbf{0.7324} & \textbf{6.9653} & \textbf{0.4845}\\
Body mass index & 1896 & 279 & \textbf{0.3638} & \textbf{2.8492} & \textbf{0.1308} & 0.3557 & 2.8879 & 0.1247\\
Diastolic pressure & 1905 & 280 & \textbf{0.2016} & \textbf{7.1894} & \textbf{0.0317} & 0.1943 & 7.2523 & 0.0312\\
Body-fat field & 1785 & 271 & 0.1933 & 0.0684 & 0.0028 & \textbf{0.4067} & \textbf{0.0637} & \textbf{0.1609}\\
HbA1c & 204 & 33 & 0.3268 & 0.3425 & -0.2109 & \textbf{0.3503} & \textbf{0.3298} & \textbf{-0.1754}\\
HDL cholesterol & 300 & 50 & 0.2171 & \textbf{10.1362} & 0.0030 & \textbf{0.2217} & 10.2496 & \textbf{0.0232}\\
LDL cholesterol & 295 & 49 & -0.0621 & 22.9361 & -0.1327 & \textbf{0.1828} & \textbf{22.8304} & \textbf{-0.0493}\\
Liver attenuation & 1623 & 245 & 0.0483 & 0.0913 & -0.0093 & \textbf{0.1161} & \textbf{0.0902} & \textbf{0.0061}\\
Liver elasticity & 1622 & 245 & -0.0224 & 0.9497 & -0.0168 & \textbf{0.1381} & \textbf{0.9270} & \textbf{0.0118}\\
Sleep duration & 1208 & 188 & \textbf{0.1255} & 3329.3065 & 0.0117 & 0.1232 & \textbf{3307.9820} & \textbf{0.0128}\\
Systolic pressure & 1904 & 280 & 0.1629 & 12.7727 & -0.0344 & \textbf{0.2156} & \textbf{12.5229} & \textbf{0.0187}\\
Triglycerides & 301 & 50 & 0.0575 & 44.5707 & -0.0290 & \textbf{0.2397} & \textbf{43.2003} & \textbf{0.0410}\\
Visceral fat mass & 1772 & 271 & 0.4001 & 436.3434 & 0.1453 & \textbf{0.4417} & \textbf{418.2026} & \textbf{0.1842}\\
Waist circumference & 1904 & 279 & 0.3763 & 8.2247 & 0.1359 & \textbf{0.4227} & \textbf{8.0062} & \textbf{0.1722}\\
\bottomrule
\end{tabular}}
\end{table}
\subsection{Meal response and short-horizon glucose prediction}
Future-glucose tasks encode the preceding 96-position daily context and predict observed glucose 60 or 120 minutes after the final input. Context coverage is at least 80\%, the last input and target must be observed, and sensor-specific timelines are kept separate. Next-day tasks predict the following non-overlapping day's observed mean, coefficient of variation (population standard deviation divided by mean), and fraction of readings above 180 mg/dL; both days require at least 80\% coverage.

For meals, baseline food records with consistent timestamps are clustered when successive entries are at most 30 minutes apart. Meals require at least 10 kcal and no subsequent meal within 120 minutes. The input ends strictly before the meal at an observed grid point no more than 15 minutes earlier. The premeal reference and eight subsequent glucose observations must be present. Peak rise is the maximum signed increase above the premeal reference. Incremental area is the signed trapezoidal glucose integral over the observed 105--120 minute response, in mg/dL$\cdot$minutes; it is neither a classification AUC nor a positive-only area. No response values enter the encoded context.

Examples are capped at eight per participant per task family and selected deterministically. The meal test contains 8,345 examples from 1,069 people; the 60- and 120-minute tests contain 9,213 and 9,214 examples from 1,218 people. Next-day tasks contain 7,975 examples from 1,196 people. Participant weighting prevents people with more observations from dominating the score. Table~\ref{tab:ds-dynamics} retains all seven tasks. GlucoFM has higher correlation and $R^2$ and lower MAE on every task in both predictor sets. Thus the broad clinical gains do not extend to glucose dynamics. These tasks depend on structure within a day, which is finer than the scale at which Trait and State are supervised and which partly shares the Sensor band (Appendix~\ref{app:block-functions}).

\begin{table}[!htbp]
\centering\small
\caption{Meal and short-horizon tasks, embedding-only probes. Next-day fraction uses the threshold 180 mg/dL. Higher $r$ and $R^2$, and lower MAE, are better. MAE retains the recorded target units.}\label{tab:ds-dynamics}
\setlength{\tabcolsep}{3pt}
\resizebox{\linewidth}{!}{\begin{tabular}{lrr rrr rrr}
\toprule
& & & \multicolumn{3}{c}{GlucoFM} & \multicolumn{3}{c}{GlucoPRISM}\\
Endpoint & Dev. & Test & $r$ & MAE & $R^2$ & $r$ & MAE & $R^2$\\
\midrule
Meal incremental area & 7079 & 1069 & \textbf{0.4936} & \textbf{1140.4195} & \textbf{0.2410} & 0.3527 & 1223.9176 & 0.1231\\
Meal peak rise & 7079 & 1069 & \textbf{0.4025} & \textbf{13.9851} & \textbf{0.1602} & 0.3163 & 14.6269 & 0.0978\\
Glucose at +120 min & 7998 & 1218 & \textbf{0.7438} & \textbf{9.2290} & \textbf{0.5530} & 0.7085 & 9.8908 & 0.5017\\
Glucose at +60 min & 7998 & 1218 & \textbf{0.7597} & \textbf{9.0759} & \textbf{0.5769} & 0.7078 & 10.0525 & 0.5009\\
Next-day glucose CV & 7819 & 1196 & \textbf{0.5504} & \textbf{0.0294} & \textbf{0.3027} & 0.5266 & 0.0302 & 0.2772\\
Next-day fraction $>180$ & 7819 & 1196 & \textbf{0.8986} & \textbf{0.0049} & \textbf{0.8065} & 0.8905 & 0.0059 & 0.7863\\
Next-day mean glucose & 7819 & 1196 & \textbf{0.9213} & \textbf{3.9300} & \textbf{0.8482} & 0.9185 & 4.1073 & 0.8422\\
\bottomrule
\end{tabular}}
\end{table}

\subsection{Paired-visit representation stability}
For 44 participants with paired eligible visits, each visit is represented by its mean window embedding. Coordinate standardization is estimated from the 8,000-person development pool and applied unchanged at both visits, with a standard-deviation floor of $10^{-6}$. The mean (median) within-person cosine is 0.2630 (0.5080) for GlucoFM and 0.4275 (0.8028) for GlucoPRISM. No supervised stability probe is fitted. Greater similarity across visits describes representational persistence, not forecast accuracy or physiological responsiveness on its own.

\paragraph{Reporting scope.} Point estimates are reported to four decimals; uncertainty intervals and clinical-unit conversions are not reconstructed from aggregates. These HPP regression tasks are separate from the four public-cohort classification benchmarks and the HPP pretraining-size sweep.


\section{Readout ablations: Trait, State and Sensor}
\label{app:block-functions}
\subsection{What is compared}
These analyses use the same original frozen GlucoPRISM encoder (the primary configuration) for all readouts. Trait comprises coordinates 0--63 (64 dimensions), State coordinates 64--111 (48), and Sensor the last 16. Trait--State retains the first 112 coordinates and All retains 128. A new Ridge probe is fitted for each readout using exactly the same people, targets, development folds, preprocessing and penalty grid described in Appendix~\ref{app:downstream-regression}. No encoder is retrained and no objective is switched off. The primary deployment readout remains Trait--State rather than whichever block wins a particular test endpoint.

The tables below include every clinical endpoint and all five readouts; for compactness we report Pearson correlation. Bold marks the best captured value for each endpoint and metric, including ties at four decimals. The readouts differ in width and the capture contains no equal-width random-coordinate control. Performance differences therefore reflect the utility of the available coordinates under the fitted probe, rather than a pure comparison of physiological factors.

\subsection{Clinical information is overlapping and partly complementary}
Trait--State has higher correlation than both individual blocks on 9/15 baseline endpoints. For example, body-fat correlation rises from 0.3669 (Trait) and 0.3781 (State) to 0.4266 jointly; visceral-fat correlation rises from 0.5042 and 0.5077 to 0.5305. However, adding a block is not uniformly helpful: LDL correlation is highest for Trait alone (0.2050 versus 0.1703 jointly). State predicts HbA1c almost as well as the joint readout (0.7263 versus 0.7307), so persistent clinical information is not exclusive to Trait.

Among the five readouts, the highest correlation occurs in All on seven endpoints, Trait--State on six, and Trait or State alone on one each. Sensor alone leads none of the 15 baseline endpoints, but retains appreciable signal, including HbA1c correlation 0.5691 and visceral-fat correlation 0.4376. Sensor predictiveness may reflect shared clinical information, population/device associations or other correlated features; this readout experiment cannot distinguish these explanations. Small numerical leads, including the 0.0004 body-fat difference between Trait--State and All, are not treated as resolved differences.

\begin{table}[!htbp]\centering\small
\caption{Baseline clinical readout ablation: Pearson correlation (higher is better).}\label{tab:block-clinical-0}
\setlength{\tabcolsep}{5pt}
\begin{tabular}{lrrrrr}\toprule
Endpoint & Trait & State & Sensor & Trait+State & All\\\midrule
Apnea--hypopnea index & 0.2843 & \underline{0.3046} & 0.2830 & 0.3026 & \textbf{0.3076}\\
Blood glucose & 0.6226 & \underline{0.6692} & 0.5811 & \textbf{0.6698} & 0.6509\\
BMI & 0.4036 & 0.4211 & 0.3202 & \textbf{0.4364} & \underline{0.4349}\\
Diastolic pressure & 0.2416 & \underline{0.2526} & 0.2067 & 0.2442 & \textbf{0.2532}\\
Body-fat field & 0.3669 & 0.3781 & 0.2434 & \textbf{0.4266} & \underline{0.4262}\\
HbA1c & 0.7116 & 0.7263 & 0.5691 & \underline{0.7307} & \textbf{0.7336}\\
HDL cholesterol & 0.2948 & 0.2712 & 0.1999 & \underline{0.3119} & \textbf{0.3150}\\
LDL cholesterol & \textbf{0.2050} & 0.1403 & 0.0132 & \underline{0.1703} & 0.1395\\
Liver attenuation & 0.2272 & 0.2163 & 0.2070 & \underline{0.2514} & \textbf{0.2523}\\
Liver elasticity & 0.2125 & 0.2596 & 0.1584 & \textbf{0.2648} & \underline{0.2600}\\
Sleep duration & 0.1377 & 0.1475 & \underline{0.1521} & 0.1407 & \textbf{0.1541}\\
Systolic pressure & 0.3033 & \underline{0.3133} & 0.2609 & 0.3094 & \textbf{0.3149}\\
Triglycerides & 0.3692 & \textbf{0.3834} & 0.3438 & \underline{0.3819} & 0.3769\\
Visceral fat mass & 0.5042 & 0.5077 & 0.4376 & \textbf{0.5305} & \underline{0.5300}\\
Waist circumference & 0.4104 & 0.4345 & 0.3498 & \textbf{0.4446} & \underline{0.4445}\\
\bottomrule\end{tabular}\end{table}
\subsection{The Sensor band carries part of the short-horizon dynamics}
The all-block readout has the highest correlation on each of the two meal and five short-horizon/next-day tasks (Table~\ref{tab:block-dynamic-0}). Adding Sensor increases meal-area correlation from 0.3527 to 0.3669 and peak-rise correlation from 0.3163 to 0.3257. The improvements are smaller for glucose forecasting: for example, +120-minute correlation changes from 0.7085 to 0.7089. Thus dropping Sensor removes some predictive utility in this setting. Nevertheless, all-block GlucoPRISM remains below GlucoFM on correlation for all seven tasks, so Sensor deletion alone does not account for the forecasting gap. The samples, target definitions and participant weighting are unchanged from Appendix~\ref{app:downstream-regression}.

\begin{table}[!htbp]\centering\small
\caption{Meal and short-horizon readout ablation: Pearson correlation (higher is better).}\label{tab:block-dynamic-0}
\setlength{\tabcolsep}{5pt}
\begin{tabular}{lrrrrr}\toprule
Endpoint & Trait & State & Sensor & Trait+State & All\\\midrule
Meal incremental area & 0.2771 & 0.3022 & 0.0848 & \underline{0.3527} & \textbf{0.3669}\\
Meal peak rise & 0.2657 & 0.2785 & 0.1038 & \underline{0.3163} & \textbf{0.3257}\\
Glucose +120 min & 0.6769 & 0.6993 & 0.4237 & \underline{0.7085} & \textbf{0.7089}\\
Glucose +60 min & 0.6714 & 0.6945 & 0.3864 & \underline{0.7078} & \textbf{0.7107}\\
Next-day glucose CV & 0.4996 & 0.4992 & 0.2833 & \underline{0.5266} & \textbf{0.5291}\\
Next-day fraction $>180$ & 0.8360 & 0.8848 & 0.2568 & \underline{0.8905} & \textbf{0.8916}\\
Next-day mean glucose & 0.8716 & 0.9094 & 0.5221 & \underline{0.9185} & \textbf{0.9194}\\
\bottomrule\end{tabular}\end{table}
\subsection{Device identity and visit-to-visit stability}
A logistic probe predicts recorded device group from pooled baseline embeddings, using 8,000 development and 1,219 common-test participants. Preprocessing is learned in development folds. Logistic regularization uses $C=1/\alpha$, with $\alpha\in\{0.1,1,10,100,1000\}$ selected by development squared probability error over the same three participant-grouped folds; the solver uses at most 1,500 iterations, seed 20260918 and no class weighting. Test positive prevalence is approximately 36.59\%. Average precision uses predicted probabilities; balanced accuracy uses threshold 0.5. This is population device identification, not simultaneous paired-device agreement or clinical transfer.

Device group is decodable from every readout (AUROC 0.83 to 0.86). The two HPP sensor variants come from one manufacturer, sample at the same 15-minute rate and are worn by different participants, so this probe measures population differences as much as measurement differences and is not a test of the Sensor contract. A same-day paired CGMacros device probe, in which only the sensor differs, is the cleaner measurement test; it is reported in Appendix~\ref{app:dann}.

For the 44 paired participants, the same development-standardized cosine procedure as Appendix~\ref{app:downstream-regression} is used. Trait has the highest mean (0.4378) and median (0.9848) cosine, followed by State and then Sensor, the order the band contracts specify. The much lower mean than median indicates that these summaries describe different aspects of the distribution; individual-pair data are needed to examine its shape. Persistent representations need not be maximally sensitive to physiological change, and stability is not a prediction metric.

\begin{table}[!htbp]\centering\small
\caption{Device information and paired-visit stability by readout. The first three columns concern device discrimination (1,219 people); the last two concern cosine similarity (44 people). These are different tasks and are not combined into a single score.}\label{tab:block-device-stability}
\begin{tabular}{lrrrrr}\toprule
Readout & Device AUROC & Device AP & Bal. acc. & Mean cosine & Median cosine\\\midrule
Trait & 0.8467 & 0.7793 & 0.7564 & 0.4378 & 0.9848\\
State & 0.8515 & 0.7836 & 0.7741 & 0.4024 & 0.9229\\
Sensor & 0.8300 & 0.7354 & 0.7277 & 0.2667 & 0.4347\\
T+S & 0.8599 & 0.7982 & 0.7703 & 0.4275 & 0.8028\\
All & 0.8607 & 0.7992 & 0.7670 & 0.4145 & 0.7113\\
\bottomrule\end{tabular}\end{table}
\subsection{Scope of contribution attribution}
These experiments localize accessible predictive information and measure the effect of changing the inference readout with the encoder fixed. They do not retrain without sensor supervision, day supervision, CMP, the cross-block correlation penalty, the variance floor or Sensor KL. Consequently, no improvement is attributed to a particular loss from these tables. Loss-level attribution requires matched retraining runs that change one term while preserving the other terms, preprocessing, training data, readout and evaluation protocol, with repeated seeds. Jointly learned blocks can contain overlapping information even when their training objectives target different roles. All figures and tables here report point estimates without uncertainty intervals.

\section{Postprandial response on HPP}\label{app:ppgr}

Section~\ref{sec:hpp} reports this cohort's trait endpoints. This appendix gives
the construction, the validity checks and the full table for its state endpoint,
postprandial glycemic response.

\paragraph{Evaluation rationale.} Postprandial response tests whether the
representation captures variation across meals within a participant. This
complements the clinical trait endpoints and probes a finer time scale than the
State contract, which is defined across days rather than across meals. We evaluate between-participant and within-participant prediction
separately and compare learned readouts with glucose and meal-based controls.

\paragraph{From food items to meals.} The diet-logging domain holds
\num{\ppgritems} logged food items. Items are not meals: a meal is a cluster of
items entered together, so items within 30 minutes of one another are pooled
into one eating occasion and their macronutrients summed, giving
\num{\ppgroccasions} occasions. An occasion is kept only if it is
\emph{isolated}, with no other occasion within two hours on either side, because
a second meal inside the response window contaminates the label and a meal
shortly before it leaves the pre-meal reading unusable as a baseline. At a
threshold of 20\,g of carbohydrate that leaves \num{\ppgrisolated} meals, of
which \num{\ppgrmeals} fall inside a CGM wear with sufficient context and a
complete response: \ppgrpermeal{} per participant across \num{\ppgrsubj}
participants, at observation fraction \ppgrobsfrac{} against the $0.333$ a
15-minute sensor can reach.

\paragraph{Temporal separation of inputs and targets.}
Postprandial response is computed from the same recording as the context, so
input and response intervals must be separated to prevent target leakage. The input is the 24
hours ending at the meal timestamp and the label is incremental area over the
two hours strictly after it. The baseline subtracted inside the label is the
last context reading, so the label reads one sample the input also saw, but no
response sample ever enters the input. The build asserts this on every window
rather than assuming it.

\paragraph{Meal-alignment checks.} A mis-anchored window
scores like a correctly anchored one on any aggregate, so the anchor was
validated before a single model was run. Mean glucose rises to
$+\ppgrpeakrise$\,mg/dL at \ppgrpeakmin{} minutes and decays monotonically
after, which is the shape a mixed meal produces when seen through interstitial
fluid. Mean incremental area increases across every carbohydrate quartile, from
\ppgrcarbqone{} to \ppgrcarbqfour{}\,mg/dL$\cdot$min. And the same windows
anchored six hours later, where no meal occurred, show smaller aggregate responses: mean peak rise \ppgrpeaknull{} against \ppgrpeakreal, median area
\ppgrmednull{} against \ppgrmedreal. These checks support meal alignment, although they do not establish that circadian influences are absent.

\paragraph{Between- and within-participant evaluation.} Variance in the label
is \ppgrbetweenvar\% between participants and \ppgrwithinvar\% within, and a
single pooled correlation answers both questions at once. \emph{Between}
correlates per-participant means and asks who is a high responder, which is a
trait a model can predict while knowing nothing about any particular meal.
\emph{Within} subtracts each participant's own mean from both truth and
prediction and asks which of that person's meals spiked, which tests meal-level
variation within a person. We report pooled for comparability
with the postprandial literature and never as the headline. The stated
between-participant share is an upper bound on the trait-like part, because a
between-group sum of squares computed from group means is inflated by
within-group noise at \ppgrpermeal{} meals per participant.

\paragraph{Baseline construction.} Self-reported gram
weights carry logging errors, and one logged meal in this cohort claims
\ppgrcarbmax\,g of carbohydrate. A raw Pearson correlation is sensitive to that
tail: carbohydrate against incremental area gives \ppgrcarbpear{} untreated
against \ppgrcarbspear{} by rank. The macronutrient features are therefore
winsorised at the 1st and 99th percentiles, which raises the correlation to
\ppgrcarbwins{} for the meal-based baseline. The stronger baseline adds
pre-meal glucose and the meal's time of day. Because each window ends at its
meal, the circadian start index encodes the meal's time of day, so time of day
is also probed alone to assess the contribution of circadian information.

\paragraph{Results.} Every encoder lands between \ppgrbetweenlo{}
and \ppgrbetweenhi{} on the trait column and between \ppgrwithinlo{} and
\ppgrwithinhi{} on the state column. Pre-meal glucose alone reaches \ppgrbaseglucwithin{}, while the raw-window baseline reaches 0.440. These comparisons indicate that retaining the pre-meal glucose value is a strong baseline for this endpoint. Stacked on the full
classical baseline, the largest within-participant gain any representation
provides is \ppgrfmgain{} with a 95\% interval of
$[\ppgrfmgainlo, \ppgrfmgainhi]$, from a participant-clustered paired bootstrap
over \num{\ppgrreps} replicates. The between-participant and within-participant results should be interpreted separately; neither band alone isolates meal-level variation, which is finer than the day-level scale at which State is supervised.

\paragraph{Measurement and evaluation scope.} A 15-minute sensor
offers eight samples in a two-hour response, so incremental area is coarse and
the peak is blunted by the sensor's own smoothing; this applies equally to every
arm. Meal logs are self-reported; winsorising reduces the influence of extreme
values but does not remove logging error. Pre-meal glucose sits inside the input window and the label is
defined as a rise above it, so part of the predictable variance is simply where
glucose started; this is why a single feature is so strong, and it is also why
the retention column, rather than the within column alone, is what carries the
argument. Finally, the ridge strength is selected on an inner subject-disjoint
split of the training folds by pooled correlation, which is the natural fitting
objective but not the within-participant metric, so a stacked readout can fall
below its own baseline on that column. The fold-shuffle spread over
\ppgriters{} repetitions is near $0.0005$ throughout, so the intervals that
matter come from the participant bootstrap and not from the folds.

\begin{table}[!htbp]\centering\footnotesize\setlength{\tabcolsep}{2.6pt}
\caption{Postprandial response for \num{\ppgrmeals} meals from \num{\ppgrsubj} participants. Frozen encoders read the preceding 24 hours; incremental area uses the following two hours. No response sample enters the input, but the last pre-meal input reading is subtracted as the label baseline. Ridge probes use subject-grouped folds. Between-participant correlation compares participant means; within-participant correlation centers within each participant. The final column separately probes retention of pre-meal glucose.}
\label{tab:ppgr}

\begin{tabular}{lr rrr >{\columncolor{oursbg}}r}
\toprule
Readout & dim & pooled $r$ & between $r$ & within $r$ & recovers pre-meal \\
\midrule
\multicolumn{6}{l}{\emph{Baselines that need no learned representation}} \\
Meal time of day (2 features) & 2 & 0.083 & 0.028 & 0.097 & -- \\
Macronutrients, winsorised & 4 & 0.195 & 0.150 & 0.218 & -- \\
Pre-meal glucose (1 feature) & 1 & 0.265 & -0.006 & 0.437 & -- \\
Macronutrients + clock + pre-meal & 7 & 0.343 & 0.067 & \textbf{0.483} & -- \\
\addlinespace[2pt]\multicolumn{6}{l}{\emph{CGM foundation models, frozen-encoder probing}} \\
Interpolated window (raw) & 288 & 0.474 & 0.550 & \underline{0.440} & \textbf{1.000} \\
CGM-JEPA & 96 & \underline{0.501} & 0.664 & 0.434 & 0.914 \\
X-CGM-JEPA & 96 & \textbf{0.507} & \underline{0.678} & 0.435 & 0.929 \\
GluFormer-tiny & 128 & 0.266 & 0.613 & 0.096 & 0.342 \\
GlucoFM (our reproduction) & 128 & 0.460 & \textbf{0.699} & 0.355 & 0.678 \\
\addlinespace[2pt]\multicolumn{6}{l}{\emph{Ours}} \\
\rowcolor{oursbg}\textbf{GlucoPRISM} $[\zt\|\zs]$ & 112 & 0.358 & 0.649 & 0.212 & 0.443 \\
\quad $\zt$ only & 64 & 0.316 & 0.635 & 0.155 & 0.361 \\
\quad $\zs$ only & 48 & 0.303 & 0.623 & 0.130 & 0.355 \\
\quad $\za$ only & 16 & 0.144 & 0.415 & 0.014 & 0.087 \\
\bottomrule
\end{tabular}

\vspace{4pt}\par\footnotesize\begin{minipage}{\textwidth}Across the \num{\ppgrnread} readouts, within-participant prediction and pre-meal retention correlate at $r=\ppgrretention$. Pre-meal glucose alone reaches \ppgrbaseglucwithin{}, while the raw-window readout reaches 0.440; this small point-estimate difference does not establish superiority. CGM-JEPA and X-CGM-JEPA reach 0.434 and 0.435. The association with baseline retention helps interpret the task, but does not identify why a particular encoder loses local information.\end{minipage}
\end{table}

\paragraph{Device-specific MAE results.} Tables~\ref{tab:ppgr-dexcom} and~\ref{tab:ppgr-libre} give the Dexcom and Libre results averaged in Table~\ref{tab:ppgr-mae}. This error-based protocol follows \citet{li2026glucofm} and differs from the correlation analysis above. GlucoPRISM has lower error than GlucoFM in 13 of 20 Dexcom settings and 14 of 20 Libre settings. X-CGM-JEPA has the lowest error in 16 of 20 Libre settings.
\begin{table}[!htbp]\centering\footnotesize\setlength{\tabcolsep}{3pt}
\caption{PPGR on the Dexcom evaluation (5-min native grid). Mean absolute error (lower is better), mean $\pm$ SD over 10 probe seeds. \textbf{Bold}: best; \underline{underline}: second best.}
\label{tab:ppgr-dexcom}
\resizebox{\linewidth}{!}{%
\begin{tabular}{llrrrr>{\columncolor{oursbg}}r}
\toprule
Context & Endpoint & GluFormer (tiny) & CGM-JEPA & X-CGM-JEPA & GlucoFM & GlucoPRISM \\
\midrule
Frozen representation & Trajectory (mg/dL) & 26.55\,{\scriptsize$\pm$\,0.26} & \underline{24.90\,{\scriptsize$\pm$\,0.28}} & \textbf{24.65\,{\scriptsize$\pm$\,0.18}} & 26.64\,{\scriptsize$\pm$\,0.33} & 26.43\,{\scriptsize$\pm$\,0.14} \\
 & Positive iAUC (mg/dL$\cdot$h) & 37.85\,{\scriptsize$\pm$\,0.75} & \underline{35.16\,{\scriptsize$\pm$\,0.50}} & \textbf{34.67\,{\scriptsize$\pm$\,0.33}} & 37.92\,{\scriptsize$\pm$\,0.71} & 37.79\,{\scriptsize$\pm$\,0.32} \\
 & Peak rise (mg/dL) & 35.07\,{\scriptsize$\pm$\,1.09} & \underline{32.66\,{\scriptsize$\pm$\,0.63}} & \textbf{32.25\,{\scriptsize$\pm$\,0.47}} & 34.91\,{\scriptsize$\pm$\,0.72} & 35.53\,{\scriptsize$\pm$\,0.31} \\
 & Peak time (min) & 30.52\,{\scriptsize$\pm$\,1.62} & \underline{30.29\,{\scriptsize$\pm$\,0.33}} & 30.44\,{\scriptsize$\pm$\,0.49} & 31.50\,{\scriptsize$\pm$\,1.77} & \textbf{30.04\,{\scriptsize$\pm$\,0.75}} \\
\midrule
+ 1 h pre-meal CGM & Trajectory (mg/dL) & 25.75\,{\scriptsize$\pm$\,0.24} & 24.37\,{\scriptsize$\pm$\,0.15} & 24.27\,{\scriptsize$\pm$\,0.17} & \underline{24.25\,{\scriptsize$\pm$\,0.13}} & \textbf{24.08\,{\scriptsize$\pm$\,0.21}} \\
 & Positive iAUC (mg/dL$\cdot$h) & 36.57\,{\scriptsize$\pm$\,0.57} & 34.61\,{\scriptsize$\pm$\,0.30} & 34.41\,{\scriptsize$\pm$\,0.30} & \underline{34.21\,{\scriptsize$\pm$\,0.29}} & \textbf{33.92\,{\scriptsize$\pm$\,0.44}} \\
 & Peak rise (mg/dL) & 33.38\,{\scriptsize$\pm$\,0.61} & 32.04\,{\scriptsize$\pm$\,0.47} & \textbf{31.71\,{\scriptsize$\pm$\,0.39}} & \underline{31.90\,{\scriptsize$\pm$\,0.36}} & 31.96\,{\scriptsize$\pm$\,0.24} \\
 & Peak time (min) & 29.81\,{\scriptsize$\pm$\,0.63} & 29.39\,{\scriptsize$\pm$\,0.45} & 29.54\,{\scriptsize$\pm$\,0.60} & \underline{29.15\,{\scriptsize$\pm$\,0.54}} & \textbf{28.92\,{\scriptsize$\pm$\,0.68}} \\
\midrule
+ Meal nutrition & Trajectory (mg/dL) & 24.43\,{\scriptsize$\pm$\,0.28} & 23.03\,{\scriptsize$\pm$\,0.15} & 22.86\,{\scriptsize$\pm$\,0.15} & \textbf{22.71\,{\scriptsize$\pm$\,0.27}} & \underline{22.78\,{\scriptsize$\pm$\,0.31}} \\
 & Positive iAUC (mg/dL$\cdot$h) & 33.88\,{\scriptsize$\pm$\,0.65} & 32.05\,{\scriptsize$\pm$\,0.35} & 31.77\,{\scriptsize$\pm$\,0.30} & \textbf{31.18\,{\scriptsize$\pm$\,0.65}} & \underline{31.57\,{\scriptsize$\pm$\,0.64}} \\
 & Peak rise (mg/dL) & 30.97\,{\scriptsize$\pm$\,0.68} & 29.70\,{\scriptsize$\pm$\,0.44} & 29.36\,{\scriptsize$\pm$\,0.33} & \textbf{28.99\,{\scriptsize$\pm$\,0.46}} & \underline{29.14\,{\scriptsize$\pm$\,0.59}} \\
 & Peak time (min) & 29.83\,{\scriptsize$\pm$\,0.56} & 29.48\,{\scriptsize$\pm$\,0.39} & 29.71\,{\scriptsize$\pm$\,0.52} & \underline{29.29\,{\scriptsize$\pm$\,0.45}} & \textbf{29.16\,{\scriptsize$\pm$\,0.50}} \\
\midrule
+ Fasting glucose & Trajectory (mg/dL) & 24.30\,{\scriptsize$\pm$\,0.34} & 22.86\,{\scriptsize$\pm$\,0.22} & 22.82\,{\scriptsize$\pm$\,0.18} & \underline{22.66\,{\scriptsize$\pm$\,0.23}} & \textbf{22.48\,{\scriptsize$\pm$\,0.20}} \\
 & Positive iAUC (mg/dL$\cdot$h) & 33.67\,{\scriptsize$\pm$\,0.81} & 31.82\,{\scriptsize$\pm$\,0.52} & 31.73\,{\scriptsize$\pm$\,0.35} & \underline{31.18\,{\scriptsize$\pm$\,0.51}} & \textbf{30.95\,{\scriptsize$\pm$\,0.45}} \\
 & Peak rise (mg/dL) & 30.66\,{\scriptsize$\pm$\,0.76} & 29.10\,{\scriptsize$\pm$\,0.42} & 28.99\,{\scriptsize$\pm$\,0.28} & \textbf{28.79\,{\scriptsize$\pm$\,0.44}} & \underline{28.81\,{\scriptsize$\pm$\,0.33}} \\
 & Peak time (min) & 29.79\,{\scriptsize$\pm$\,0.73} & \textbf{28.44\,{\scriptsize$\pm$\,0.30}} & \underline{29.02\,{\scriptsize$\pm$\,0.53}} & 29.12\,{\scriptsize$\pm$\,0.52} & 29.12\,{\scriptsize$\pm$\,0.61} \\
\midrule
+ BMI + diabetes status & Trajectory (mg/dL) & 24.54\,{\scriptsize$\pm$\,0.32} & 23.23\,{\scriptsize$\pm$\,0.15} & \textbf{22.99\,{\scriptsize$\pm$\,0.22}} & 23.17\,{\scriptsize$\pm$\,0.32} & \underline{23.03\,{\scriptsize$\pm$\,0.30}} \\
 & Positive iAUC (mg/dL$\cdot$h) & 34.17\,{\scriptsize$\pm$\,0.63} & 32.47\,{\scriptsize$\pm$\,0.32} & \textbf{31.90\,{\scriptsize$\pm$\,0.42}} & 32.09\,{\scriptsize$\pm$\,0.55} & \underline{31.97\,{\scriptsize$\pm$\,0.63}} \\
 & Peak rise (mg/dL) & 30.95\,{\scriptsize$\pm$\,0.68} & 29.79\,{\scriptsize$\pm$\,0.46} & \textbf{29.08\,{\scriptsize$\pm$\,0.38}} & 29.50\,{\scriptsize$\pm$\,0.52} & \underline{29.37\,{\scriptsize$\pm$\,0.62}} \\
 & Peak time (min) & 29.81\,{\scriptsize$\pm$\,0.81} & 29.39\,{\scriptsize$\pm$\,0.56} & \underline{29.28\,{\scriptsize$\pm$\,0.49}} & 29.81\,{\scriptsize$\pm$\,0.65} & \textbf{29.07\,{\scriptsize$\pm$\,0.59}} \\
\bottomrule
\end{tabular}}
\end{table}
\begin{table}[!htbp]\centering\footnotesize\setlength{\tabcolsep}{3pt}
\caption{PPGR on the Libre evaluation (15-min native grid). Mean absolute error (lower is better), mean $\pm$ SD over 10 probe seeds. \textbf{Bold}: best; \underline{underline}: second best.}
\label{tab:ppgr-libre}
\resizebox{\linewidth}{!}{%
\begin{tabular}{llrrrr>{\columncolor{oursbg}}r}
\toprule
Context & Endpoint & GluFormer (tiny) & CGM-JEPA & X-CGM-JEPA & GlucoFM & GlucoPRISM \\
\midrule
Frozen representation & Trajectory (mg/dL) & 25.17\,{\scriptsize$\pm$\,0.21} & \underline{22.56\,{\scriptsize$\pm$\,0.47}} & \textbf{22.21\,{\scriptsize$\pm$\,0.49}} & 25.03\,{\scriptsize$\pm$\,0.13} & 25.31\,{\scriptsize$\pm$\,0.10} \\
 & Positive iAUC (mg/dL$\cdot$h) & 34.24\,{\scriptsize$\pm$\,0.53} & \underline{30.03\,{\scriptsize$\pm$\,0.82}} & \textbf{29.41\,{\scriptsize$\pm$\,0.97}} & 33.68\,{\scriptsize$\pm$\,0.17} & 34.61\,{\scriptsize$\pm$\,0.30} \\
 & Peak rise (mg/dL) & 30.12\,{\scriptsize$\pm$\,0.88} & \underline{26.82\,{\scriptsize$\pm$\,0.68}} & \textbf{26.38\,{\scriptsize$\pm$\,0.70}} & 30.33\,{\scriptsize$\pm$\,0.26} & 31.10\,{\scriptsize$\pm$\,0.38} \\
 & Peak time (min) & 31.26\,{\scriptsize$\pm$\,0.54} & 30.73\,{\scriptsize$\pm$\,0.64} & \textbf{30.51\,{\scriptsize$\pm$\,0.63}} & 31.04\,{\scriptsize$\pm$\,0.86} & \underline{30.68\,{\scriptsize$\pm$\,0.55}} \\
\midrule
+ 1 h pre-meal CGM & Trajectory (mg/dL) & 24.18\,{\scriptsize$\pm$\,0.40} & \underline{22.17\,{\scriptsize$\pm$\,0.47}} & \textbf{21.90\,{\scriptsize$\pm$\,0.44}} & 22.66\,{\scriptsize$\pm$\,0.20} & 22.54\,{\scriptsize$\pm$\,0.20} \\
 & Positive iAUC (mg/dL$\cdot$h) & 32.81\,{\scriptsize$\pm$\,0.75} & \underline{29.71\,{\scriptsize$\pm$\,0.84}} & \textbf{29.31\,{\scriptsize$\pm$\,0.79}} & 30.19\,{\scriptsize$\pm$\,0.42} & 30.24\,{\scriptsize$\pm$\,0.36} \\
 & Peak rise (mg/dL) & 28.21\,{\scriptsize$\pm$\,0.80} & \underline{26.32\,{\scriptsize$\pm$\,0.52}} & \textbf{25.84\,{\scriptsize$\pm$\,0.59}} & 27.11\,{\scriptsize$\pm$\,0.25} & 27.20\,{\scriptsize$\pm$\,0.33} \\
 & Peak time (min) & 31.64\,{\scriptsize$\pm$\,0.94} & 31.47\,{\scriptsize$\pm$\,0.55} & \underline{30.87\,{\scriptsize$\pm$\,0.87}} & 31.66\,{\scriptsize$\pm$\,0.39} & \textbf{30.57\,{\scriptsize$\pm$\,0.53}} \\
\midrule
+ Meal nutrition & Trajectory (mg/dL) & 22.46\,{\scriptsize$\pm$\,0.29} & \underline{20.65\,{\scriptsize$\pm$\,0.16}} & \textbf{20.57\,{\scriptsize$\pm$\,0.19}} & 21.13\,{\scriptsize$\pm$\,0.31} & 20.72\,{\scriptsize$\pm$\,0.21} \\
 & Positive iAUC (mg/dL$\cdot$h) & 29.54\,{\scriptsize$\pm$\,0.43} & \underline{26.78\,{\scriptsize$\pm$\,0.30}} & \textbf{26.68\,{\scriptsize$\pm$\,0.39}} & 27.10\,{\scriptsize$\pm$\,0.50} & 26.89\,{\scriptsize$\pm$\,0.43} \\
 & Peak rise (mg/dL) & 24.85\,{\scriptsize$\pm$\,0.27} & \underline{23.58\,{\scriptsize$\pm$\,0.43}} & \textbf{23.40\,{\scriptsize$\pm$\,0.21}} & 24.17\,{\scriptsize$\pm$\,0.42} & 23.76\,{\scriptsize$\pm$\,0.43} \\
 & Peak time (min) & 31.36\,{\scriptsize$\pm$\,1.02} & \underline{30.92\,{\scriptsize$\pm$\,0.42}} & 30.95\,{\scriptsize$\pm$\,0.55} & 31.98\,{\scriptsize$\pm$\,0.48} & \textbf{29.89\,{\scriptsize$\pm$\,0.58}} \\
\midrule
+ Fasting glucose & Trajectory (mg/dL) & 22.57\,{\scriptsize$\pm$\,0.32} & \underline{20.62\,{\scriptsize$\pm$\,0.23}} & \textbf{20.52\,{\scriptsize$\pm$\,0.27}} & 21.19\,{\scriptsize$\pm$\,0.34} & 20.70\,{\scriptsize$\pm$\,0.17} \\
 & Positive iAUC (mg/dL$\cdot$h) & 29.76\,{\scriptsize$\pm$\,0.72} & \underline{26.84\,{\scriptsize$\pm$\,0.42}} & \textbf{26.53\,{\scriptsize$\pm$\,0.53}} & 27.40\,{\scriptsize$\pm$\,0.59} & 26.95\,{\scriptsize$\pm$\,0.33} \\
 & Peak rise (mg/dL) & 24.77\,{\scriptsize$\pm$\,0.28} & \underline{23.12\,{\scriptsize$\pm$\,0.47}} & \textbf{23.02\,{\scriptsize$\pm$\,0.43}} & 23.99\,{\scriptsize$\pm$\,0.46} & 23.50\,{\scriptsize$\pm$\,0.31} \\
 & Peak time (min) & 31.76\,{\scriptsize$\pm$\,0.78} & 30.52\,{\scriptsize$\pm$\,0.50} & \underline{30.43\,{\scriptsize$\pm$\,0.73}} & 30.90\,{\scriptsize$\pm$\,0.46} & \textbf{29.43\,{\scriptsize$\pm$\,0.56}} \\
\midrule
+ BMI + diabetes status & Trajectory (mg/dL) & 22.94\,{\scriptsize$\pm$\,0.45} & \underline{21.01\,{\scriptsize$\pm$\,0.20}} & \textbf{20.84\,{\scriptsize$\pm$\,0.29}} & 21.37\,{\scriptsize$\pm$\,0.19} & 21.14\,{\scriptsize$\pm$\,0.12} \\
 & Positive iAUC (mg/dL$\cdot$h) & 30.50\,{\scriptsize$\pm$\,0.91} & \underline{27.47\,{\scriptsize$\pm$\,0.45}} & \textbf{27.13\,{\scriptsize$\pm$\,0.45}} & 27.81\,{\scriptsize$\pm$\,0.30} & 27.84\,{\scriptsize$\pm$\,0.30} \\
 & Peak rise (mg/dL) & 25.12\,{\scriptsize$\pm$\,0.35} & \underline{23.48\,{\scriptsize$\pm$\,0.37}} & \textbf{23.20\,{\scriptsize$\pm$\,0.32}} & 24.19\,{\scriptsize$\pm$\,0.44} & 23.80\,{\scriptsize$\pm$\,0.28} \\
 & Peak time (min) & 31.61\,{\scriptsize$\pm$\,0.54} & 31.16\,{\scriptsize$\pm$\,0.62} & \underline{30.99\,{\scriptsize$\pm$\,0.40}} & 31.33\,{\scriptsize$\pm$\,0.78} & \textbf{29.93\,{\scriptsize$\pm$\,0.55}} \\
\bottomrule
\end{tabular}}
\end{table}

\section{Block dependence and representation stability}\label{app:dependence}\label{sec:fails}
\label{app:fewshot}
These diagnostics evaluate block dependence, accessible device information and across-day consistency.

\head{Cross-block dependence.} $L_{\mathrm{indep}}$ penalises off-diagonal
\emph{correlation}, which is a linear measure, so we also measure HSIC with RBF
kernels, which detects dependence a correlation matrix cannot see. Averaged over
three seeds and three cohorts, Trait and State remain strongly dependent by both
measures ($|r| = \corrts$, HSIC $= \hsicts$). Sensor has lower measured dependence on both (HSIC $\hsicta$ against Trait and $\hsicsa$ against State).
Lower dependence is compatible with distinct accessible information, but does not identify a pure device factor. Both blocks outperform their width-matched controls
(Figure~\ref{fig:readouts}), showing useful predictive information alongside
substantial dependence between the blocks.

\head{Device information in acquisition features.} A linear probe under
subject-grouped cross-validation recovers which sensor produced a window at
$\devmask$ \auc{} from the \emph{observation mask alone}, and at $\devcount$
from a single scalar, the number of observed samples. Three glucose-level scalars, without mask information, also recover device
identity at $\devlevel$ \auc. Real paired recordings also exhibit a systematic offset of $-31.1 \pm 15.8$~mg/dL, with 43 of 44 subjects in the same direction. Acquisition differences can therefore remain visible in glucose values even when observation masks are omitted. Population-level device predictability may additionally reflect cohort composition.

\section{Embedding visualisation}\label{app:embed}
\begin{figure}[!htbp]\centering
\includegraphics[width=\linewidth]{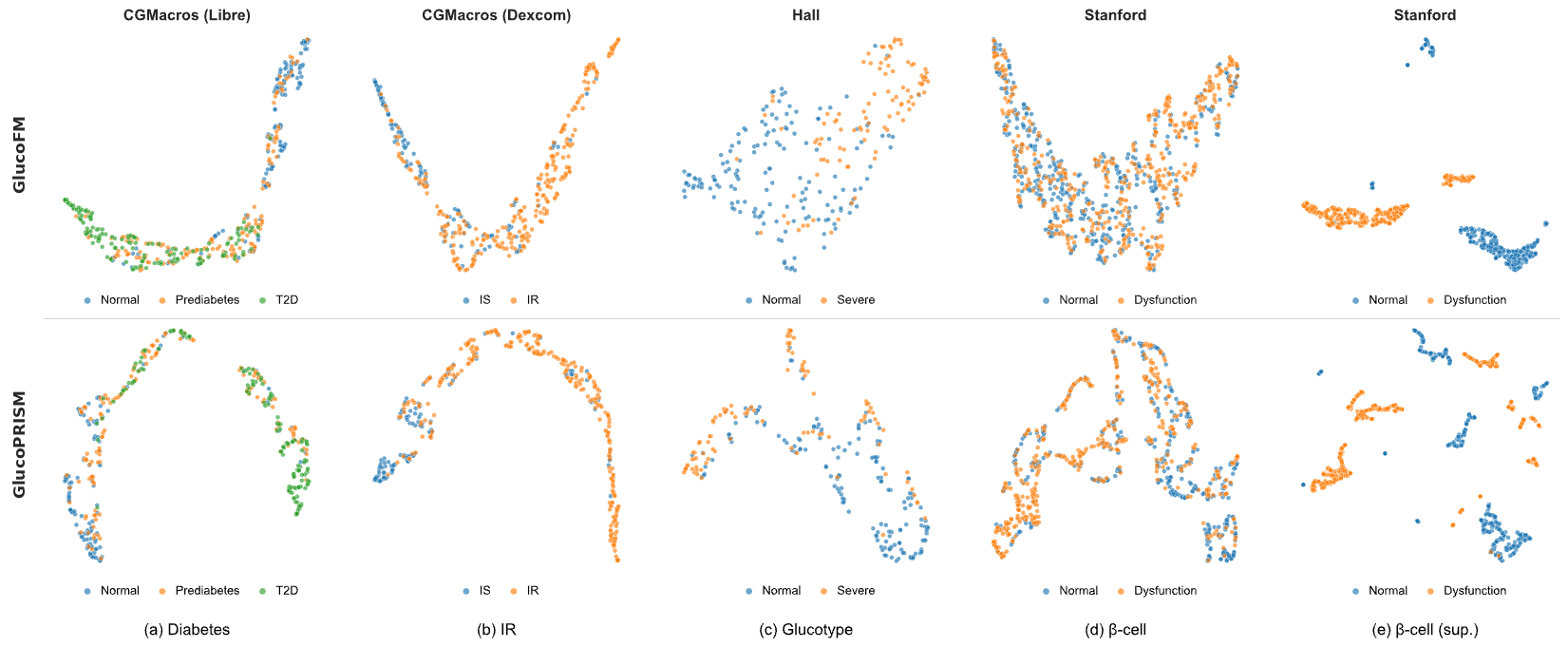}
\caption{Two-dimensional projections of frozen representations for GlucoFM (top) and GlucoPRISM (bottom), coloured by label: (a) diabetes status on CGMacros (Libre), (b) insulin resistance on CGMacros (Dexcom), (c) glucotype on Hall, (d) \(\beta\)-cell dysfunction on Stanford, and (e) the same Stanford labels with a label-supervised projection. Both rows use the same windows. Panel (e) separates classes by construction and is shown for reference only.}
\label{fig:embed-compare}
\end{figure}

Figure~\ref{fig:embed-compare} projects frozen GlucoFM and GlucoPRISM representations of the same evaluation windows to two dimensions and colours them by label. The projections are qualitative illustrations; quantitative comparisons are reported in Section~\ref{sec:headline} and Appendix~\ref{app:percell}. Panels (a) to (d) are unsupervised projections. Panel (e) uses the labels to fit the projection, so its separation is expected by construction and is not evidence about either encoder.

\section{Multi-day aggregation}\label{app:multiday}
\begin{figure}[!htbp]\centering
\includegraphics[width=0.79\linewidth]{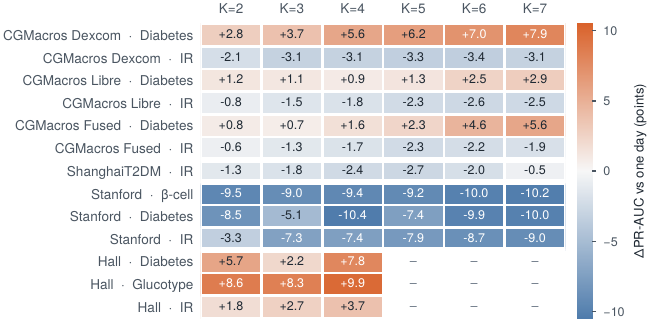}
\caption{Effect of multi-day observation for GlucoPRISM: paired PR-AUC change from one-day to $K$-day subject representations ($K=2,\dots,7$), in percentage points, for each task and cohort. Following \citet{li2026glucofm}, Hall uses $K\le 4$ and the other cohorts use $K\le 7$; dashes mark values of $K$ that are not evaluated. Cells are point estimates without uncertainty intervals.}
\label{fig:multiday}
\end{figure}

Following \citet{li2026glucofm}, Figure~\ref{fig:multiday} reports the paired PR-AUC change for GlucoPRISM when a subject is represented by $K$ days instead of one. Pooling more days increases diabetes PR-AUC on CGMacros and Hall and increases glucotype PR-AUC on Hall, but decreases PR-AUC for every Stanford task and for insulin resistance on CGMacros and Shanghai. The effect of multi-day aggregation therefore depends on the cohort and task. These per-cell changes are relative to each model's own one-day score; against GlucoFM, GlucoPRISM remains higher at every number of days in the subject-level analysis of Table~\ref{tab:multiday}. Within the prism view, a Trait band that is already stable across days leaves less for extra days to add, so small or negative per-cell changes are expected where one day already describes the person.

\begin{figure}[!htbp]\centering
\includegraphics[width=0.46\linewidth]{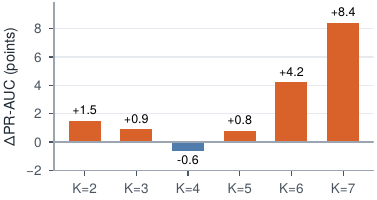}
\caption{Multi-day observation for GlucoPRISM on ShanghaiT2DM insulin resistance under concat(mean, max) aggregation, following \citet{li2026glucofm}: paired PR-AUC change from one-day to $K$-day subject representations, in percentage points. Bars are point estimates without uncertainty intervals.}
\label{fig:multiday-shanghai}
\end{figure}

GlucoFM also reports one cell under an alternative aggregation. Figure~\ref{fig:multiday-shanghai} gives the corresponding GlucoPRISM result for ShanghaiT2DM insulin resistance with concat(mean, max) aggregation. Under this aggregation the gain rises from 1.5 points at $K=2$ to 8.4 points at $K=7$, with a small loss at $K=4$. This differs from the mean-aggregation result for the same cell in Figure~\ref{fig:multiday}, so the multi-day effect depends on the aggregation as well as the cohort.

\section{Additional ablation studies}\label{app:extendedablations}
\head{Monitoring duration.} The probe reads the mean of embeddings from $K$ randomly drawn 24-hour windows per subject (Table~\ref{tab:multiday}). At one day, the gain over GlucoFM is $\mdgainone$, and at seven days it is $\mdgainseven$. GlucoPRISM at three days ($\mdcthree$) exceeds its own full readout at seven ($\mdcfullseven$). These comparisons characterize monitoring-duration tradeoffs in this benchmark, without establishing a general substitution between deletion and additional wear days or independently identifying a mechanism.

\begin{table}[!htbp]\centering\footnotesize\setlength{\tabcolsep}{5pt}
\caption{Subject-level ROC-AUC using mean embeddings from $K$ randomly drawn 24-hour windows per subject; the windows need not be consecutive. The effect of additional days differs by model and readout; these comparisons do not establish that a representation change substitutes for a fixed number of clinical monitoring days.}
\label{tab:multiday}

\begin{tabular}{lrrr}
\toprule
Days $K$ & GlucoFM & \textbf{GlucoPRISM} & GlucoPRISM [$z_A$ kept] \\
\midrule
1 & 61.42 & \textbf{64.94} & \underline{63.94} \\
2 & 64.80 & \textbf{67.15} & \underline{66.37} \\
3 & 65.74 & \textbf{68.41} & \underline{67.13} \\
5 & \underline{68.26} & \textbf{68.88} & 67.71 \\
7 & \underline{68.81} & \textbf{69.01} & 68.06 \\
\bottomrule
\end{tabular}

\vspace{4pt}\par\footnotesize\begin{minipage}{\textwidth}GlucoPRISM at three days (\mdcthree) exceeds GlucoFM at five days (68.26) and the same model with $z_A$ retained at seven (\mdcfullseven); these comparisons use randomly drawn windows and do not establish an equivalence to additional consecutive days of clinical monitoring.\end{minipage}
\end{table}

\head{Device-adversarial supervision.} We also evaluated a
gradient-reversal adversary~\citep{ganin2015dann} on the Sensor head to reduce
the predictability of device identity from $\za$. At adversarial weight~3 the Sensor block's own clinical
probe score \emph{rose} to $\grladv$ \auc{} against $\grlctrl$ for the control.
The adversary had suppressed one discriminator while leaving the block
\emph{more} readable rather than less~\citep{elazar2018adversarial}. This result motivated using a KL penalty to constrain Sensor capacity in the
released models.

\section{Design ablations}\label{app:ablations}
\begin{figure}[!htbp]\centering
\includegraphics[width=0.9\linewidth]{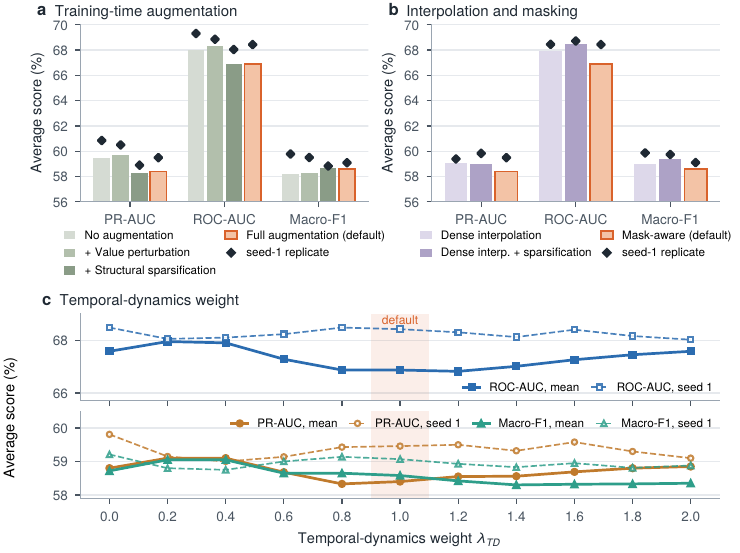}
\caption{Design ablations for GlucoPRISM, reported as scores averaged over the within-cohort benchmark cells. (a) Training-time augmentation. (b) Input interpolation and masking. (c) Temporal-dynamics weight $\lambda_{TD}$; the shaded band marks the default. The default configuration is shown in orange. Bars and solid lines are means; diamonds and dashed lines are a single seed-1 run. Values come from the ablation protocol and are not directly comparable with Table~\ref{tab:master}.}
\label{fig:ablations}
\end{figure}

Following the ablations of \citet{li2026glucofm}, Figure~\ref{fig:ablations} varies three design choices for GlucoPRISM: training-time augmentation, input interpolation and masking, and the temporal-dynamics weight $\lambda_{TD}$. Bars and solid lines are means; diamonds and dashed lines show a single seed-1 run. All tested settings stay close to the default, which indicates that the dispersion does not depend on these inherited design choices; the default is not the best-scoring setting in any panel. Removing augmentation or keeping only value perturbation gives higher ROC-AUC (68.0 and 68.3 against 66.9), dense interpolation with sparsification gives 68.5, and $\lambda_{TD}=1.0$ lies near the ROC-AUC minimum of the sweep. The largest ROC-AUC difference from the default is 1.6 points, and the seed-1 runs differ from the means by up to about 1.5 points, so these ablations do not identify a clearly better setting.

\section{Comparison with a GMI rule}\label{app:gmi}
Following \citet{li2026glucofm}, Table~\ref{tab:gmi} compares GlucoPRISM with a rule that thresholds the glucose management indicator (GMI) computed from seven days of CGM. GlucoPRISM has higher Macro-F1 on Stanford, CGMacros-Dexcom and CGMacros-Fused, with the largest difference on CGMacros-Dexcom (19.5 points). The GMI rule is higher on CGMacros-Libre by 5.8 points. The GlucoPRISM standard deviations are large relative to some differences, particularly on Stanford.

\begin{table}[!htbp]\centering\small
\caption{Macro-F1 (\%) of a seven-day GMI threshold rule and GlucoPRISM. GlucoPRISM values are mean $\pm$ SD; the rule is deterministic. $\Delta$ is GlucoPRISM minus the rule, in percentage points. \textbf{Bold} marks the higher score in each row.}
\label{tab:gmi}
\begin{tabular}{l r >{\columncolor{oursbg}}r r}
\toprule
Cohort & 7-day GMI rule & GlucoPRISM & $\Delta$ \\
\midrule
Stanford & 47.6 & \textbf{57.1}$\pm$6.2 & $+$9.5 \\
CGMacros-Dexcom & 39.6 & \textbf{59.1}$\pm$2.3 & $+$19.5 \\
CGMacros-Libre & \textbf{57.1} & 51.3$\pm$2.6 & $-$5.8 \\
CGMacros-Fused & 44.9 & \textbf{54.1}$\pm$3.3 & $+$9.2 \\
\midrule
Average (4 cohorts) & 47.3 & \textbf{55.4} & $+$8.1 \\
\bottomrule
\end{tabular}
\end{table}

\section{Corpus composition and window geometry}\label{app:corpusfrac}
\begin{figure}[!htbp]
\centering
\includegraphics[width=0.84\linewidth]{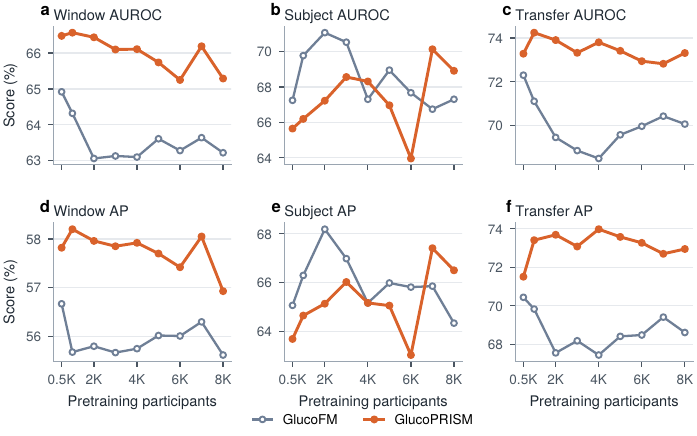}
\caption{HPP pretraining corpus-size comparison on public cohorts. AUROC (top) and average precision (AP; bottom) are percentages; higher is better. Each panel compares identical nested participant subsets and downstream splits. Window and subject scores average 14 task--cohort cells; transfer scores average 12 directions. One pretraining seed is available. Lines connect observations without smoothing; no uncertainty or significance is inferred. Axis ranges differ across panels.}
\label{fig:hpp_scaling_abs}
\end{figure}
\begin{figure}[!htbp]\centering
\includegraphics[width=0.97\linewidth]{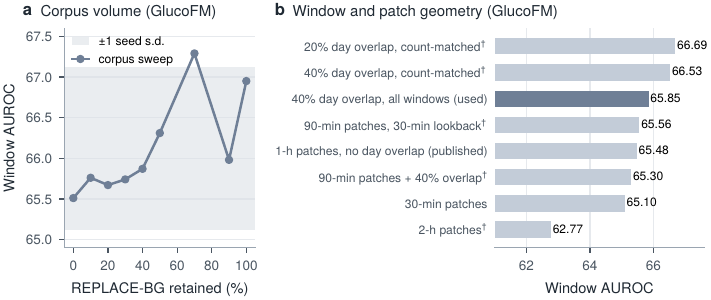}
\caption{\textbf{Corpus composition and window geometry for the GlucoFM backbone.} (a) Window AUROC as the retained fraction of REPLACE-BG varies; the shaded band is $\pm1$ seed s.d. This panel comes from a separate sweep and is not seed-matched to panel (b). (b) Window AUROC by window and patch geometry. The dark bar is the geometry used for every model in this paper (40\% day overlap, all windows kept); \emph{published} is the original GlucoFM geometry. Bars marked $\dagger$ are single runs; the others are three-seed means.}
\label{fig:sweeps}
\end{figure}

\head{Pretraining corpus composition.} Figure~\ref{fig:sweeps} compares corpus size and window geometry for the GlucoFM backbone, not GlucoPRISM.

Retaining between 20\% and 100\% of REPLACE-BG moves the task-averaged score
across a range narrower than the seed band, so hours of signal are not the
binding constraint. Composition is a
different matter. In a leave-one-cohort-out run with one pretraining seed, compared against the three-seed full-corpus means, removing the 177 BIG~IDEAs windows lowers window ROC-AUC by 1.79 points for GlucoFM (64.06 against 65.85) and by 1.20 for GlucoPRISM (66.81 against 68.01), whereas removing all 9{,}040 REPLACE-BG windows lowers it by only 0.23 and 0.27 (65.62 and 67.74), about a 51-fold difference in volume. The same ordering holds for transfer ROC-AUC for both models, and for subject ROC-AUC for GlucoPRISM only. Each removal is a single run compared with a three-seed mean, so these differences are indicative rather than resolved.

 This difference supports retaining varied cohorts in the pretraining corpus.
The released models use the full corpus, including the smaller contributing
cohorts. These sweeps motivate evaluating cohort composition separately from volume; they do not establish that balancing the corpus and deleting Sensor are equivalent interventions.

\head{Window geometry.} Relative to GlucoFM's published geometry (one-hour patches, no day overlap; 65.48), two-hour patches cost $2.72$ \auc{} and 30-minute patches $0.39$, so the one-hour choice is bracketed on both sides. A 30-minute lookback changes the score by $+0.08$. Overlapping consecutive days by 40\% and keeping all windows gives 65.85; this is the configuration used for every model in the paper and the GlucoFM value in Table~\ref{tab:master}. Count-matched overlap variants reach 66.53 and 66.69, but these are single runs; only the published geometry, the used geometry and 30-minute patches have three seeds. This sweep assesses sensitivity to the window geometry inherited from the backbone.

\section{Extended related work}\label{app:related}

\head{CGM foundation models, in more detail.}
GluFormer~\citep{lutsker2026gluformer} established that generative pretraining
on CGM transfers to long-horizon health outcomes. It uses an autoregressive
objective over tokenised glucose, at a scale of 10,812 adults that no public
corpus can currently match. CGM-JEPA~\citep{muhammad2026cgmjepa} argues for the
opposite inductive bias. Predicting \emph{latent} representations of masked
patches rather than raw values yields abstraction that survives a change of
view, and its X-CGM-JEPA variant adds a masked Glucodensity cross-view on the
principle that abstraction is additive across views. GlucoFM~\citep{li2026glucofm},
our backbone, is the most architecturally opinionated of the three. It combines
chronological grid alignment, mask preservation and an explicit slow/fast
decomposition through a learnable causal filter, and reaches the strongest
published results on this benchmark at 0.72M parameters.
CGMformer~\citep{cgmformer2025} takes the complementary discretisation route.

These models learn reusable glucose representations through temporal or distributional prediction. Our addition is a block-directed training objective and a fixed coordinate selection at deployment. The paired-sensor analysis in Appendix~\ref{app:genrobust} characterizes acquisition differences used to design synthetic views. GlucoPRISM does not apply the measured level offset in its synthetic views, so that sensor views do not change the glucose level that Trait is supervised to predict.

\head{Identifiability.} The disentanglement literature's central negative result
is that, without inductive bias, infinitely many generative models produce the
same marginal while differing in their
factorization~\citep{locatello2019challenging}. Weak supervision helps only
under assumptions about how paired samples are
generated~\citep{locatello2020weakly}. Protocol views supply relationships between recordings, but synthetic sensor partners and different-day pairs are imperfect proxies for changes in measurement and physiology. Persistent Trait--State dependence therefore leaves several explanations open, including the supervision, objectives and shared encoder.

\head{Invariance versus erasure.} The distinction that organises our result is
\emph{when} the nuisance is removed. Invariance objectives remove it during
training. That forces the encoder to choose between representing a factor and
suppressing it, and when the factor shares a channel with the label, the choice
is lossy. Post-hoc erasure removes it after the representation already exists.
That avoids the trade, but it requires discovering the subspace by fitting a
probe in the deployment embedding space, with device labels available at that
time. Dispersing the measurement component into a fixed band during training and
filtering it at inference sits between the two. The discovery cost is paid once, at design time, and the
removal itself is free. In prism terms, the measurement band is dispersed during training and filtered
at deployment, rather than suppressed throughout the encoder or estimated afterwards.

\head{Multi-task optimization.} GradNorm adapts task-gradient magnitudes~\citep{chen2018gradnorm}, whereas PCGrad modifies conflicting gradient directions~\citep{yu2020pcgrad}. Our objective-weight sweep tests fixed weights, not either adaptive method. The stronger degradation at higher capacity motivates measuring gradient alignment during training, but does not identify directional conflict. Testing these methods would help distinguish optimization effects from other consequences of the auxiliary objectives (Appendix~\ref{app:arch}).

\end{document}